\documentclass[preprint,12pt]{elsarticle}
\usepackage{amsmath,amsthm}
\usepackage{graphicx}
\usepackage{indentfirst}
\usepackage{verbatim}
\usepackage{algorithm}
\usepackage{algpseudocode}
\usepackage{float}
\usepackage{cancel}
\usepackage{soul}
\usepackage{subfigure}
\usepackage{multicol}
\usepackage{listings}
\usepackage{epsfig}
\usepackage{amsfonts}
\usepackage{amssymb}
\usepackage[para,online,flushleft]{threeparttable}
\usepackage{epstopdf}
\usepackage{color}
\usepackage{epstopdf}
\usepackage{hyperref}
\usepackage{soul}
\usepackage[left=2.54cm, right=2.54cm, top=2.54cm, bottom=2.54cm]{geometry}
\usepackage{titlesec}
\titlespacing\section{0pt}{12pt}{12pt}
\titlespacing\subsection{0pt}{12pt}{12pt}
\titlespacing\subsubsection{0pt}{12pt}{12pt}
\usepackage{bm}
\usepackage{xcolor}
\usepackage{mathtools}
\newtheorem{definition}{Definition}
\newcommand{\bs}{\boldsymbol}
 
\newcommand{\splitatcommas}[1]{	\begingroup
	\begingroup\lccode`~=`, \lowercase{\endgroup
		\edef~{\mathchar\the\mathcode`, \penalty0
			\noexpand\hspace{0pt plus 1em}}		}\mathcode`,="8000 #1	\endgroup
}
\usepackage{tabulary}
\newcolumntype{L}[1]{>{\raggedright\let\newline\\\arraybackslash\hspace{0pt}}m{#1}}
\newcolumntype{C}[1]{>{\centering\let\newline\\\arraybackslash\hspace{0pt}}m{#1}}
\newcolumntype{R}[1]{>{\raggedleft\let\newline\\\arraybackslash\hspace{0pt}}m{#1}}
\usepackage{multirow}

\usepackage{lineno}
\usepackage{setspace}

\begin{document}

\begin{frontmatter}
\title{Multi-scale DeepOnet (Mscale-DeepOnet) for Mitigating Spectral Bias in Learning High Frequency Operators of Oscillatory Functions}
\author[1]{B. Wang}
\ead{bowang@hunnu.edu.cn}
\author[2]{Lizuo Liu}
\ead{lizuo.liu@dartmouth.edu}
\author[3]{Wei Cai\corref{cor1}}
\ead{cai@smu.edu}
\affiliation[1]{organization={LCSM(MOE), School of Mathematics and Statistics, Hunan Normal University},  
            city={Changsha},  
            country={China}}  
\affiliation[2]{organization={Department of Mathematics, Dartmouth College},  
            city={Hanover},  
            state={NH},  
            country={USA}}
\affiliation[3]{organization={Department of Mathematics, Southern Methodist University},  
            city={Dallas},  
            state={TX},  
            country={USA}}  
\date{April 14, 2025}

\cortext[cor1]{Corresponding author}

\begin{abstract}
In this paper, a multi-scale DeepOnet (Mscale-DeepOnet) is proposed to reduce the spectral bias of the DeepOnet in learning  high-frequency mapping between highly oscillatory functions, with an application to the nonlinear mapping between the coefficient of the Helmholtz equation and its solution. The Mscale-DeepOnet introduces the multiscale neural network in the branch and trunk networks of the original DeepOnet, the resulting Mscale-DeepOnet is shown to be able to capture various high-frequency components of the mapping itself and its image. Numerical results demonstrate the substantial improvement of the Mscale-DeepOnet for the problem of wave scattering in the high-frequency regime over the normal DeepOnet with a similar number of network parameters.
\end{abstract}
  
\end{frontmatter}

\section{Introduction}

The DeepONet has shown its ability to learn not only explicit mathematical operators like integration and fractional derivatives, but also PDE operators \cite{cai2021deepm,deng2022approximation,di2021deeponet,lin2021operator,lu2021learning}. However,  like neural networks such as PINN in solving PDEs, neural operators including DeepOnet and FNO demonstrates the behaviour of spectral bias where learning is preferred for low frequency component of the approximation \cite{rahaman2019spectral,Zhi_Qin_John_Xu_2020,sbreview25}. 
The high frequency aspect of a mapping between functions reflects the dependence of the output function on the value of the input function such as in the simple nonlinear mapping of $u(x)=\sin (m a(x))$ for large value of $m$. This kind of functional relationship appears in a similar manner in the relation between the material dielectric properties of a scatterer and the scattering field.  Meanwhile, for high frequency wave scattering, the scattering field $u(x)$ by itself could be an oscillatory function. However, we will need to distinguish the high frequency behavior of the scattering field $u(x)$ and that of the nonlinear mapping as the latter comes from the form of the Green's function of the Helmholtz equations for the scattering problem. It will be seen that the large value $m$ related to the size of the scattering and a larger value corresponds to scattering problems in high frequency regime for a given wavelength of the incidence wave. In this paper, we proposed an enhanced variant of the DeepOnet by introducing the technique of multiscale deep neural network proposed in \cite{mscalednn} into the trunk-net of the DeepOnet structure to reduce the spectral bias of DeepOnet in learning high frequency mapping between functions, especially oscillatory functions.  A previous work on mutiscale FNO \cite{mscaleFNO} has been shown to reduce much the spectral bias of high-frequency mapping.

The rest of the paper is organized as follows. In Section 2, we will review the DeepOnet and the multiscale DeepOnet will be introduced in Section 3. Numerical results for the Mscale-DeepOnet will be given in Section 4. Finally, a conclusion and future work is discussed in Section 5.

\section{DeepOnet}

 The universal approximation theory of neural networks is one of the mathematical basis for the broad applications of neural network, which justifies approximating functions or operators by weighted compositions of some functions, whose inputs are also weighted. The parameters, i.e., the weights and bias, could be obtained by minimizing a loss function with optimization algorithms such as stochastic gradient descent and its variants. Thus, neural network learning turns an approximation problem to one of optimization with respect to the parameters. In this section, we focus on the approximation of operators. 

The work of \cite{tianpingchen1995} gives a constructive procedure for
approximating nonlinear operator \(\mathcal{G}\) between continuous functions $\mathcal{G}\left( f \right)\left( x \right)$ in a compact subset of $C(\mathcal{X}) $ with \(\mathcal{X} \subseteq\mathbb{R}^{d} \) and continuous functions \(f\left( x \right)\) in a compact subset of \(C\left( \mathcal{F} \right)\) with $ \mathcal{F} \subseteq \mathbb{R}^{d}$
\begin{equation}
    \mathcal{G}: f(x)\in C(\mathcal{F}) \rightarrow \mathcal{G}(f)(x)\in
C(\mathcal{X}),
\end{equation}
where \(C\left( \mathcal{X} \right)\) and \(C\left( \mathcal{F} \right)\) are the continuous function spaces over \(\mathcal{X}\) and \(\mathcal{F}\), respectively, and \(\mathcal{X}\) and \(\mathcal{F}\) are compact subsets of \(\mathbb{R}^{d}\), the Euclidean space of dimension \(d\).
The universal approximations with respect to operators are based on the two following results:
\begin{itemize}
\item \textbf{Universal Approximation of Functions \cite{tianpingchen1995}:} Given any $\varepsilon
_{1}>0,$ there exists a positive integer \(N\), \(\left\{  \bm{w}_{k}\right\}_{k=1}^{N}  \in \mathbb{R}^{d}, \left\{ {b}_{k} \right\}_{k=1}^{N}  \in \mathbb{R}\), such that functions $\mathcal{G}(f)(x)$ selected from a compact subset \(\mathcal{U}\) of
$C(\mathcal{X})$ could be uniformly approximated by a one-hidden-layer neural network with
any Tauber-Wiener $\left(  \text{TW}\right)  $\ activation function
$\sigma_{t}$
\begin{equation}
\left|\mathcal{G}(f)(x)-\sum_{k=1}^{N}{\color{black}\mathcal{J}_{k}\left(\mathcal{G}(f)\right)}\sigma_{t}\left(  \bm{w}_{k}\cdot
x+b_{k}\right)  \right|<\varepsilon_{1},\quad\forall\text{ } x\in \mathcal{X},
\label{thm: UAT_function}
\end{equation}
where $\color{black}\mathcal{J}_{k}\left(\mathcal{G}(f)\right)$ is a linear continuous functional defined on $\mathcal{V}$
(a compact subset of $C(\mathcal{F})$), and all $\bm{w}_{k},b_{k}$ are independent of $x$ and \(f\left( x \right)\).
A Tauber-Wiener activation function is defined as follows.
\begin{definition}
Assume \(\mathbb{R}\) is the set of real numbers. $\sigma: \mathbb{R} \rightarrow \mathbb{R}$ is
called a {Tauber-Wiener $\left(  \text{TW}\right) $} function if all the linear combinations
\( g(x) = \sum\limits_{i=1}^{I}c_{i}\sigma\left(  {w}_{i}x+b_{i}\right)
\)
are dense in every $C\left[  a,b\right]  $, where \(\left\{ {w}_{i} \right\}^{I}_{i=1} , \left\{ b_{i} \right\}^{I}_{i=1} , \left\{ c_{i} \right\}^{I}_{i=1} \in\mathbb{R}\) are real constants.
\end{definition}

\medskip
\item \textbf{Universal Approximation of Functionals \cite{tianpingchen1995}:} Given any $\varepsilon_{2}>0$, there exists a positive integer \(M\), \(m\) points  \(\left\{ x_{j} \right\}_{j=1}^{m}  \in \mathcal{F}\) with  real constants \(c_{i}^{k}, {W}_{ij}^{k}, B_{i}^{k} \in \mathbb{R}, i=1,\ldots, M, j=1,\ldots ,m,\) such that a functional $\color{black}\mathcal{J}_{k}\left(\mathcal{G}(f)\right)$ could be approximated by a one-hidden-layer neural
network with any $\text{TW}$ activation
function $\sigma_{b}$
\begin{equation}
\left|{\color{black}\mathcal{J}_{k}\left(\mathcal{G}(f)\right)}-\sum_{i=1}^{M}c_{i}^{k}\sigma_{b}\left(  \sum_{j=1}^{m}{W}_{ij}
^{k}f\left(  x_{j}\right)  +B_{i}^{k}\right)  \right|<\varepsilon
_{2},\quad\forall f\in \mathcal{V}, \label{thm: UAT_functional}
\end{equation}
where the coefficients $c_{i}^{k},{W}_{ij}^{k},B_{i}^{k}$ and nodes
$\{x_{j}\}_{j=1}^{m}$ and $m,M$ are all independent of $f\left( x \right)$.
\end{itemize}

Combining these two universal approximations, the authors of \cite{tianpingchen1995} propose the universal
approximations of nonlinear operators by neural networks when restricted to the compact subset  \(\mathcal{V}\) of
the continuous function space \(C\left( \mathcal{F} \right)\) defined on a compact domain $\mathcal{F}$ in \(\mathbb{R}^{d}\). Namely, given any
$\varepsilon>0,$
 there exists positive integers \(M, N\), \(m\) points  \(\left\{ x_{j} \right\}_{j=1}^{m}  \in \mathcal{F} \subseteq \mathbb{R}^{d} \) with  real constants \(c_{i}^{k}, W_{ij}^{k}, B_{i}^{k} \in \mathbb{R}, i=1,\ldots, M, j=1,\ldots ,m,\)
 \(\left\{  \bm{w}_{k}\right\}_{k=1}^{N}  \in \mathbb{R}^{d}, \left\{ b_{k} \right\}_{k=1}^{N}  \in \mathbb{R}\) that are all independent of continuous functions $f \in \mathcal{V}\subseteq C(\mathcal{F})$ and $ x\in \mathcal{X} \subseteq \mathbb{R}^{d}$ such that
\begin{equation}
\left|\mathcal{G}(f)(x)-\sum_{k=1}^{N} \sum_{i=1}^{M}c_{i}^{k}\sigma_{b}\left(  \sum_{j=1}^{m}W_{ij}
^{k}f\left(  x_{j}\right)  +B_{i}^{k}\right)\sigma_{t}\left(  \bm{w}_{k}\cdot
x+b_{k}\right)  \right|<\varepsilon
\label{thm: UAT_operator}
\end{equation}

\bigskip

\section{Multiscale DeepOnet}
In this section, we introduce a multiscale technique into the DeepOnet framework to enhance its performance in learning operators whose outputs are highly oscillatory functions. The high oscillation could come from the oscillation of the input function or the high frequency nature of the objective operator. 
\subsection{MscaleDNN - multi-scale deep neural network}
\label{subsec3_1}
 The multi-scale DNN(MscaleDNN) \cite{mscalednn} uses a simple parallel architectural framework to approximate functions $f:\Omega \to \mathbb{R}$ with a wide range of spectral content for a general domains $\Omega\subset \mathbb{R}^d$. The MscaleDNN is designed to reduce the spectral bias of DNNs \cite{Zhi_Qin_John_Xu_2020,rahaman2019spectral},
which characterizes the preference toward lower frequencies in the learning dynamics of DNNs. 

The main idea of MscaleDNN can be understood from the following approximation problem. Consider a band-limited function $f$ with a limited frequency range,  
\begin{equation}  
    \hbox{supp}\hat{f}(\bm k)\subset \{\bm k\in \mathbb{R}^d,\;|\bm k| \leq k_{\mathrm{max}}\} . 
\end{equation}  
First, we partition the frequency domain of $f(\bm{x})$  
\begin{equation}  
A_i = \{\bm{k}\in \mathbb{R}^d,\; K_{i-1}\leq|\bm{k}| \leq K_i\},\quad i=1,2,\cdots,M,
\end{equation}  
where $0=K_0 < K_1 <\cdots<K_M=k_{\mathrm{max}}$ and $\hbox{supp}\hat{f}(\bm{k})\subset\bigcup_{i=1}^{M} A_i$. Then, $f$ can be represented as a sum of functions with non-overlapping frequency information  
\begin{equation}  
f(\bm{x}) = \sum_{i=1}^{M}f_i(\bm{x}),\quad f_i(\bm{x}) = \int_{A_i}\hat{f}(\bm{k})e^{j \bm{k}\cdot \bm{x}}\,d\bm{k}.  
\end{equation}  
Next, one performs a radial scaling on in the frequency  domain 
\begin{equation}  
  \hat{f}^{(scale)}_i(\bm{k})  = \hat{f}_i(\alpha_i\bm{k}),  
\end{equation}  
which corresponds to scaling in the physical domain
$f_i(\bm{x}) = \alpha_i^n f^{(scale)}_i(\alpha_i\bm{x}).$
If the scaling factor is sufficiently large, the scaled function becomes a low-frequency function with a Fourier transform compactly supported at lower frequencies,
\begin{equation}  
 \hbox{supp}\hat{f}_i^{(scale)}(\bm{k})\subset \Big\{\bm{k}\in \mathbb{R}^d,\;\frac{K_{i-1}}{\alpha_i}\leq|\bm{k}| \leq \frac{K_i}{\alpha_i}\Big\},\quad i=1,2,\cdots,M.  
\end{equation}  

Due to the spectral bias of the DNNs toward low frequencies, $f^{(scale)}_i(\bm{x})$ can be quickly learned by a DNN $f_{\theta_i}(x)$ parameterized by $\theta_i$,   
\begin{equation}  
    f_i(\bm{x}) =\alpha_i^n f^{(scale)}_i(\alpha_i\bm{x}) \sim \alpha_i^n f_{\theta_i}(\alpha_i \bm{x}).  
\end{equation}  
Since $f(\bm{x})= \sum_{i=1}^{M}f_i\left(\bm{x}\right),$ the MscaleDNN approximation can be found by  
\begin{equation}
f(\bm{x}) \sim \sum_{i=1}^{M} \alpha_i^n f_{\theta_i}(\alpha_i\bm{x}).
\label{MscaleDNN_e}
\end{equation}


The MscaleDNN uses a series of sub-networks with different scaled inputs to approximate the target function $f(\bm{x})$ at different frequency ranges. It has been shown that it is able to capture features across multiple scales simultaneously through this parallel network structure. $f_{\theta_i}(\alpha_i\bm{x})$ denotes an individual neural network with parameters $\theta_i$ operating on scaled input $\alpha_i\bm{x}$, where $\alpha_i$  represents the scaling factor. A property of this structure is that larger values of $\alpha_i$ enable the corresponding sub-network to capture and learn higher-frequency components of the target function $f(\bm{x})$. Moreover, the scaling factors $\alpha_i$ could also be made as trainable parameters for optimal performance.

\subsection{Multiscale DeepOnet}
Following the idea in MscaleDNN \cite{mscalednn}, it is quite easy to introducing multi-scale techniques in the DeepOnet framework. An operator given by a multi-scale DeepOnet has the form 
\begin{equation}
\mathcal{G}(f)(x)\sim\sum\limits_{s=1}^{S_{\rm trunk}}\sum_{k=1}^{n_t}\underbrace{\left[\sum\limits_{i=1}^{S_{\rm branch}}\mathcal N_{Br,sn_t+k}\left(
\left\{\alpha_if\left( x_{j}\right)\right\}_{j=1}^{m}  \right)\right]}_{Br_k}
\underbrace{\mathcal N_{T,k}\left( \beta_s  x\right)}_{T_k}.
\label{msdon_formula}
\end{equation}
Here $S_{\rm branch}, S_{\rm trunk}$ is the number of scales and $\{\alpha_s\}_{s=1}^{S_{\rm branch}}$, $\{\beta_s\}_{s=1}^{S_{\rm trunk}}$ are the scaling factors. They are hyper parameters and usually depend on what kind of problem to solve. In this architecture, neural networks with different scales are stacked to build the trunk net $[\mathcal N_{T}(\beta_1x), \mathcal N_{T}(\beta_2x), \cdots, \mathcal N_{T}(\beta_{S_{\rm trunk}}x)]$, while the branch net consists of a summation of neural networks with different scales. In the multiscaled trunk net, each sub-network has $n_t$ outputs. Thus, the total number of basis functions generated by the trunk net is $S_{trunk}\times n_t$ which should be equal to the number of outputs of each sub-network in the multiscale branch net.

In this work, we are particularly interested in using the multiscale DeepOnet to learn the solution operator of Helmholtz equation when the wave number is large. As the solutions are complex valued functions, we use two branch nets to generate real and imaginary parts of the expansion coefficients associated to the basis functions learned by the trunck nets. Then, the complex valued  multiscale DeepOnet 
\begin{equation}
\mathcal{G}(f)(x)\sim\sum\limits_{s=1}^{S_{\rm trunk}}\sum_{k=1}^{n_t}\underbrace{\left[\sum\limits_{i=1}^{S_{\rm branch}}\mathcal N_{Br}^{c}\left(
\left\{\alpha_if\left( x_{j}\right)\right\}_{j=1}^{m}  \right)\right]}_{Br_k}
\underbrace{\mathcal N_{T,k}\left( \beta_s  x\right)}_{T_k}.
\label{cmsdon_formula}
\end{equation}
use the following complex valued branch net
\begin{equation}
 \mathcal N_{Br}^{c}\left(
\left\{\alpha_if\left( x_{j}\right)\right\}_{j=1}^{m}  \right)=   \mathcal N_{Br}^{\rm real}\left(
\left\{\alpha_i f\left(  x_{j}\right)\right\}_{j=1}^{m}  \right)+{\rm i}\sum\limits_{i=1}^{S_{\rm branch}}\mathcal N_{Br}^{\rm imag}\left(
\left\{\alpha_i f\left(  x_{j}\right)\right\}_{j=1}^{m}  \right).
\end{equation}

\section{Numerical Results}

\subsection{Nonlinear mapping}	We will consider the approximation of the nonlinear operator 
\begin{equation}
\mathcal G_K[a](x)=\sum\limits_{n=0}^{K}[A_n\sin(na(x))+B_n\cos(na(x))],\quad  x\in\Omega,
\label{nlmap}
\end{equation}
where $\{A_n, B_n\}_{n=0}^K$ are given coefficients. 
This operator embodies the nonlinear connection between the material property of a media  $a(x)$ and the scattering wave $u(x)$ in a wave scattering problem described by the following Helmholtz equation
\begin{equation}
     u^{\prime \prime} + k^2(1+a(x)) u =g(x), \qquad x \in \mathbb R,
    \label{helm1}
\end{equation}
with a Sommerfeld boundary condition 
\begin{equation}
    \frac{du^{\rm sc}(x)}{dx}\pm i ku^{\rm sc}(x)\to 0, \quad \bm{x}\to \pm\infty,
\end{equation}
and the support of $a(x)$, corresponding to the scatterer, is compact-supported inside $\Omega$.

The solution of Eq. \eqref{helm1} can be found by a volume integral equation n term of the Green's function $G(x,x^\prime)$,
\begin{equation}
    u(x)=\int_\Omega G(x,x^\prime)    f(x^\prime) d x^\prime
    \label{ie}
\end{equation}
where
\begin{equation}
    f(x)=g(x)-k^2a(x)u
\end{equation}
and the Green's function is given by
\begin{equation}
        G(x,x^\prime) =\frac{\rm i}{2k}e^{{\rm i} k |x-x^\prime|}.
    \label{gf}
\end{equation}

\begin{figure}[ht!]
\center
\subfigure[$S_{branch}=S_{trunk}=1$]{\includegraphics[scale=0.25]{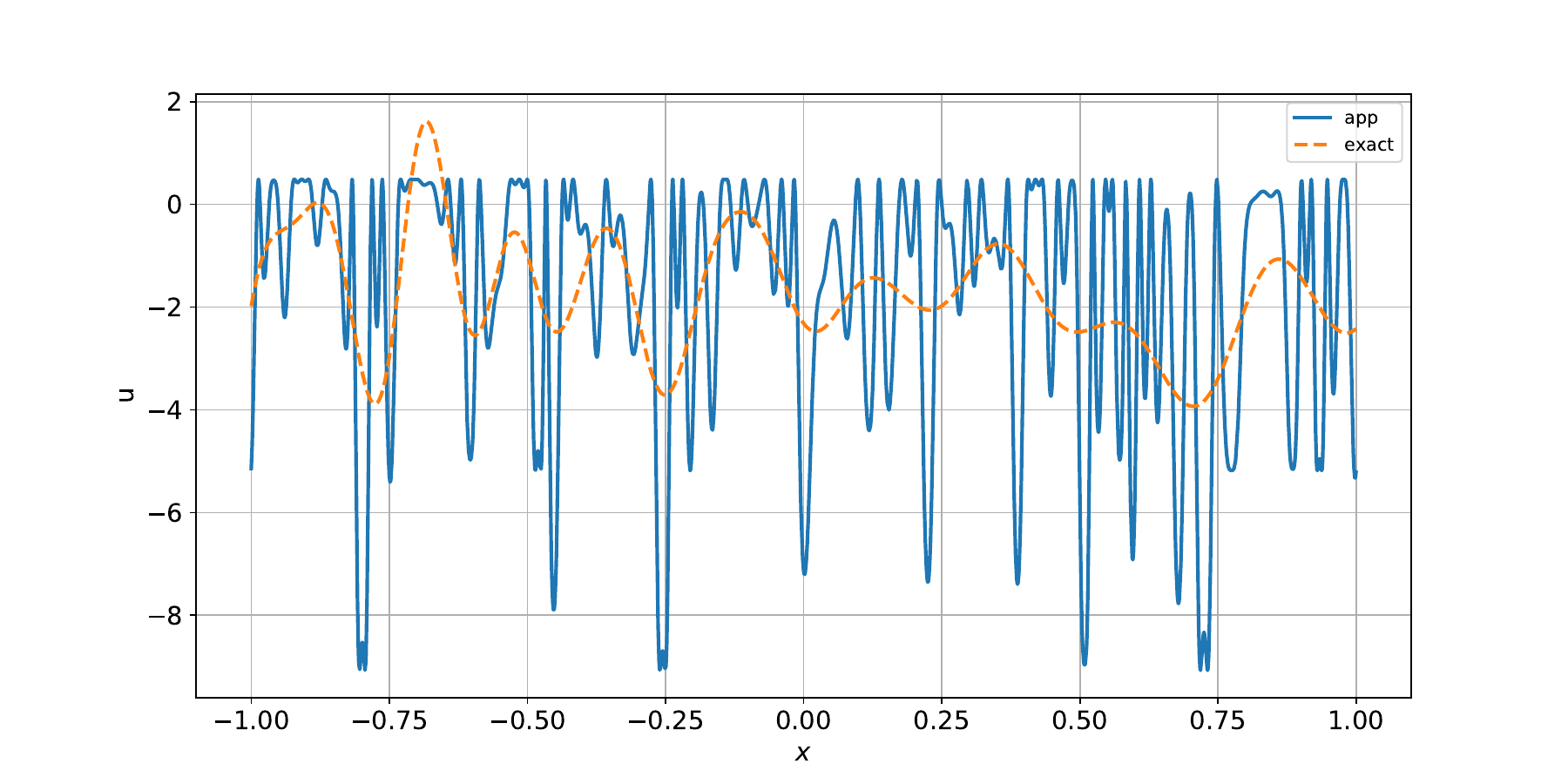}}
\subfigure[$S_{branch}=S_{trunk}=1$]{\includegraphics[scale=0.25]{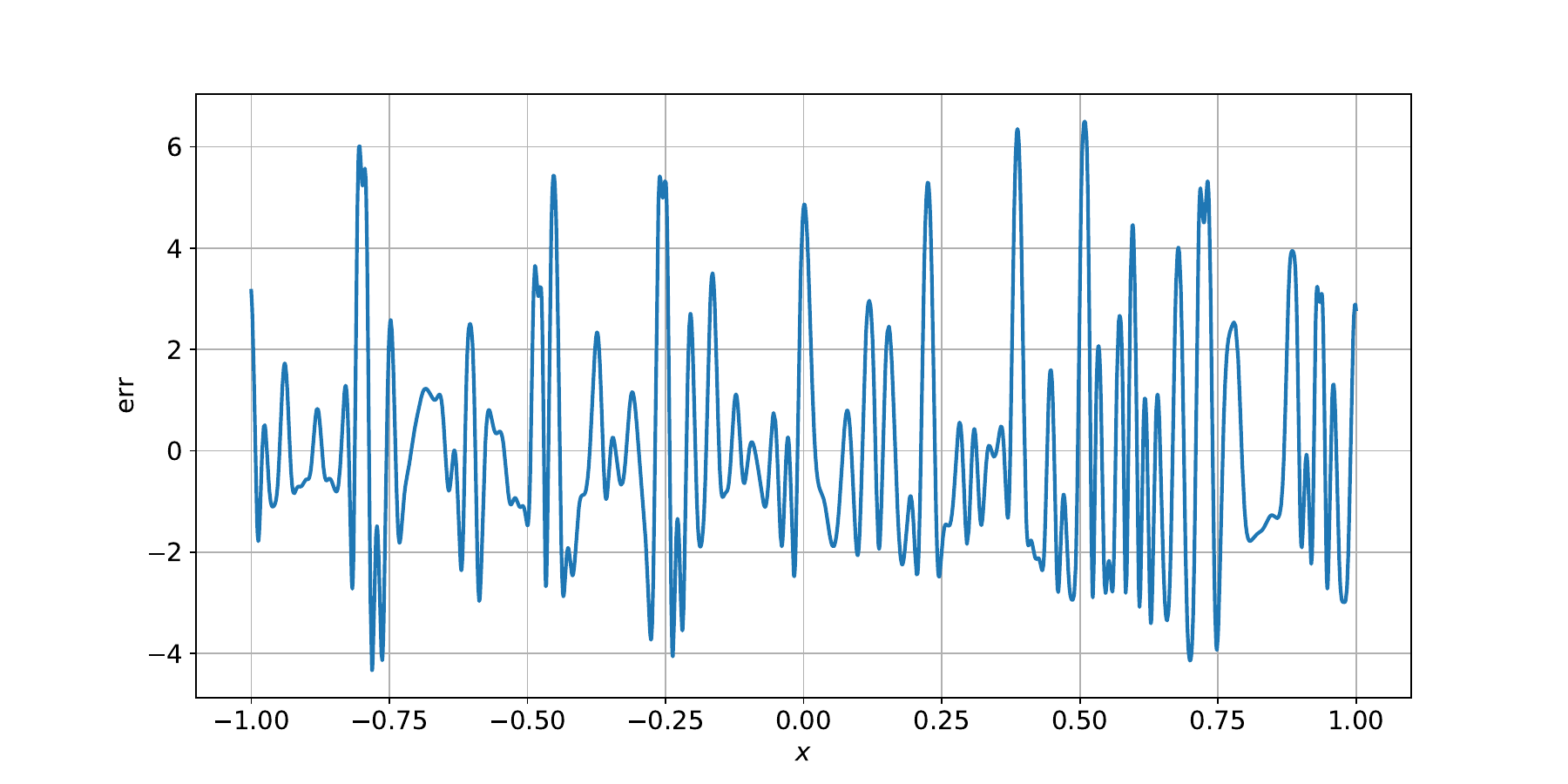}}\\
\subfigure[$S_{branch}=1,S_{trunk}=10$]{\includegraphics[scale=0.25]{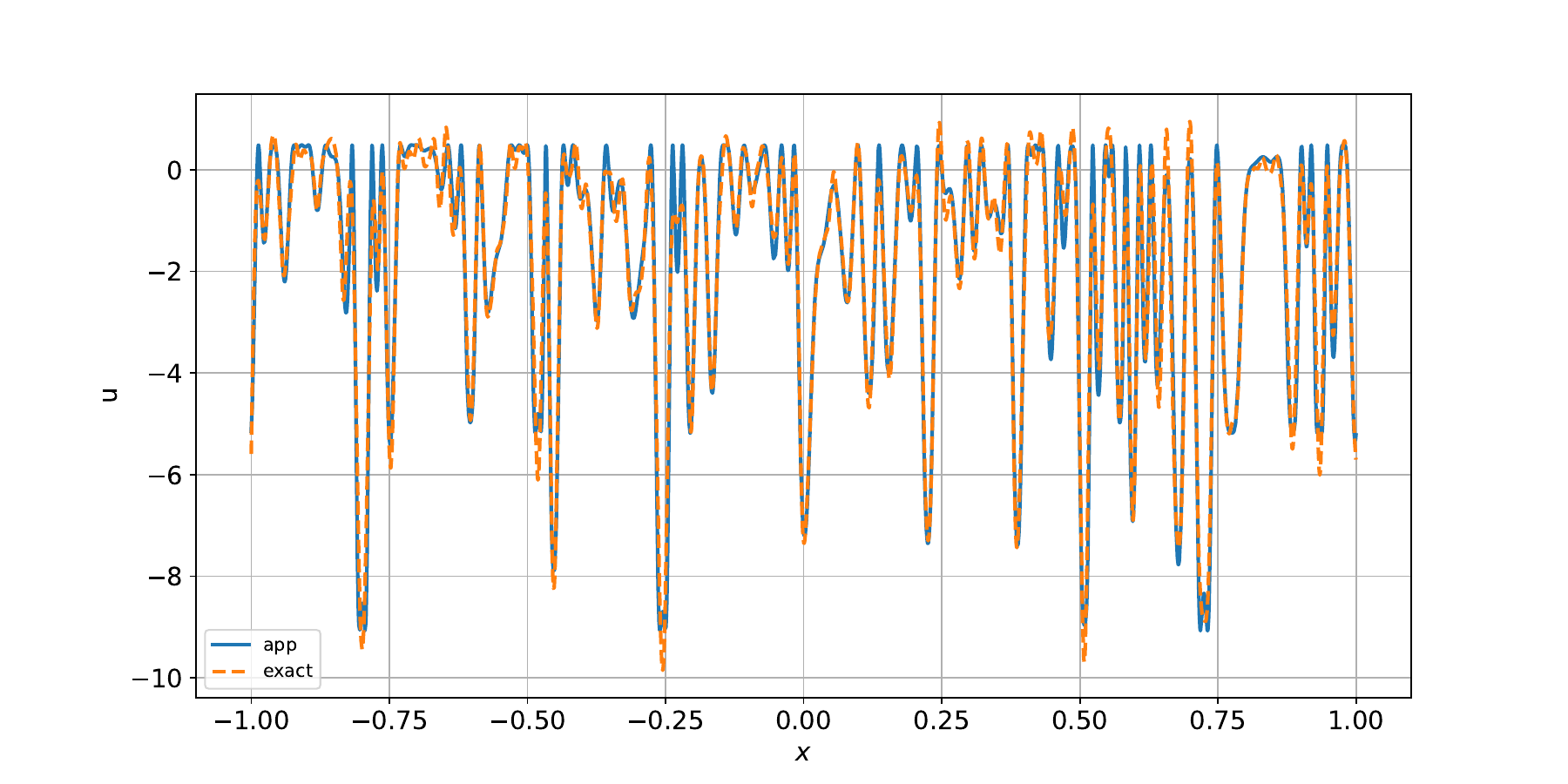}}
\subfigure[$S_{branch}=1,S_{trunk}=10$]{\includegraphics[scale=0.25]{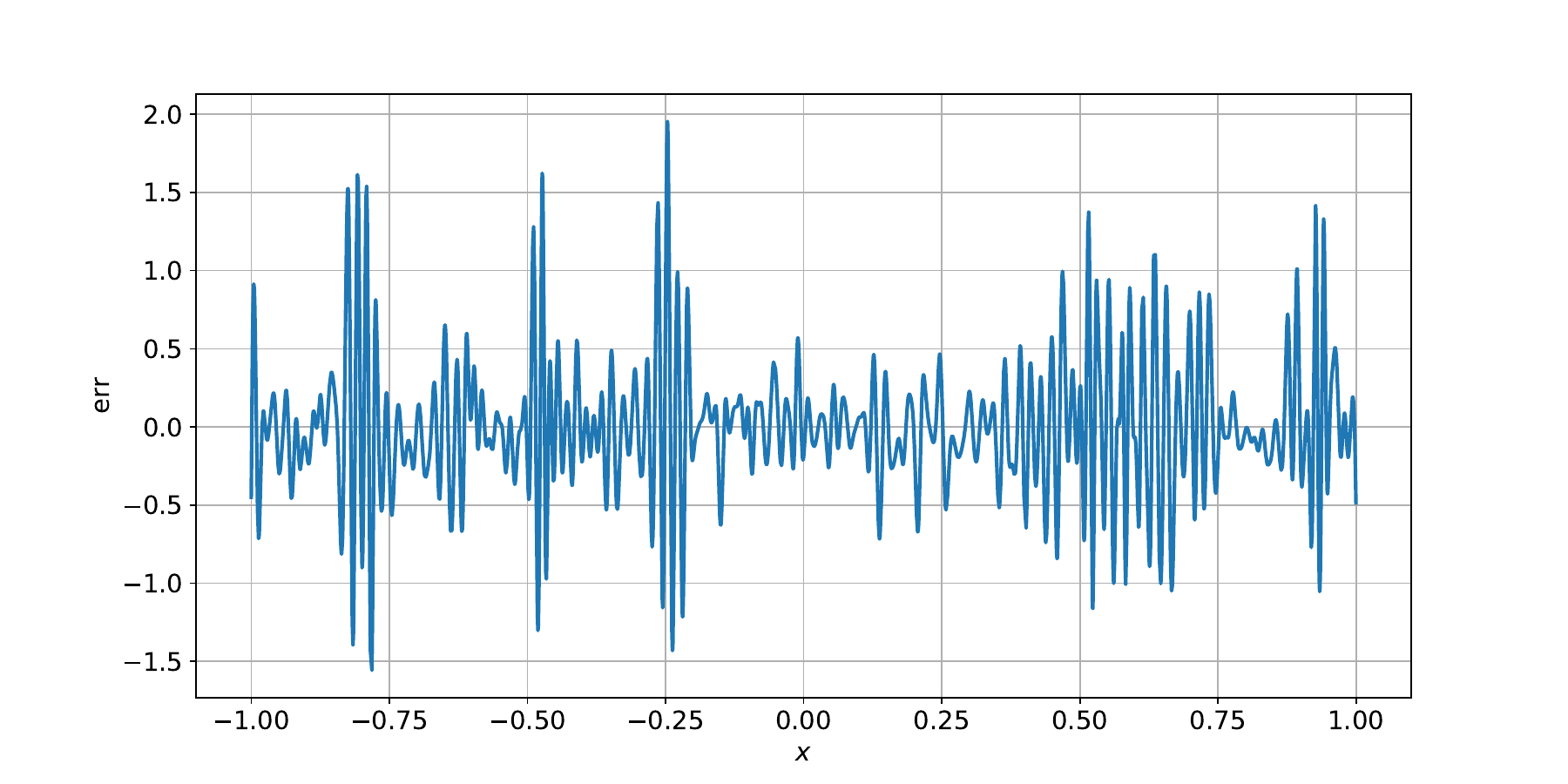}}\\
\subfigure[$S_{branch}=5,S_{trunk}=10$]{\includegraphics[scale=0.25]{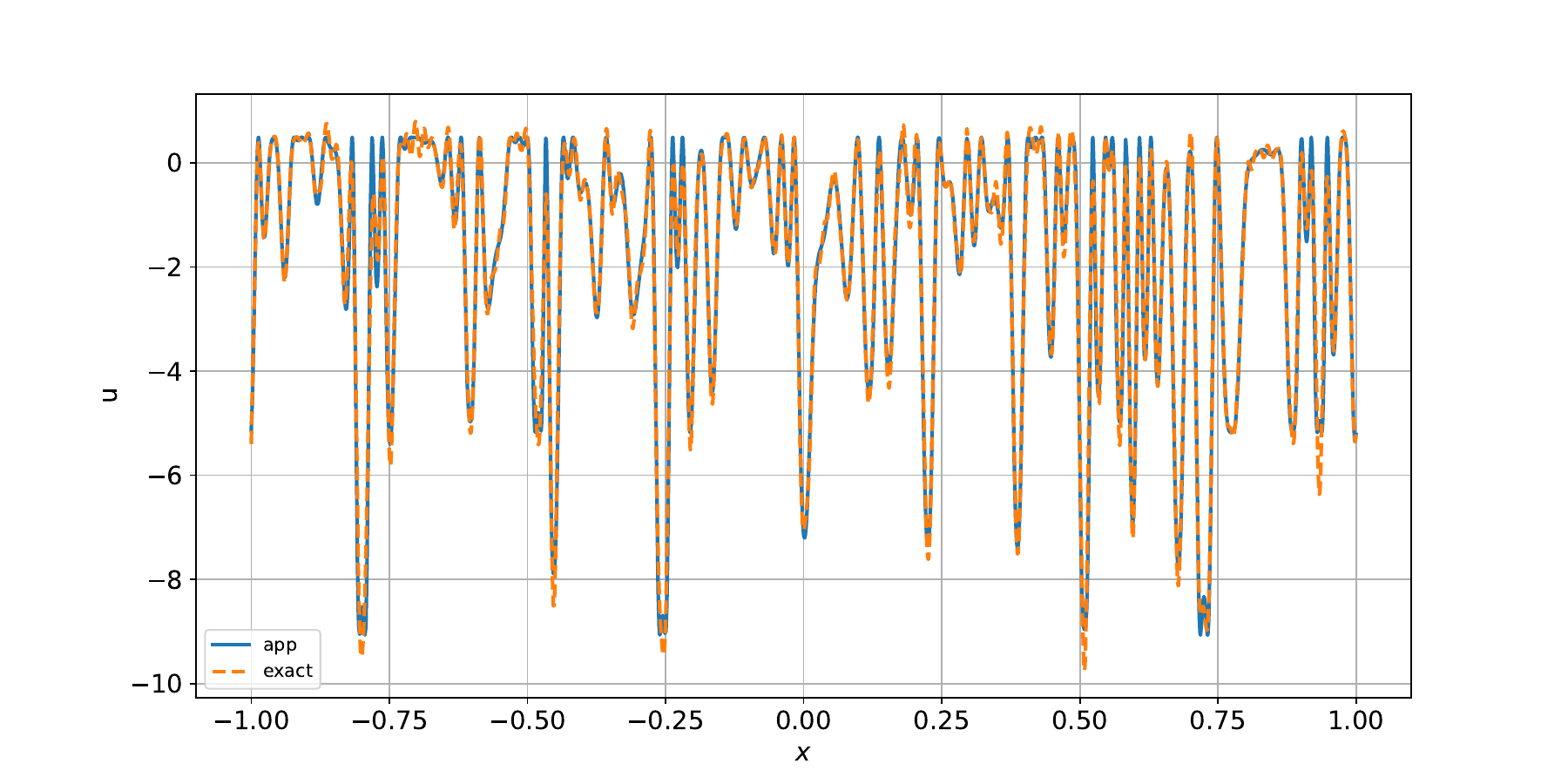}}
\subfigure[$S_{branch}=5,S_{trunk}=10$]{\includegraphics[scale=0.25]{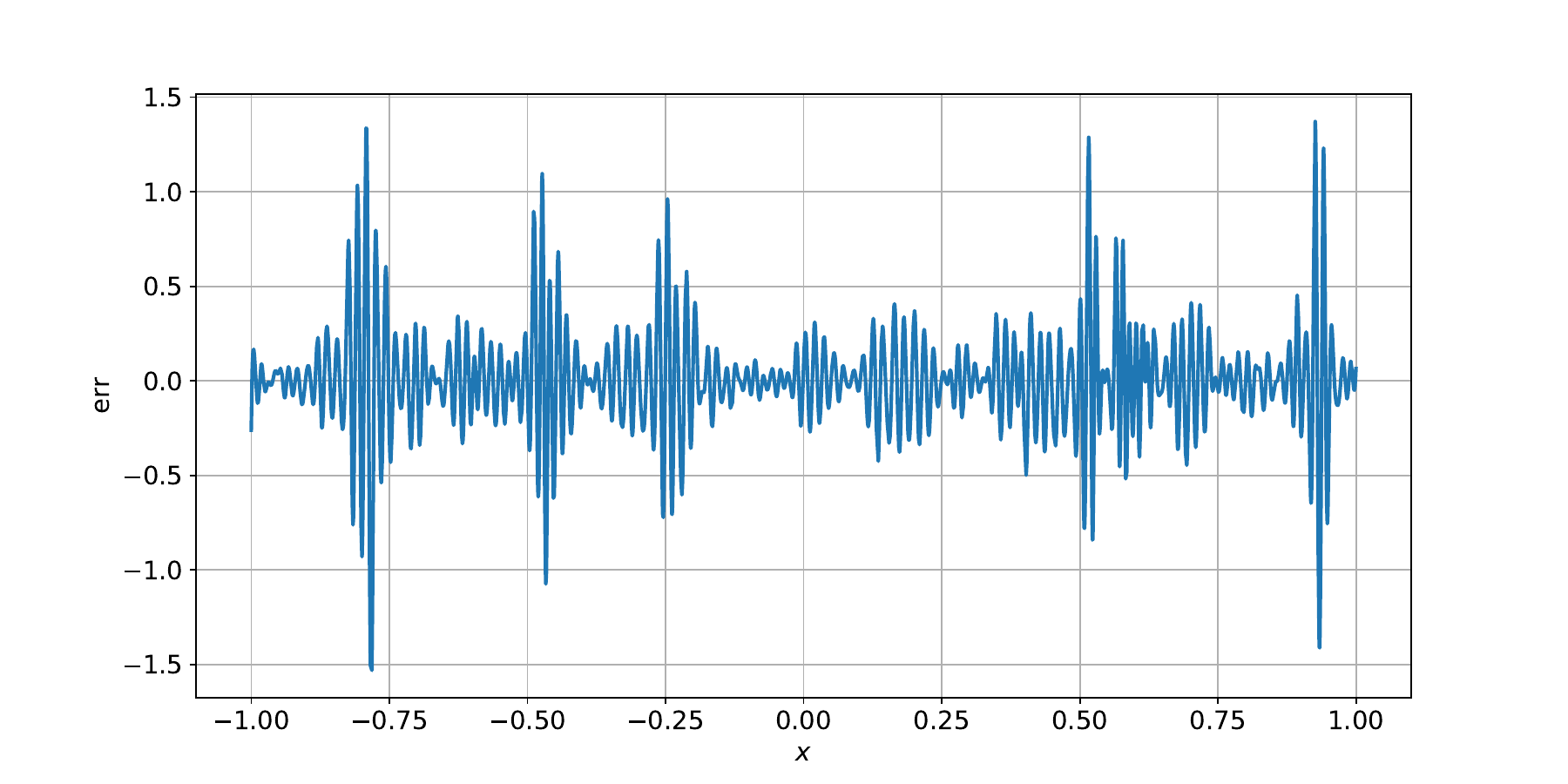}}\\
\caption{Approximations of the nonlinear mapping $\mathcal G_{50}[a]$ on a training data.}
\label{mapping_test_1_fig_1}
\end{figure}

\begin{figure}[ht!]
\center
\subfigure[$S_{branch}=S_{trunk}=1$]{\includegraphics[scale=0.25]{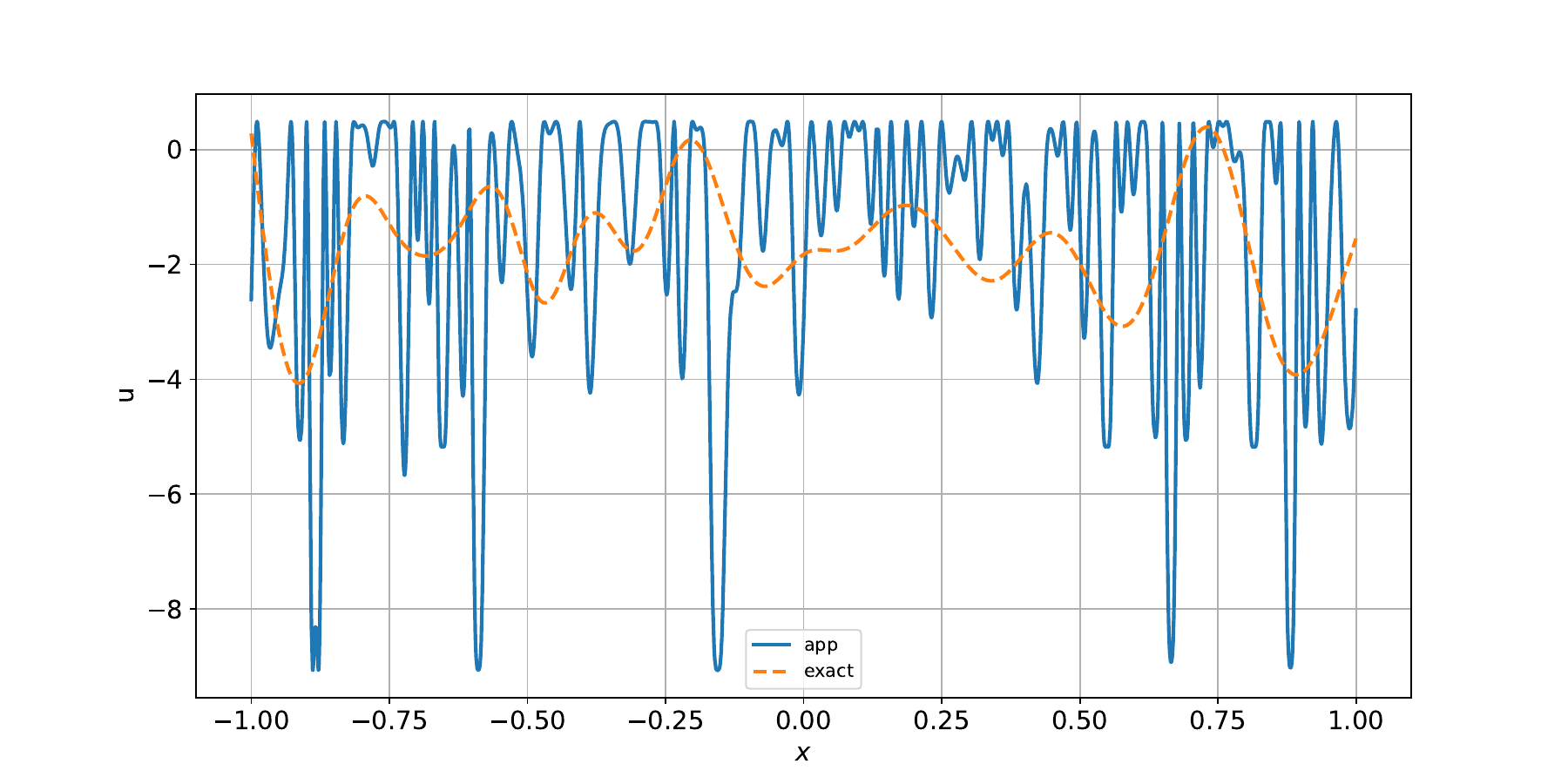}}
\subfigure[$S_{branch}=S_{trunk}=1$]{\includegraphics[scale=0.25]{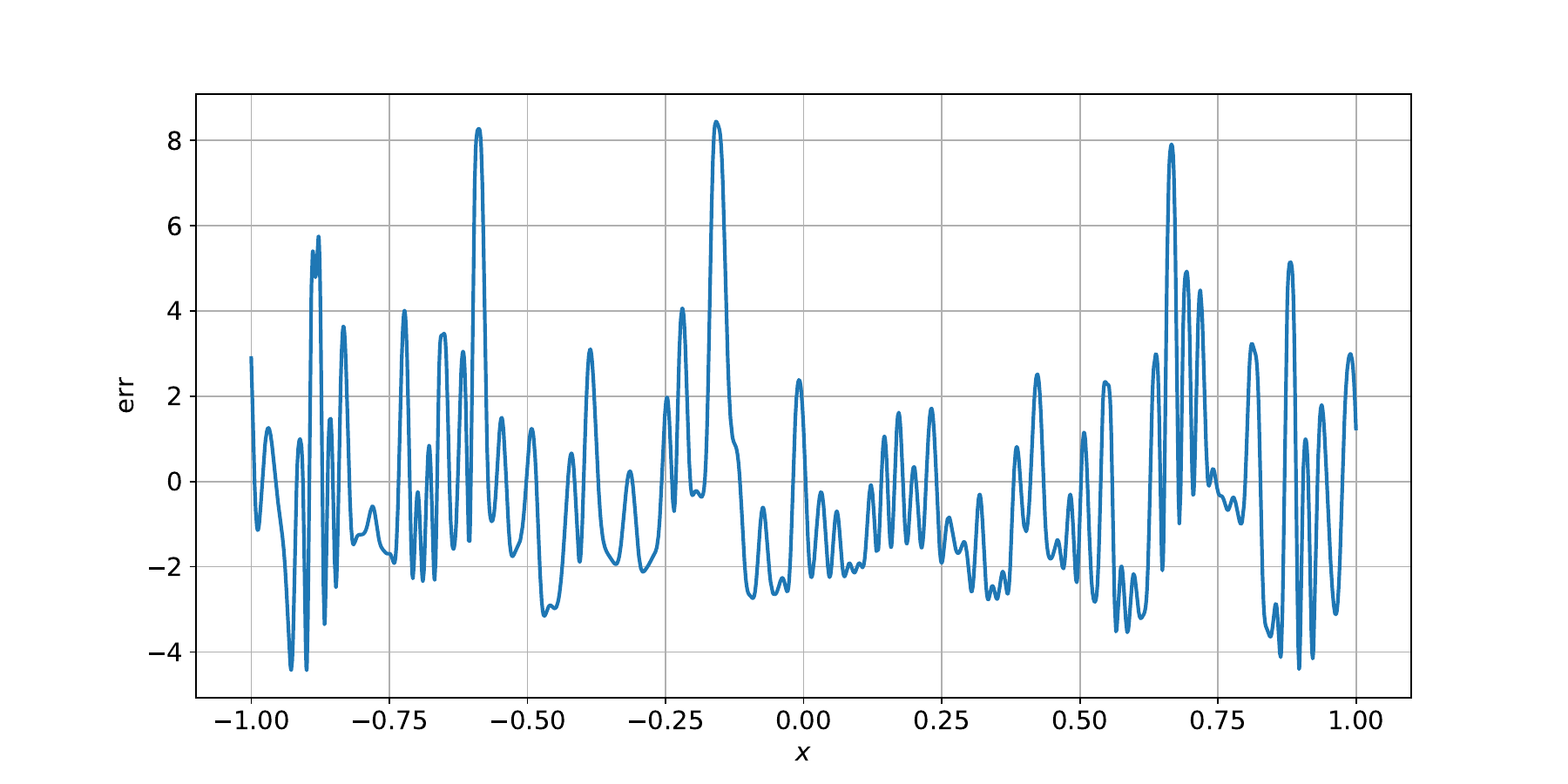}}\\
\subfigure[$S_{branch}=1,S_{trunk}=10$]{\includegraphics[scale=0.25]{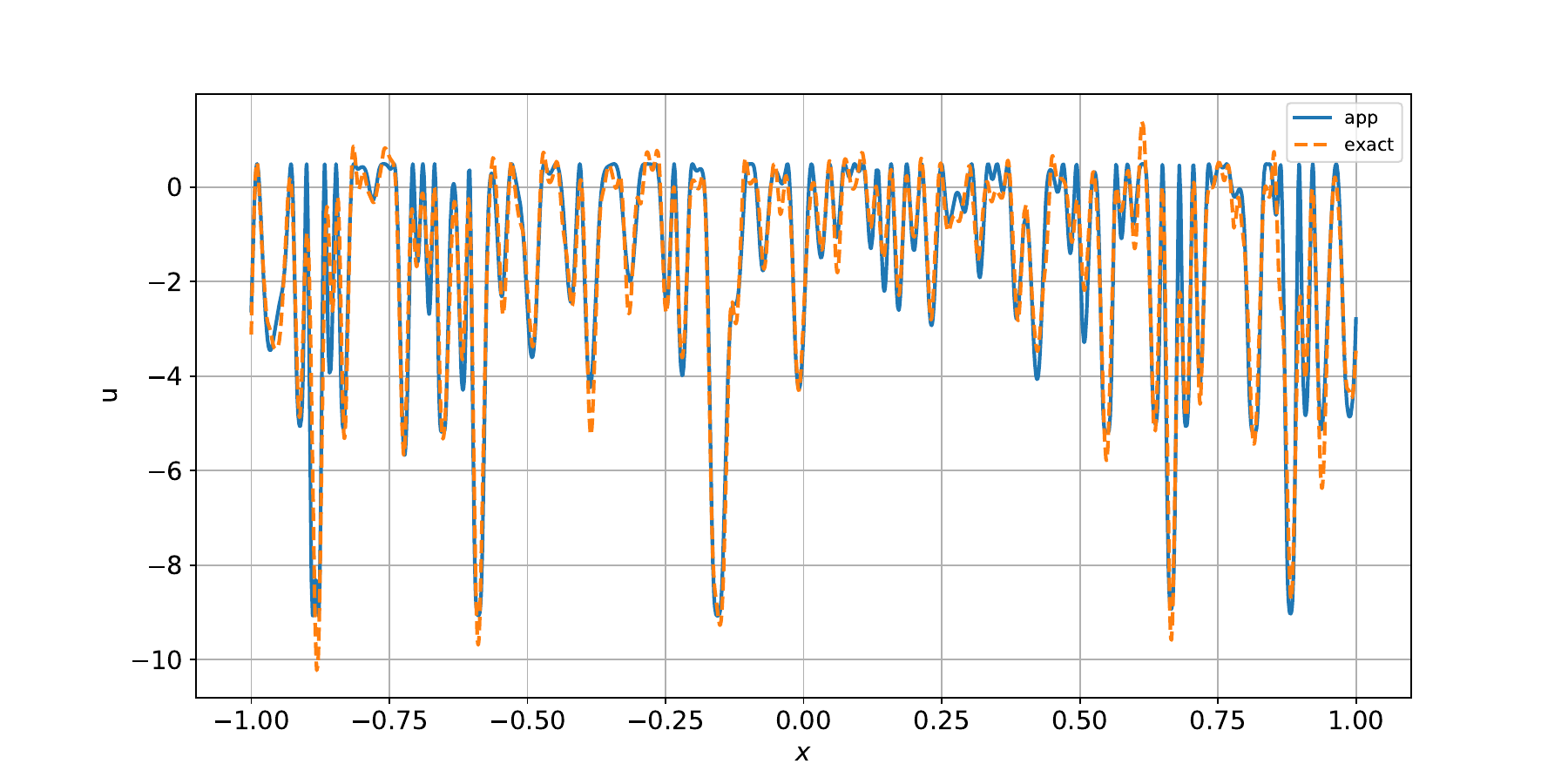}}
\subfigure[$S_{branch}=1,S_{trunk}=10$]{\includegraphics[scale=0.25]{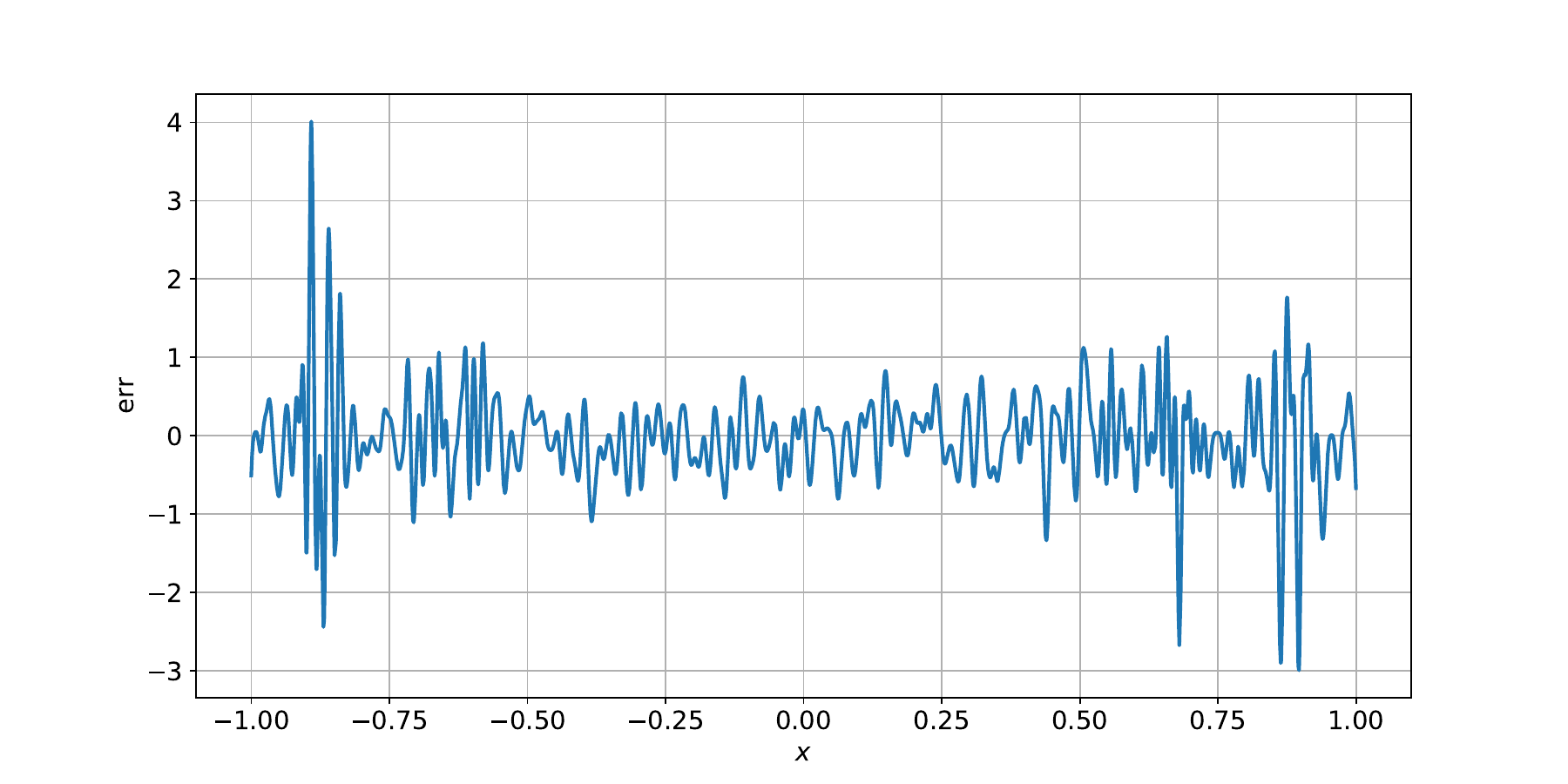}}\\
\subfigure[$S_{branch}=5,S_{trunk}=10$]{\includegraphics[scale=0.25]{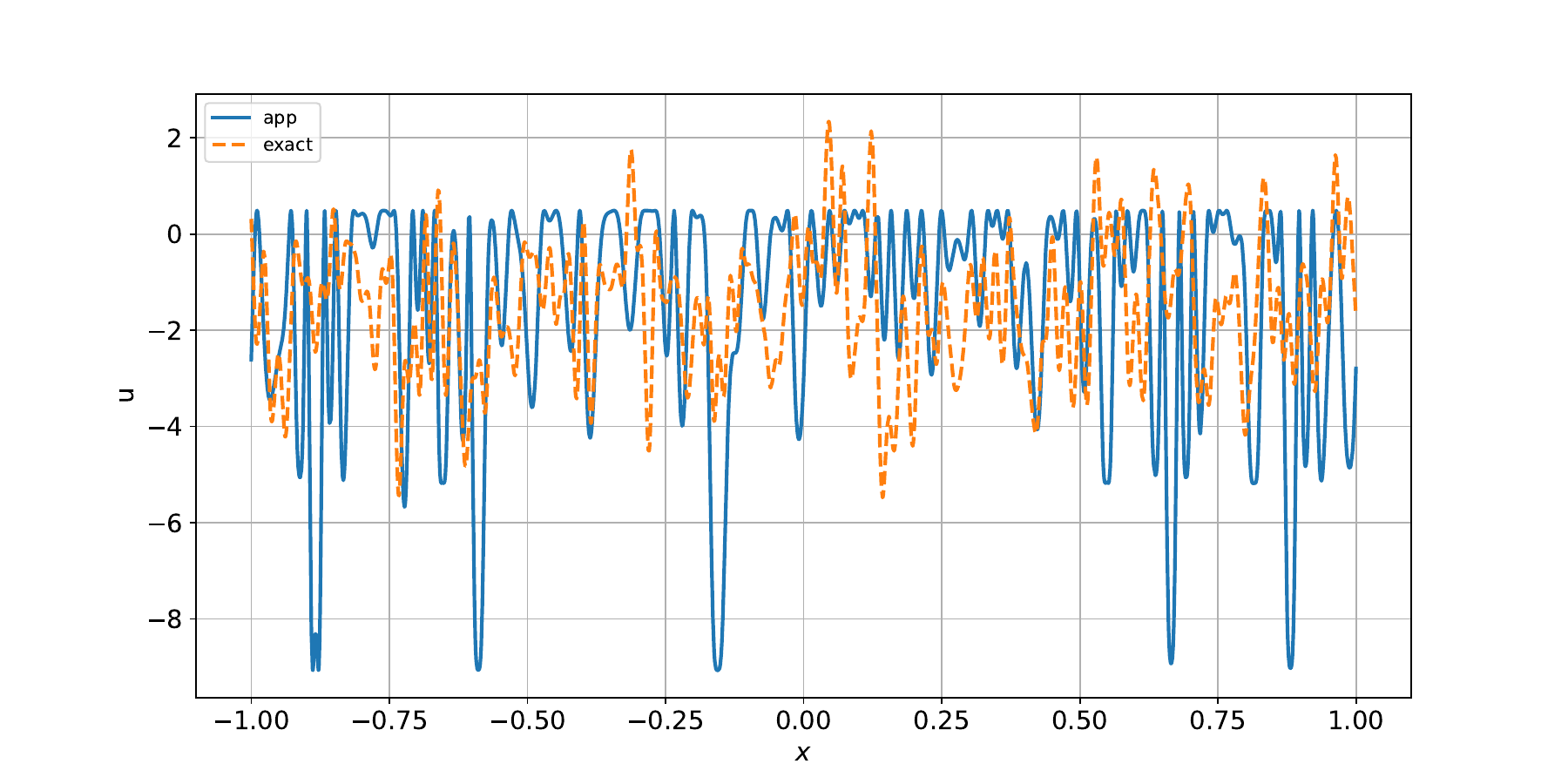}}
\subfigure[$S_{branch}=5,S_{trunk}=10$]{\includegraphics[scale=0.25]{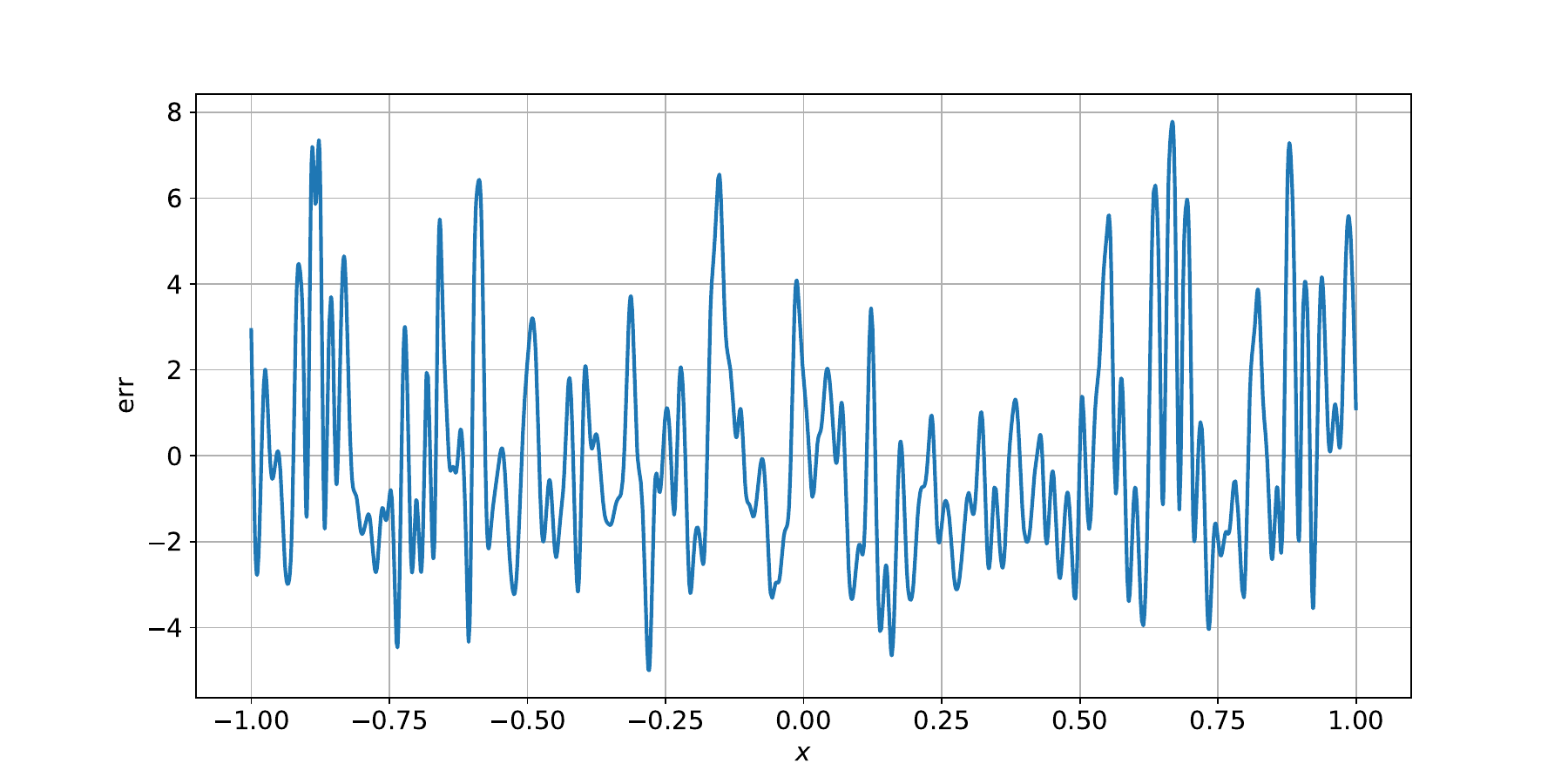}}\\
\caption{Approximations of the nonlinear mapping $\mathcal G_{50}[a]$ on a testing data.}
\label{mapping_test_1_fig_2}
\end{figure}

For numerical test on the nonlinear mapping $\mathcal G_K[a]$  in  \eqref{nlmap} , we consider the domain $\Omega=[-1,1]$, $K=50$ and use $M$ - mode Fourier series to generate training and testing data sets as follows
\begin{equation}
    a_n(x)=c\hat a_n(x),\quad \hat a_n(x)=b_0^{(n)}+\sum\limits_{j=1}^{M}[b_j^{(n)}\sin(j\pi x)+c_j^{(n)}\cos(j\pi x)], \quad  M=50,
\end{equation}
where $\{b_j^{(n)}\}_{j=0}^{50},\{c_j^{(n)}\}_{j=1}^{50}$ are uniformly distributed random numbers in $[-1, 1]$.
We generate $5000$ and $100$ $(a_n(x), \mathcal G_{50}[a_n](x))$ pairs  for training and testing, respectively. For each train function $a_n(x)$, we randomly sample $3000$ values $\{a_n(x_j)\}_{j=1}^{3000}$, at uniformly distributed points $x_j\in [-1, 1]$. We tested the original and Multi-Scale DeepOnet with training rate equal to $1e-3$. Three neural networks with architecture given in Table \ref{mapping_test_1_table_1} are tested. The approximation errors on an arbitrary selected training and testing data are plotted in Fig. \ref{mapping_test_1_fig_1}-\ref{mapping_test_1_fig_2}. The numerical results show that multiscale technique applied to the trunk net can significantly improve the performance in learning operators with highly oscillated output functions. However, multiscale technique in branch net has much less impact on the performance of operator learning, which might be due to the fact that the range of the value of the input function $a(x)$ is limited.
\begin{table}[htbp]
	\centering
	\begin{tabular}{|c|c|c|c|}
    \hline
     $S_{branch}$ & $S_{trunk}$   & architecture of the subnetworks    &    $\#$ parameters  \\
    \hline
            1   &   1  &  $\begin{array}{l}
        \mathcal N_{Br}: [3000, 2000,1000, 500, 501]\\
        \mathcal N_T: [ 1,   500, 500, 500, 500]
        \end{array}$ & 9,506,000\\
        \hline
		   1  &    10     &     $\begin{array}{l}
        \mathcal N_{Br}: [3000, 2000,1000, 500, 501]\\
       \mathcal N_T: [ 1,   50, 50, 50, 50]
        \end{array}$    &   8,831,000\\
        \hline
            5  &    10      &    $\begin{array}{l}
        \mathcal N_{Br}: [3000, 2000,1000, 500, 501]\\
       \mathcal N_T: [ 1,   50, 50, 50, 50]
        \end{array}$    &   43,847,000\\
         \hline
	\end{tabular}
	\caption{Network architectures and number of parameters for the case $M=50$ Fourier modes for the input function. }
    \label{mapping_test_1_table_1}
\end{table}

\subsection{Helmholtz scattering problem: mapping from dielectric parameter $a(x)$ to the scattering field} We will consider the approximation of the solution operator 
\begin{equation}
    \mathcal S: a(x)\rightarrow u(x)
\end{equation}
 of the $1$-dimensional wave scattering problem 
\begin{equation}
    \Delta u(x) + k^2(1+a(x)) u(x) =0, \qquad x \in \mathbb{R} ,
    \label{helm}
\end{equation}
with Sommerfeld radiation condition 
\begin{equation}
    \frac{du^{\rm sc}(x)}{dx}\pm ku^{\rm sc}(x)\to 0, \quad \bm{x}\to \pm\infty,
\end{equation}
on the scattering field $u^{\rm sc}(x)$, where  $u(x)=u^{\rm sc}(x)+u^{\rm inc}(x)$ is the total field and the incident field $u^{\rm inc}$ is assumed to satisfy homogeneous Helmholtz equation
$$\Delta u^{\rm inc}(x)+k^2u^{\rm inc}(x)=0,\quad x\in\mathbb R.$$
Subtracting the above equation from \eqref{helm} gives
\begin{equation}
    \Delta u^{\rm sc}+k^2(1+a(x))u^{sc}=g(x),\quad x\in \mathbb R,
\end{equation}
where the source term is given by the incident filed $u^{\rm inc}(x)$ as follows
\begin{equation}
    g(\bs x)=-k^2a(x)u^{\rm inc}(x).
\end{equation}
Further, the scatterer described by $a(x)$ is assumed to be finite, i.e, $a({x})$ has compact-support inside a bounded domain $\Omega$. 

 \bigskip
 
\noindent{\bf Training data} We use an integral method to generate training and testing data. The solution of the scattering problem can be written into a integral formulation
\begin{equation}
\begin{split}
u^{\rm sc}(x)=&\int_RG(x-x')g(x')dx'-k^2\int_RG(x-x')a(x')u^{\rm sc}(x')dx'\\
=&\int_{\Omega}G(x-x')g(x')dx'-k^2\int_{\Omega}G(x-x')a(x')u^{\rm sc}(x')dx'\\
\triangleq&\mathcal N[g](x)-\mathcal K[u](x)\quad x\in\Omega,
\end{split}
\end{equation}
where
$$G(x,x^\prime) = \frac{\rm i}{2k}e^{{\rm i} k |x-x^\prime|},$$
is the Green's function of the one-dimensional Helmholtz equation.

Many numerical method has been proposed to discretize the integral equation 
\begin{equation}
( \mathcal I+\mathcal K)[u](x)=\mathcal N[g](x),\quad x\in\Omega.
\end{equation}
The Nystrom method is adopted here. Assume $\mathcal T_h=\{K_e=[x_e, x_{e+1}]\}_{e=0}^{N-1}$ is a mesh of the domain $\Omega$, $\{u_e\}_{e=0}^{N}$ are the nodal values of a linear approximation of $u(x)$ on $\mathcal T_h$. The numerical scheme using the mesh points $\{x_e\}_{e=0}^N$ as collocation points is given by 
\begin{equation}\label{equ1-8}
u_i+\sum\limits_{e=0}^{N-1}(A_{ie}^+u_e+A_{ie}^-u_{e+1})=b_i,\quad i=0, 1, \cdots, N.
\end{equation}
where 
\begin{equation*}
\begin{split}
A_{ie}^+=&\frac{k^2h_e}{2}\int_{-1}^1G(x^i, X^e(\xi))a(X^e(\xi))\dfrac{1-\xi}{2}\mathrm{d}\xi,\quad A_{ie}^-=\frac{k^2h_e}{2}\int_{-1}^1G(x^i, X^e(\xi))a(X^e(\xi))\dfrac{1+\xi}{2}\mathrm{d}\xi,\\
b_i=&\sum\limits_{e=0}^{N-1}\frac{h_e}{2}\int_{-1}^{1}G(x^i, X^e(\xi))g(X^e(\xi))d\xi.
\end{split}
\end{equation*}
The above integrals can be approximated by simply using a two points Gauss quadrature. 

\noindent {\bf $M$ - Fourier mode representation of input data $a(x)$.} We use the M-mode Fourier series to represent the function space of training and testing function sets as follows
\begin{equation}
    a_n(x)=c\hat a_n(x),\quad \hat a_n(x)=b_0^{(n)}+\sum\limits_{j=1}^{M}(b_j^{(n)}\sin(j\pi x)+c_j^{(n)}\cos(j\pi x)
\end{equation}
where $\{b_j^{(n)}\}_{j=0}^M, \{c_j^{(n)}\}_{j=0}^M$ are random numbers in $[-1, 1]$, $c$ is a given constant to control the perturbation of the inhomogeneous media. The number of the input functions in the training and testing datasets are denoted by $N_{train}$ and $N_{test}$, and the number of sampling points for the input and target functions are denoted by $N_{sample}^{in}$ and $N_{sample}^{out}$. Although the sampling points in the training and testing datasets could be different, we used the same group of sampling points for both training and testing datasets while keeping the sampling points for input and target functions different. 

For the numerical tests, we consider the cases $M=10, 50$ for different wave numbers $k=50, 100$ The training and test datasets are generated according to the parameters provided in Table \ref{test_2_table_1}. Both the conventional and multi-scale DeepOnets, with architectures specified in Tables \ref{test_2_table_2}–\ref{test_2_table_3} are evaluated. All test are run for 1500 epoches with a learning rate of $1 \times 10^{-4}$.


\begin{table}[htbp]
	\centering
	\begin{tabular}{|c|c|c|c|c|}
    \hline
		\multicolumn{2}{|c|}{Mode M / wave number k}&   $k=10$   &     $k=50$    &   $k=100$ \\
        \hline
		\multirow{2}{*}{$M=10$} & $(N_{train}, N_{test})$    &    (2000, 100)      &     (3000, 100)     &   (5000, 100)   \\
         \cline{2-5}
		&$(N_{sample}^{in}, N_{sample}^{out})$    &    (500, 500)     &   (500, 500)     &  (500, 1000)\\
        \hline
		\multirow{2}{*}{$M=50$} & $(N_{train}, N_{test})$    &    (2000, 100)      &     (3000, 100)     &   (5000, 100)   \\
         \cline{2-5}
		&$(N_{sample}^{in}, N_{sample}^{out})$    &    (3000, 500)     &   (3000, 500)     &  (3000, 1000)\\
         \hline
	\end{tabular}
	\caption{Number of functions and sampling points in training and testing datasets for different mode $M$ and wave number $k$. }
    \label{test_2_table_1}
\end{table}
\begin{table}[htbp]
	\centering
	\begin{tabular}{|c|c|c|c|c|}
    \hline
    $k$      & $S_{branch}$ & $S_{trunk}$   & architecture of the subnetworks    &    $\#$ parameters  \\
    \hline
         \multirow{3}{*}{$50$}  &   1   &   1  &  $\begin{array}{l}
        \mathcal N_{B}:[500,500,500,500,500,501]\\
        \mathcal N_{T}:[500,500,500,500,500,500]
        \end{array}$  & 19,852,300\\
        \cline{2-5}
		  &   1  &    10     &     $\begin{array}{l}
        \mathcal N_{B}:[500,500,500,500,500,501]\\
        \mathcal N_{T}:[50,50,50,50,50,50]
        \end{array}$     &  18,952,300\\
        \cline{2-5}
         &   5  &    10     &     $\begin{array}{l}
        \mathcal N_{B}:[500,500,500,500,500,501]\\
        \mathcal N_{T}:[50,50,50,50,50,50]
        \end{array}$     &  94,351,500\\
         \hline
         \multirow{3}{*}{$100$}  &   1   &   1  &  $\begin{array}{l}
        \mathcal N_{B}:[500,500,500,500,500,501]\\
        \mathcal N_{T}:[500,500,500,500,500,500]
        \end{array}$  & 19,852,300\\
        \cline{2-5}
		 &   1  &    10     &    $\begin{array}{l}
        \mathcal N_{B}:[500,500,500,500,500,501]\\
        \mathcal N_{T}:[50,50,50,50,50,50]
        \end{array}$   &   18,952,300\\
        \cline{2-5}
         &   5  &    10      &   $\begin{array}{l}
        \mathcal N_{B}:[500,500,500,500,500,501]\\
        \mathcal N_{T}:[50,50,50,50,50,50]
        \end{array}$     &  94,351,500\\
         \hline
	\end{tabular}
	\caption{Network architectures and number of parameters for the case $M=10$. }
    \label{test_2_table_2}
\end{table}
\begin{table}[htbp]
	\centering
	\begin{tabular}{|c|c|c|c|c|}
    \hline
    $k$      & $S_{branch}$ & $S_{trunk}$   & architecture of the subnetworks    &    $\#$ parameters  \\
    \hline
         \multirow{3}{*}{$50$}  &   1   &   1  &  $\begin{array}{l}
        \mathcal N_{B}:[3000,2000,1000,700,600,501]\\
        \mathcal N_{T}:[500,500,500,500,500,500]
        \end{array}$  & 19,852,300\\
        \cline{2-5}
		  &   1  &    10     &     $\begin{array}{l}
        \mathcal N_{B}:[3000,2000,1000,700,600,501]\\
        \mathcal N_{T}:[50,50,50,50,50,50]
        \end{array}$     &  18,952,300\\
        \cline{2-5}
         &   5  &    10     &     $\begin{array}{l}
        \mathcal N_{B}:[3000,2000,1000,700,600,501]\\
        \mathcal N_{T}:[50,50,50,50,50,50]
        \end{array}$     &  94,351,500\\
         \hline
         \multirow{3}{*}{$100$}  &   1   &   1  &  $\begin{array}{l}
        \mathcal N_{B}:[3000,2000,1000,700,600,501]\\
        \mathcal N_{T}:[500,500,500,500,500,500]
        \end{array}$  & 19,852,300\\
        \cline{2-5}
		 &   1  &    10     &    $\begin{array}{l}
        \mathcal N_{B}:[3000,2000,1000,700,600,501]\\
        \mathcal N_{T}:[50,50,50,50,50,50]
        \end{array}$   &   18,952,300\\
        \cline{2-5}
         &   5  &    10      &   $\begin{array}{l}
        \mathcal N_{B}:[3000,2000,1000,700,600,501]\\
        \mathcal N_{T}:[50,50,50,50,50,50]
        \end{array}$     &  94,351,500\\
         \hline
	\end{tabular}
	\caption{Network architectures and number of parameters for the case $M=50$. }
    \label{test_2_table_3}
\end{table}

\begin{figure}[ht!]
\center
\subfigure[ $M=10, k=100$]{\includegraphics[scale=0.25]{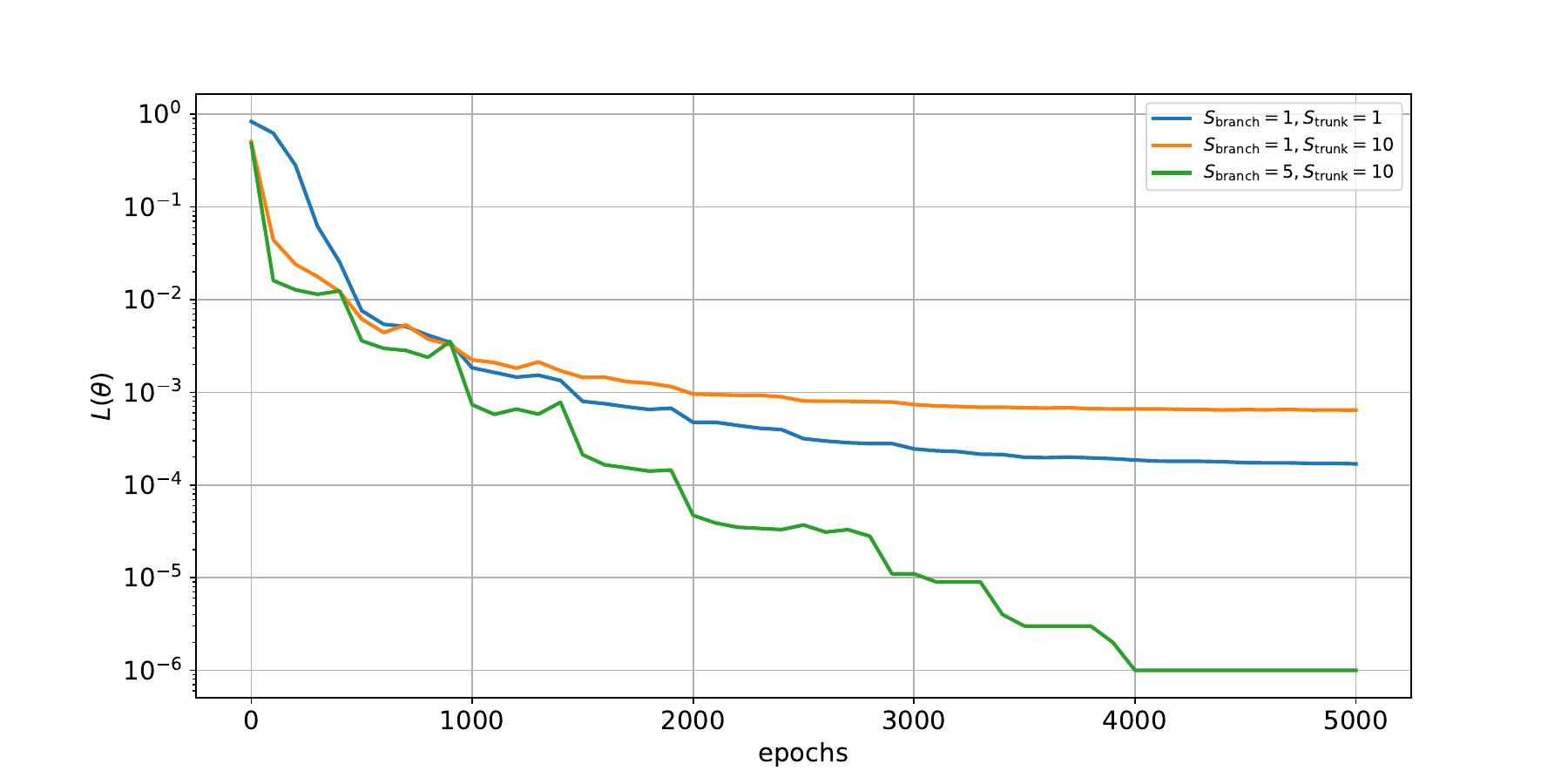}}
\subfigure[ $M=10, k=100$]{\includegraphics[scale=0.25]{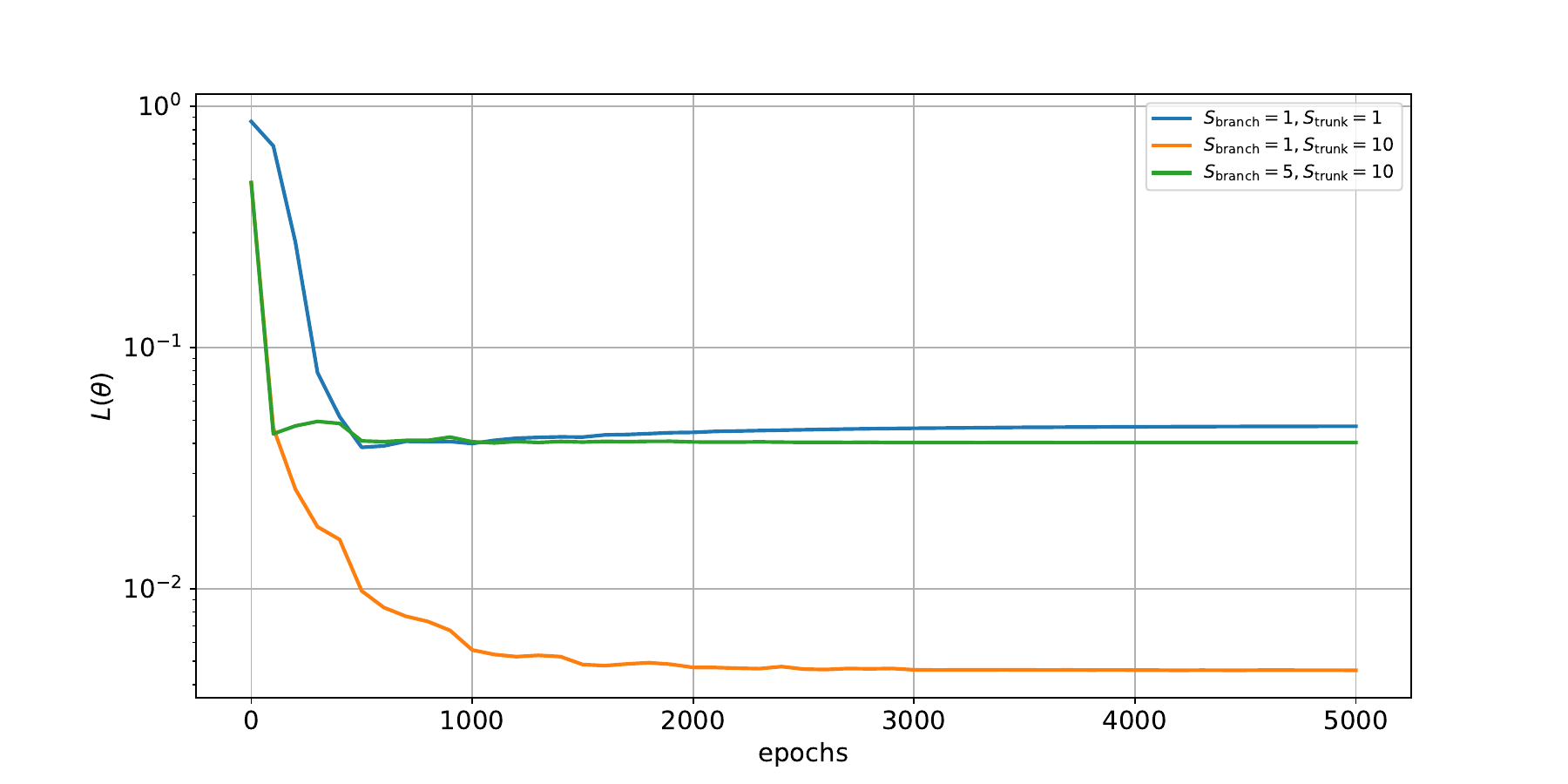}}\\
\subfigure[ $M=50, k=50$]{\includegraphics[scale=0.25]{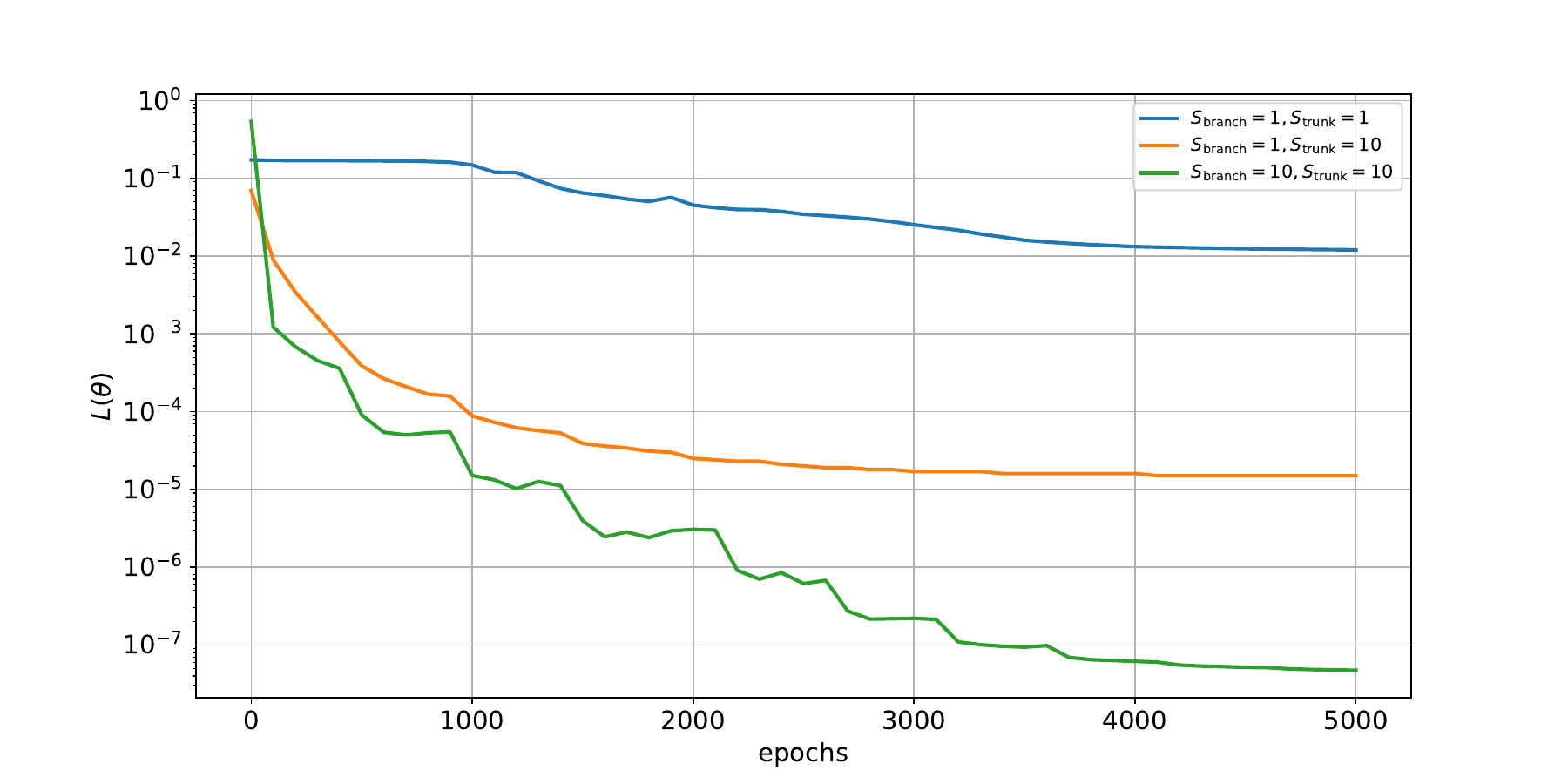}}
\subfigure[$M=50, k=50$]{\includegraphics[scale=0.25]{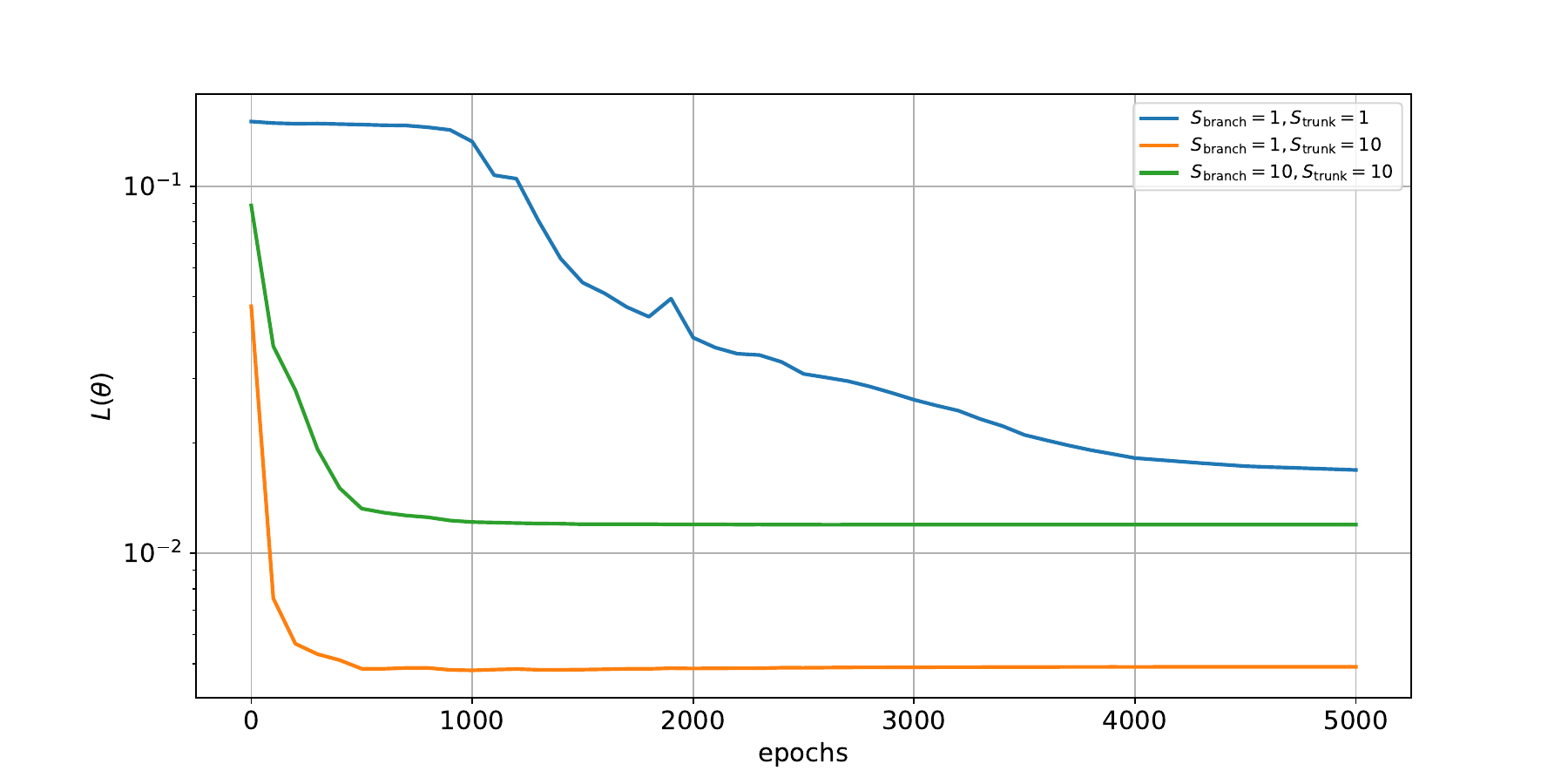}}\\
\subfigure[$M=50, k=100$]{\includegraphics[scale=0.25]{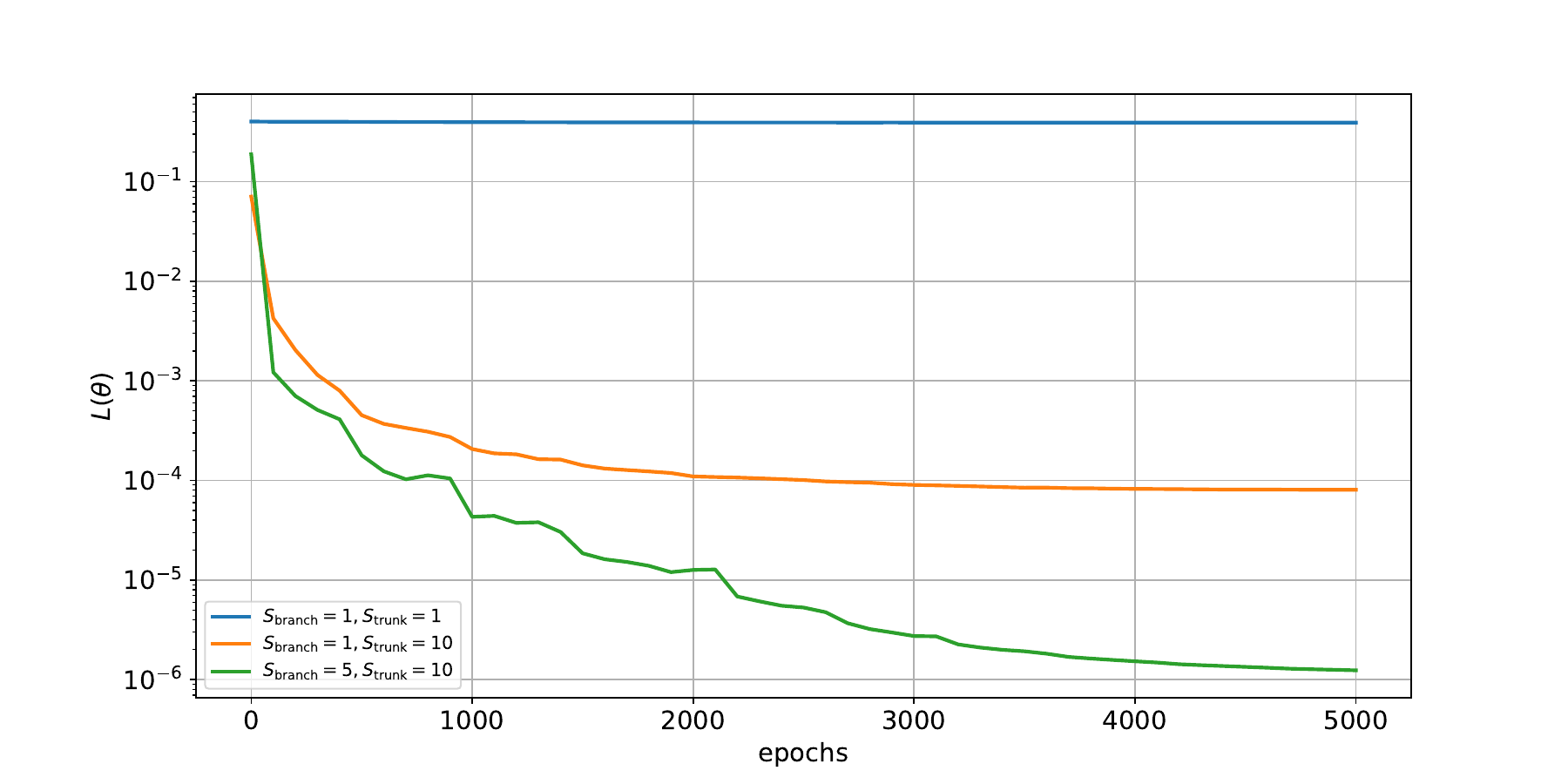}}
\subfigure[$M=50, k=100$]{\includegraphics[scale=0.25]{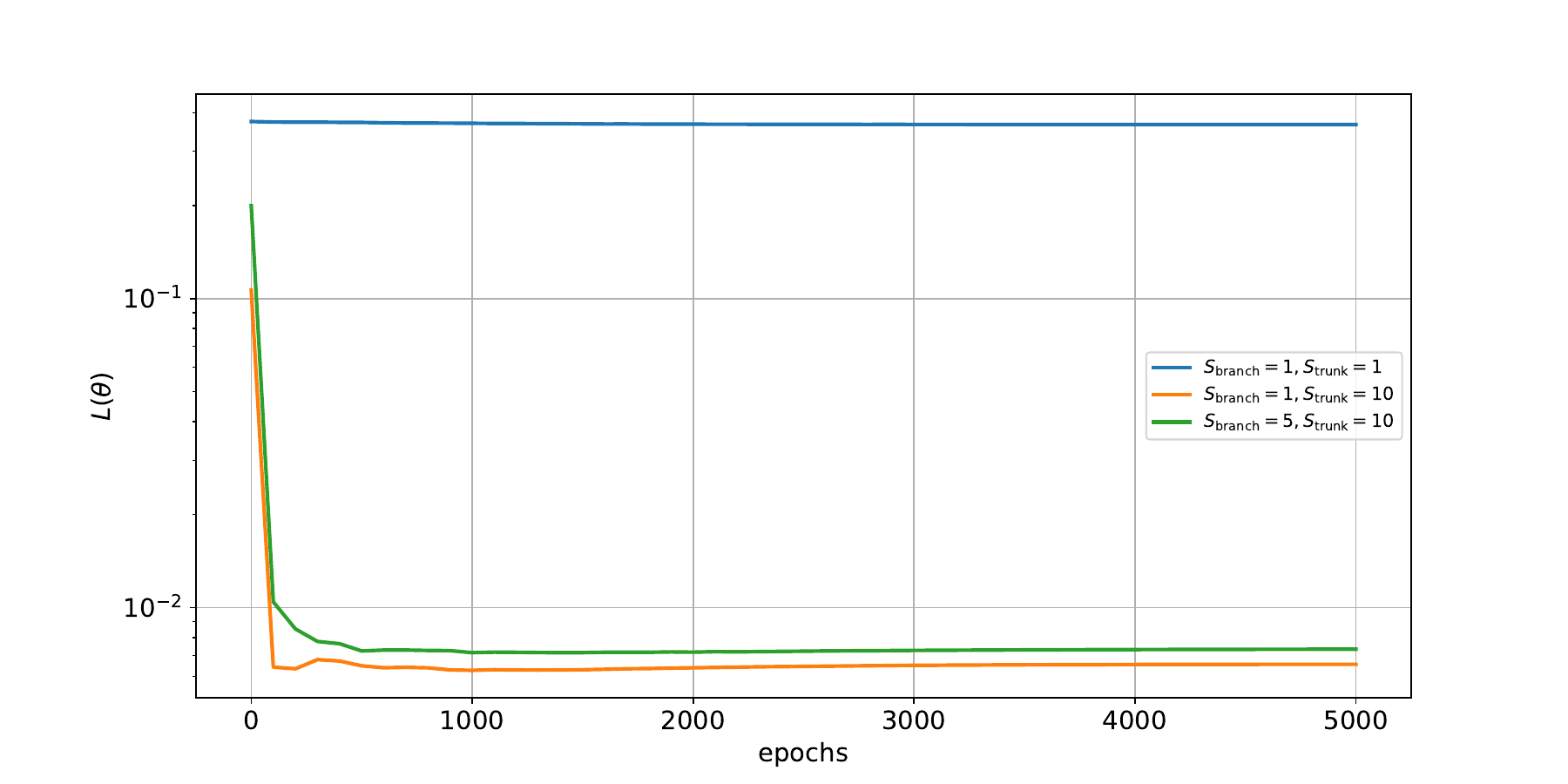}}
\caption{Loss on training (left) and testing (right) data for the various cases.}%
\label{test_1_fig_1}%
\end{figure}

\begin{figure}[ht!]
\center
\subfigure[real part on training data]{\includegraphics[scale=0.25]{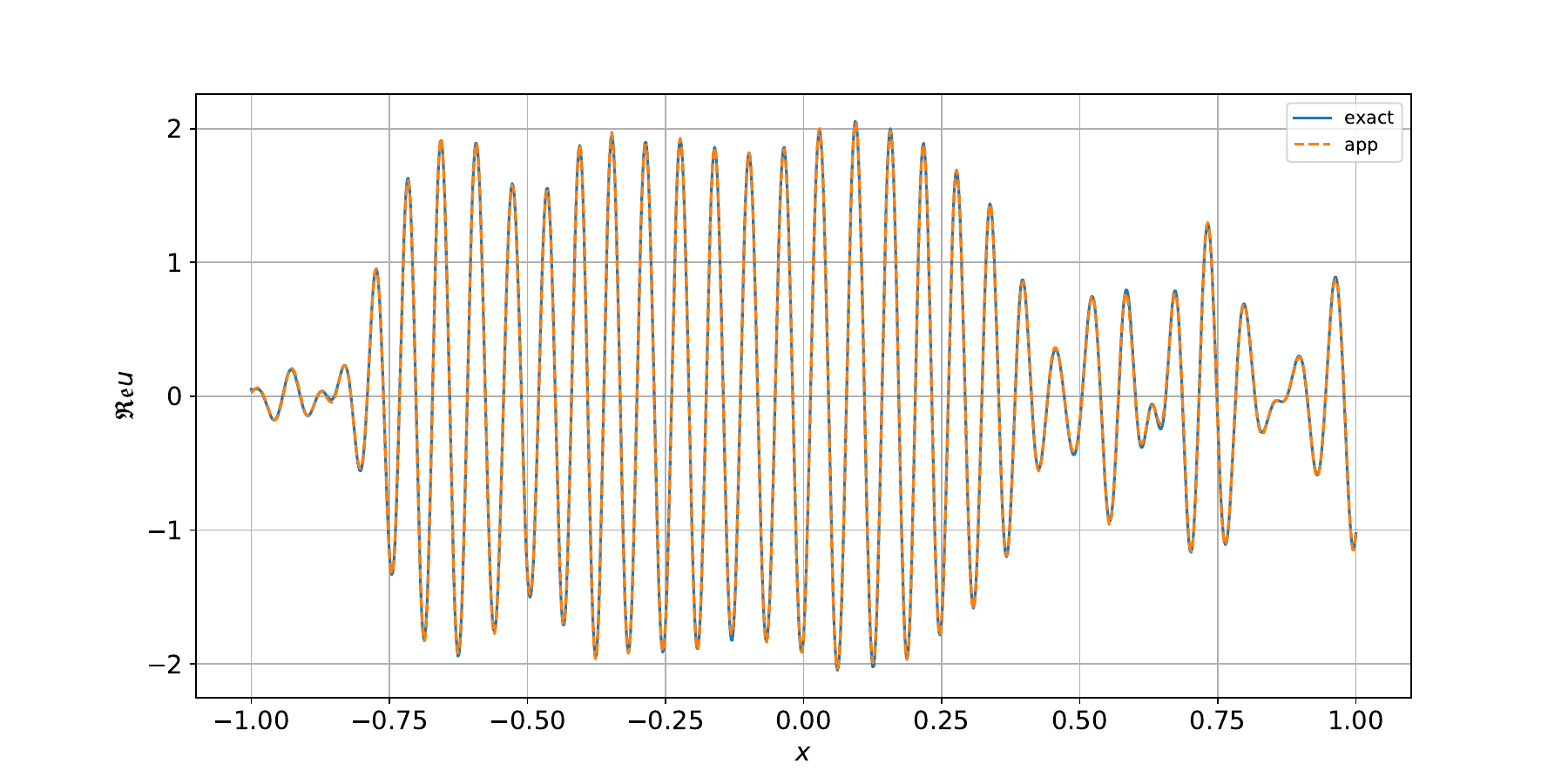}}
\subfigure[real part on test data]{\includegraphics[scale=0.25]{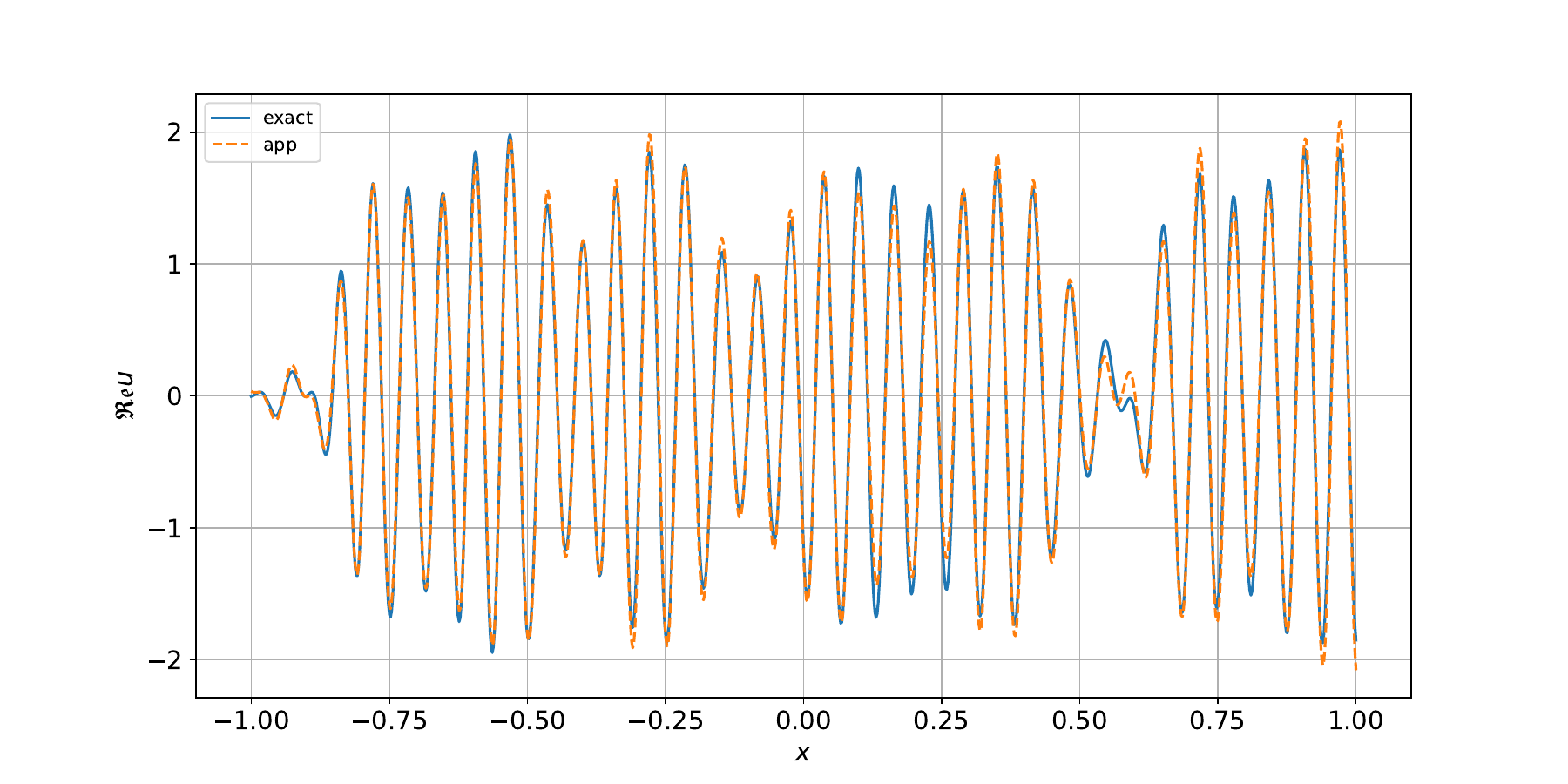}}
\subfigure[imaginary part on training data]{\includegraphics[scale=0.25]{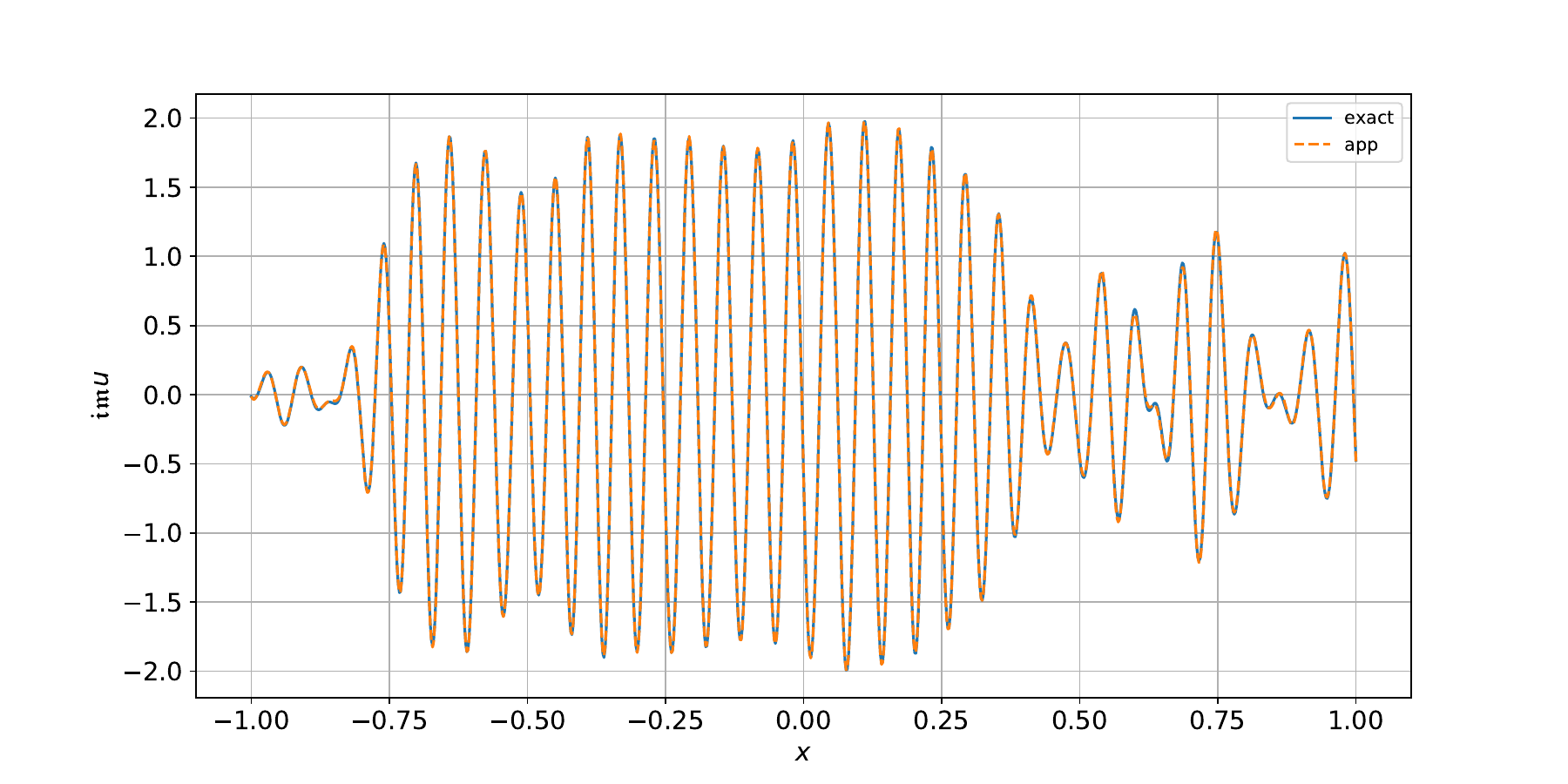}}
\subfigure[imaginary part on test data]{\includegraphics[scale=0.25]{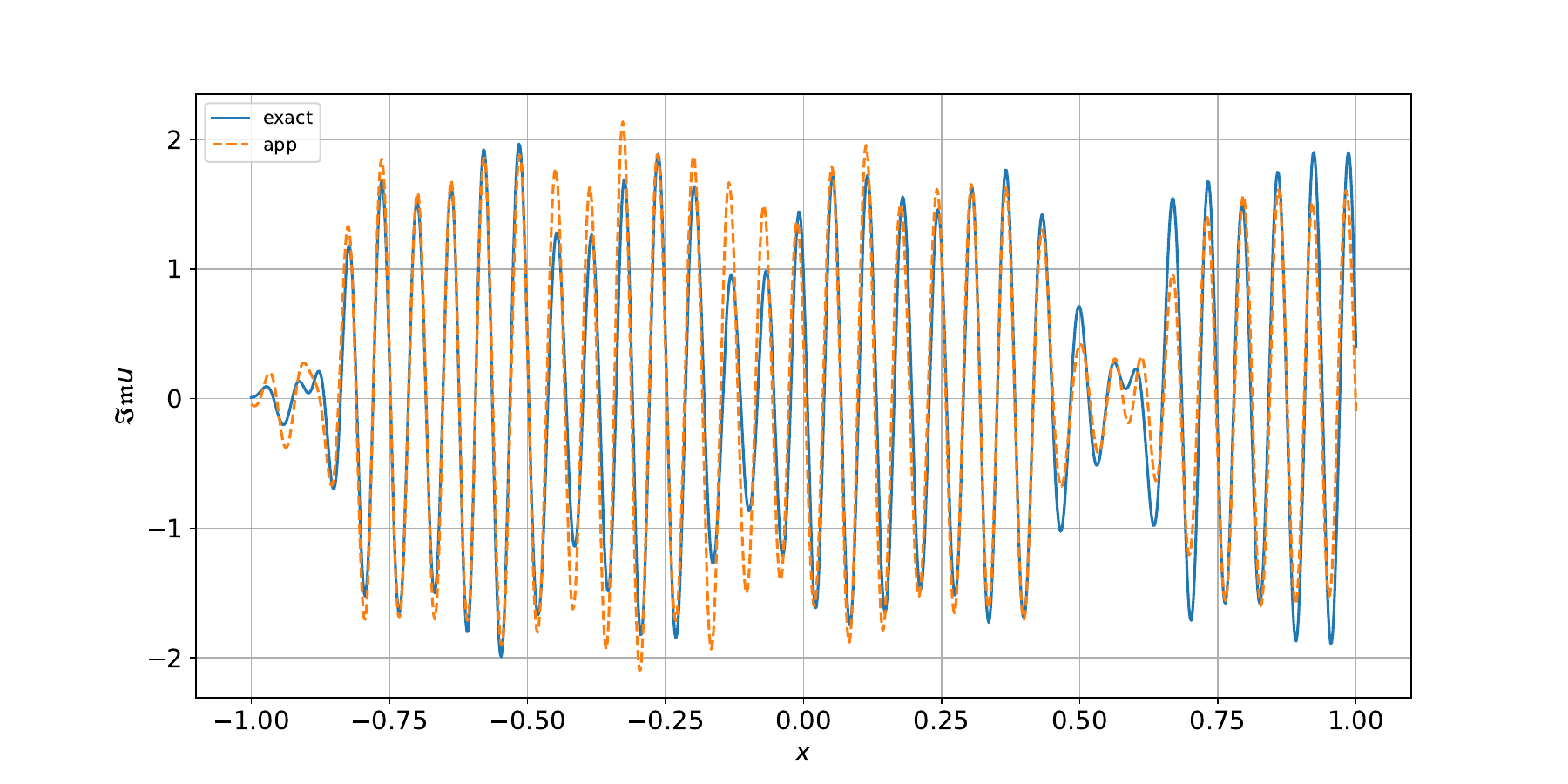}}
\caption{Training results using DeepONet ($S_{\rm branch}=1, S_{\rm trunk}=1$) for $M=10, k=100$.}%
\label{test_1_fig_2}%
\end{figure}

\begin{figure}[ht!]
\center
\subfigure[real part on training data]{\includegraphics[scale=0.25]{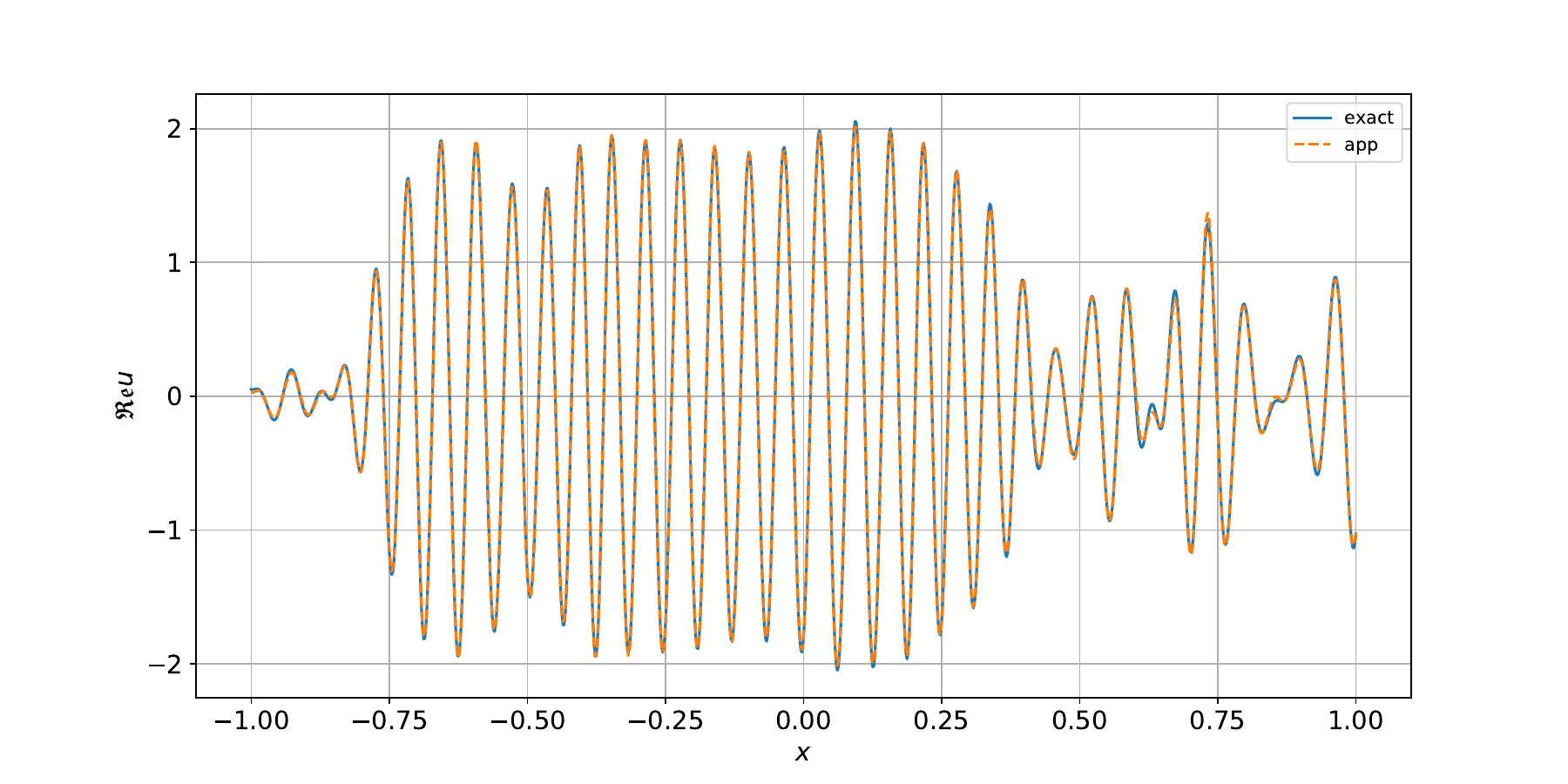}}
\subfigure[real part on test data]{\includegraphics[scale=0.25]{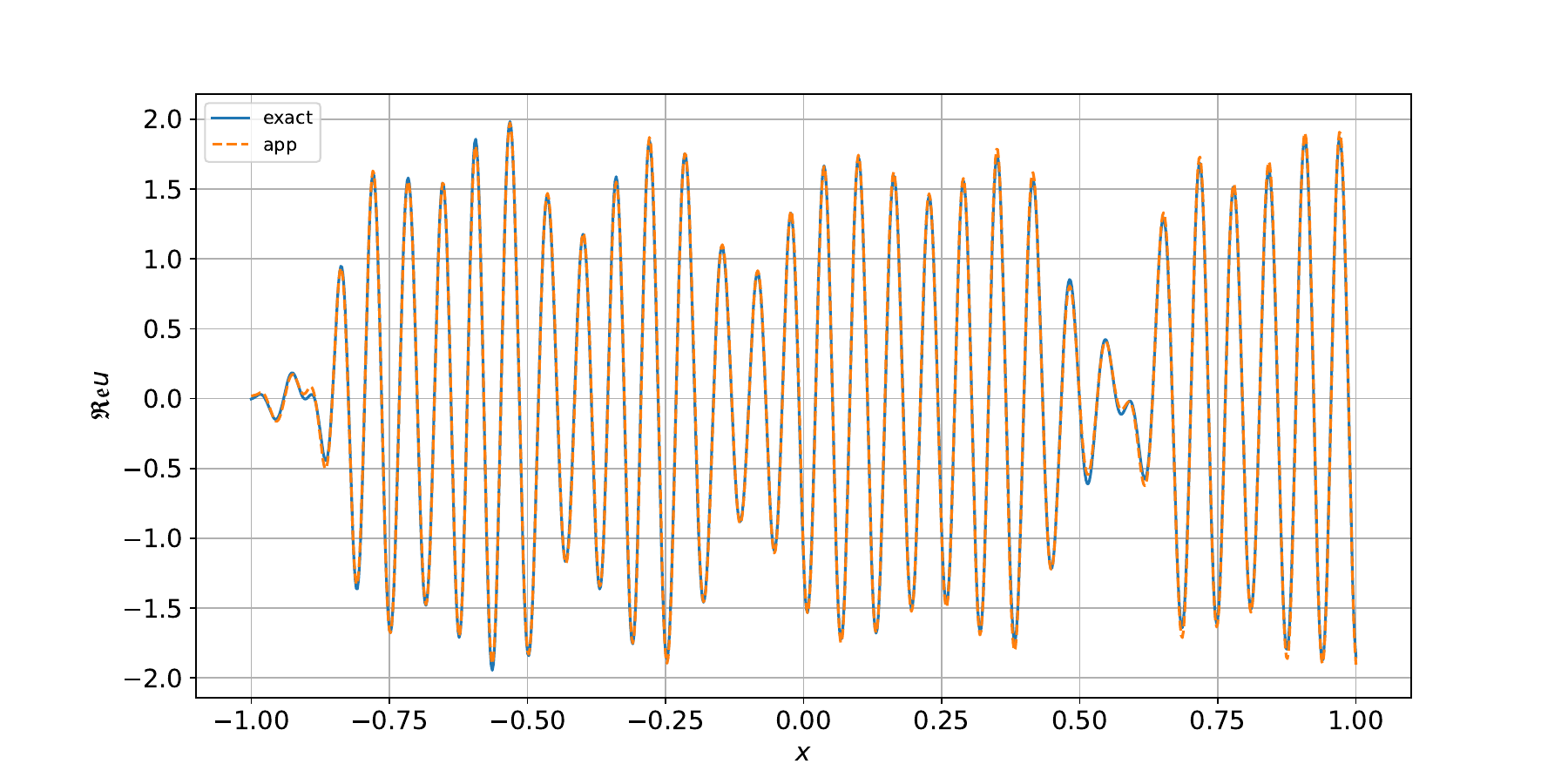}}
\subfigure[imaginary part on training data]{\includegraphics[scale=0.25]{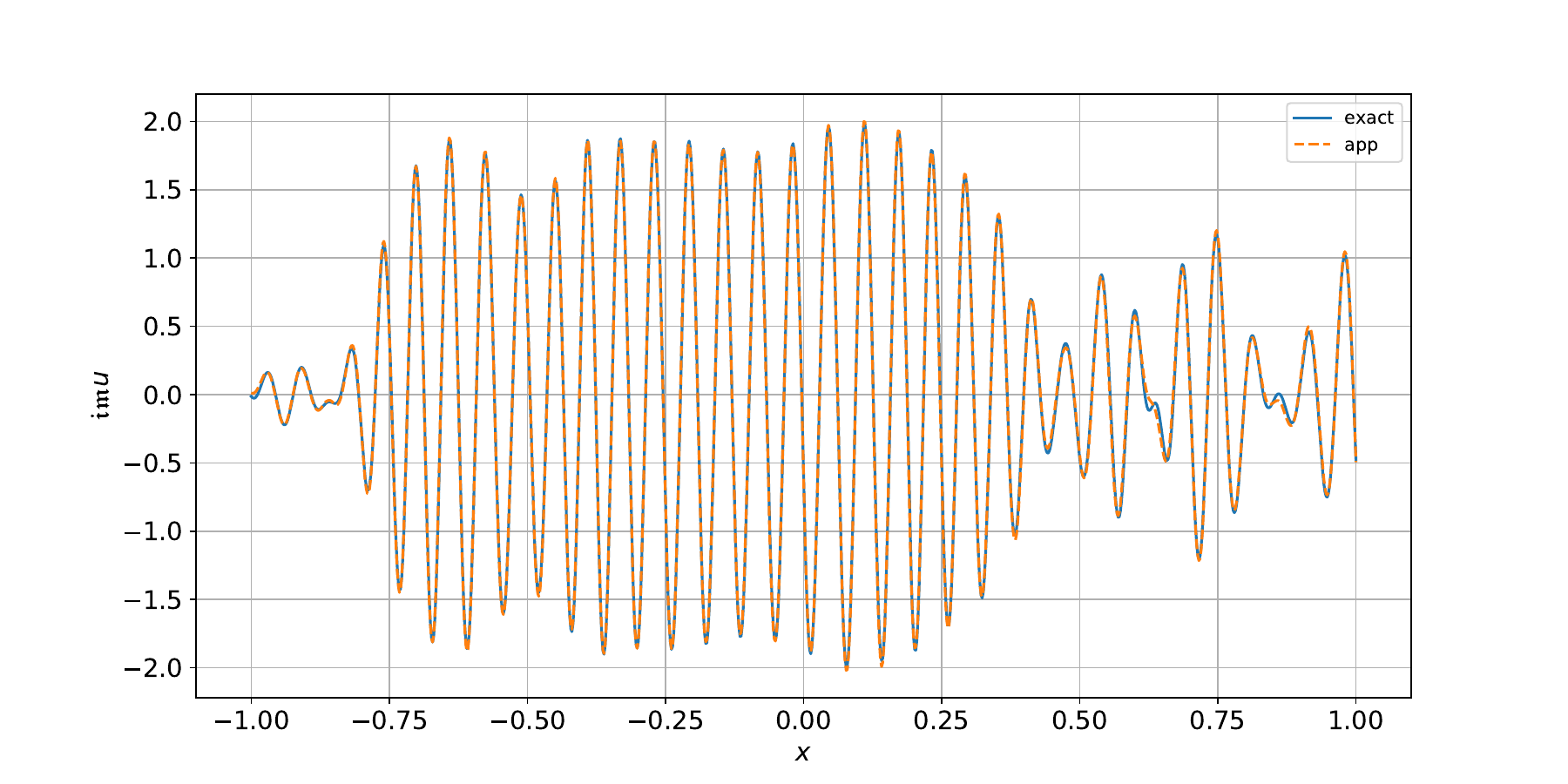}}
\subfigure[imaginary part on test data]{\includegraphics[scale=0.25]{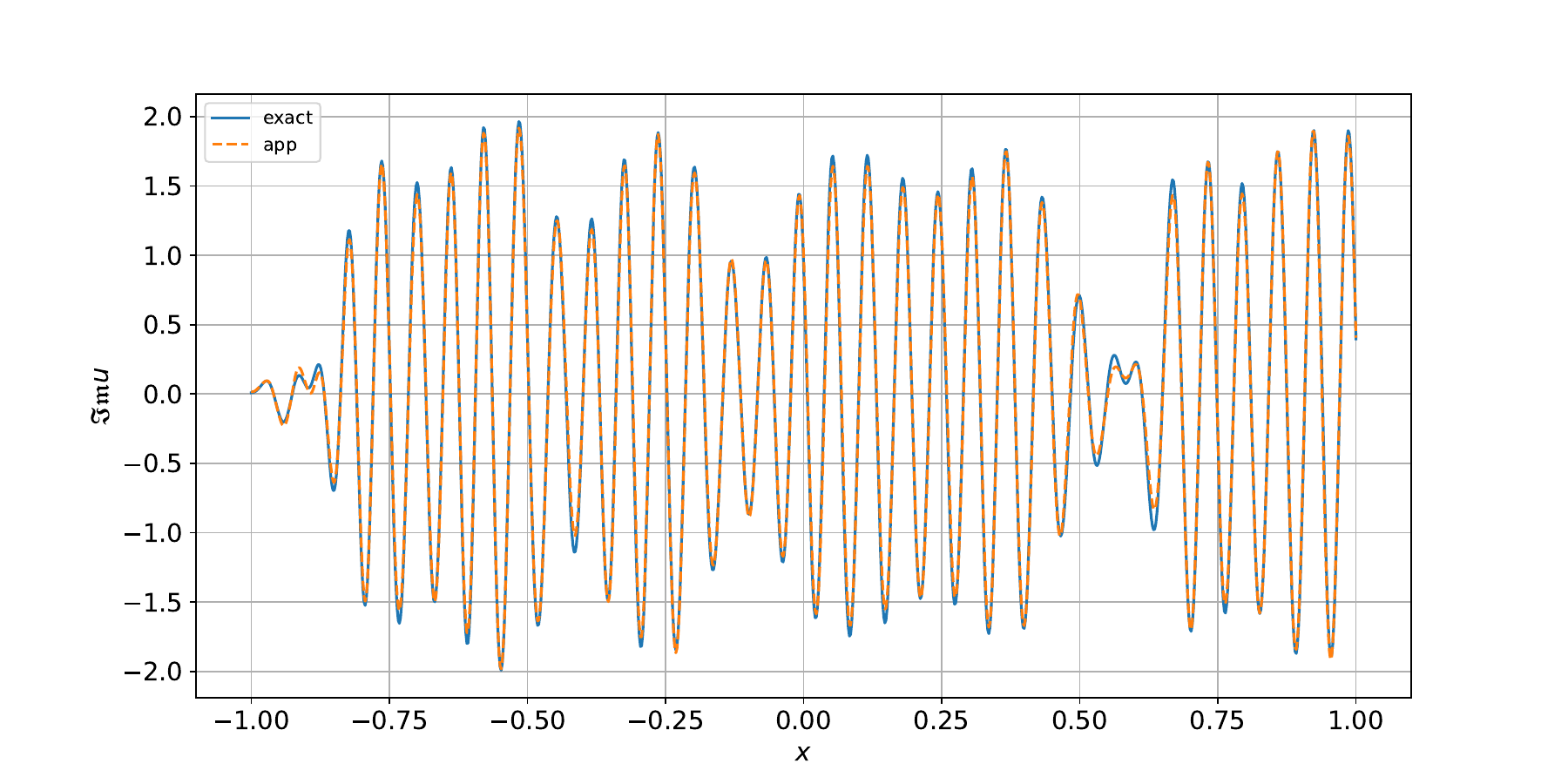}}
\caption{Training results using multi-scale DeepONet ($S_{\rm branch}=1, S_{\rm trunk}=10$) for the case $M=10, k=100$.}%
\label{test_1_fig_3}%
\end{figure}

\begin{figure}[ht!]
\center
\subfigure[real part on training data]{\includegraphics[scale=0.25]{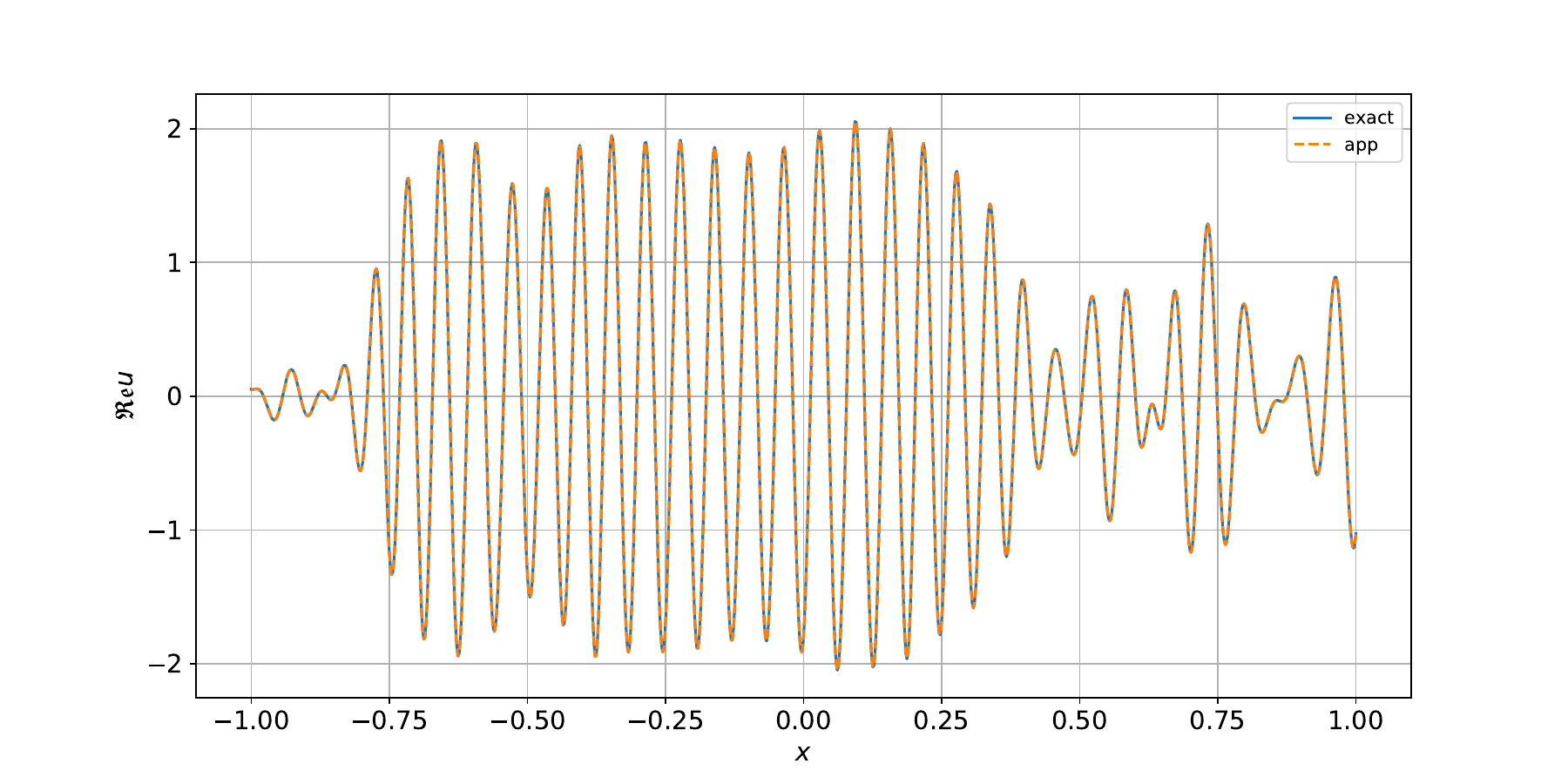}}
\subfigure[real part on test data]{\includegraphics[scale=0.25]{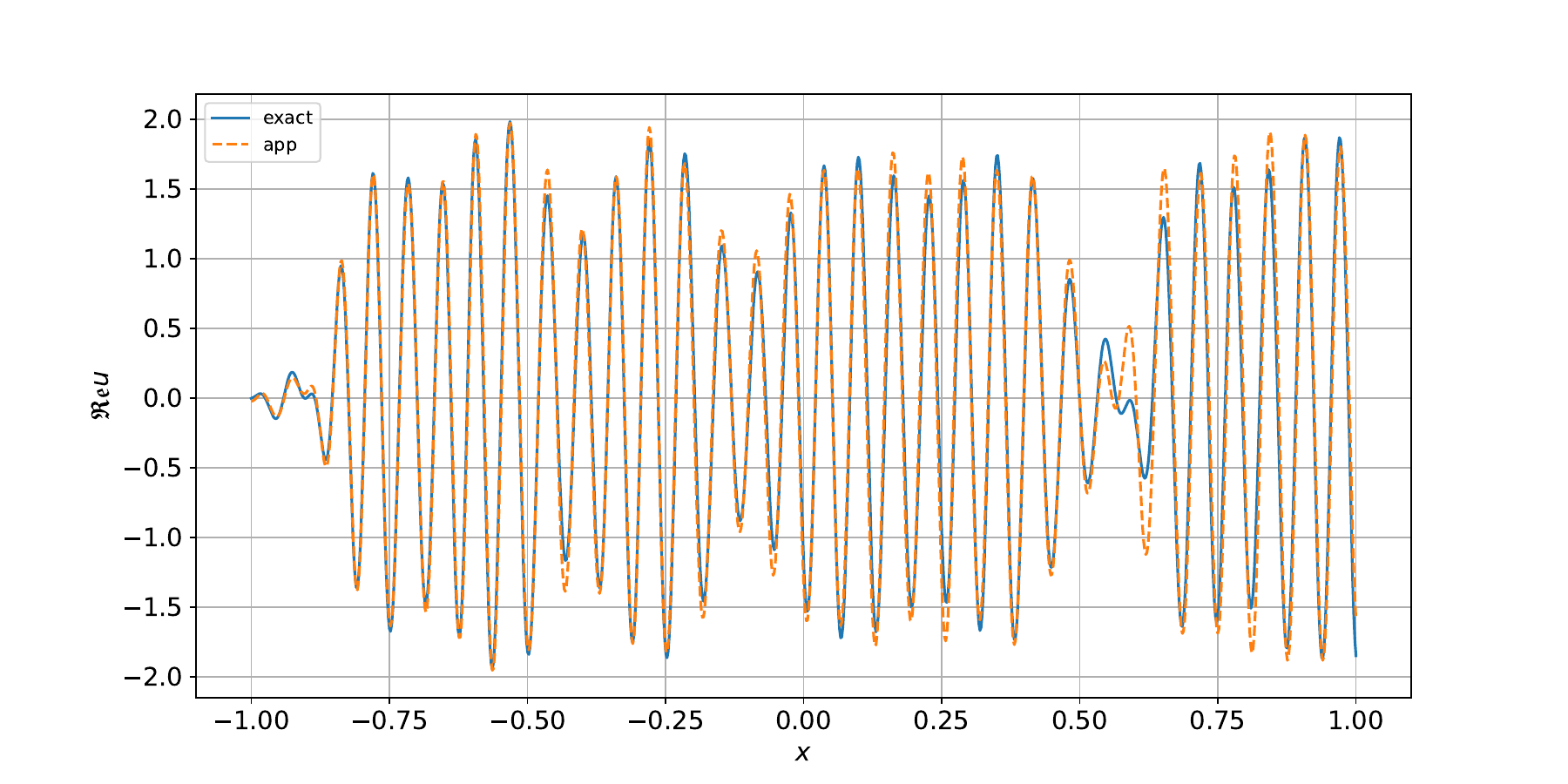}}
\subfigure[imaginary part on training data]{\includegraphics[scale=0.25]{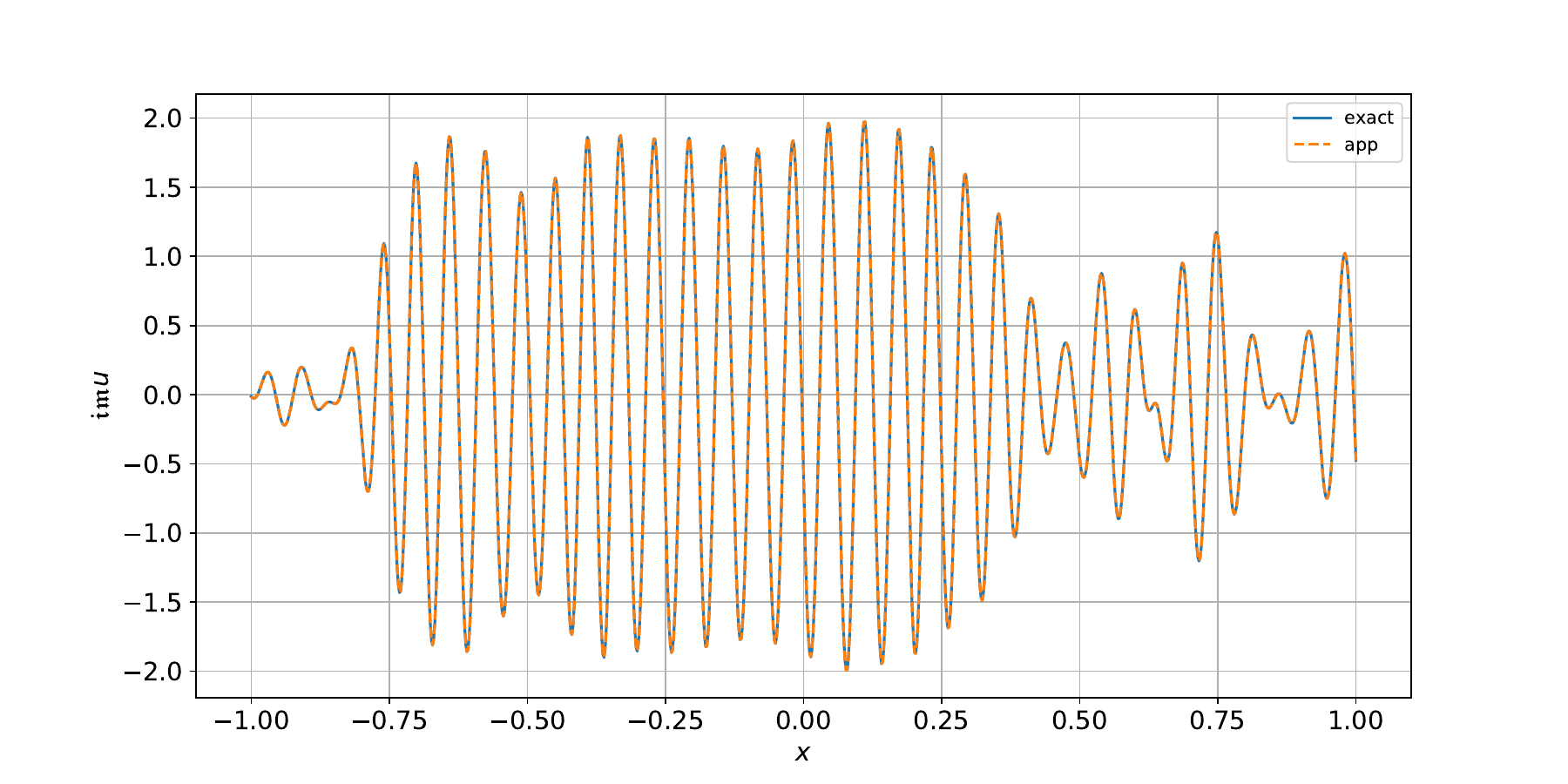}}
\subfigure[imaginary part on test data]{\includegraphics[scale=0.25]{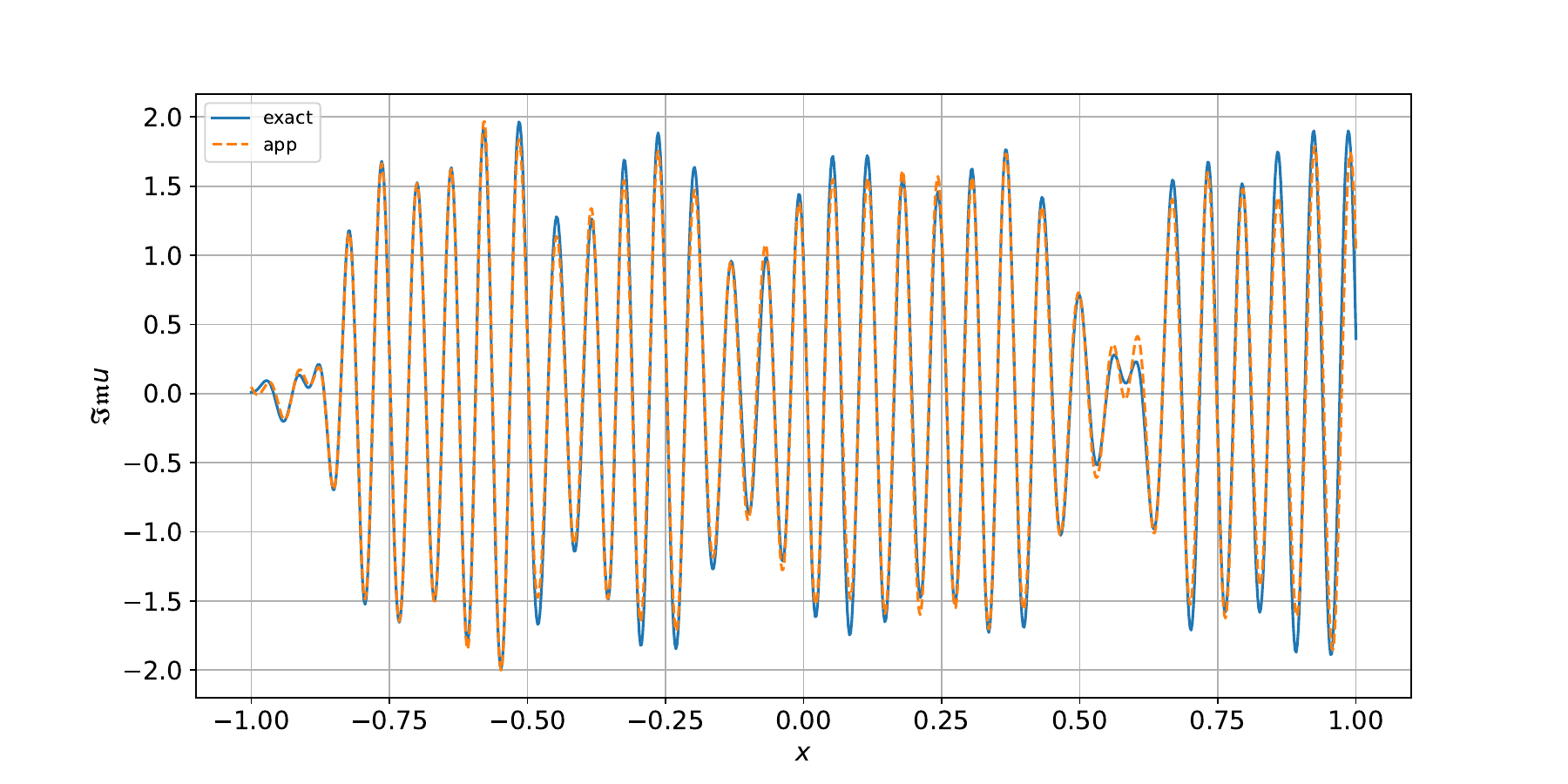}}
\caption{Training results using multi-scale DeepONet ($S_{\rm branch}=5, S_{\rm trunk}=10$) for the case $M=10, k=100$.}%
\label{test_1_fig_4}%
\end{figure}

\begin{figure}[ht!]
\center
\subfigure[real part on training data]{\includegraphics[scale=0.25]{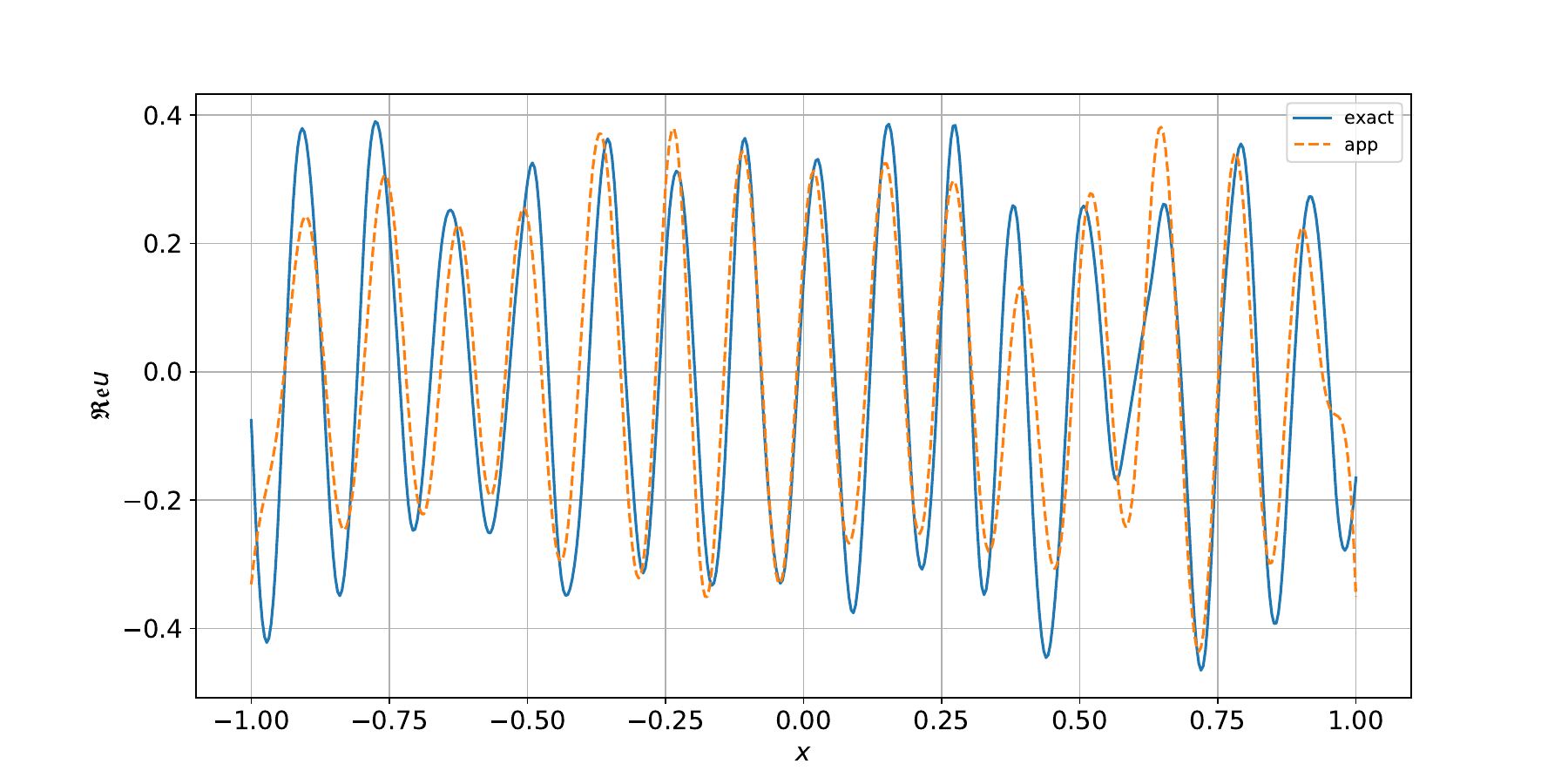}}
\subfigure[real part on test data]{\includegraphics[scale=0.25]{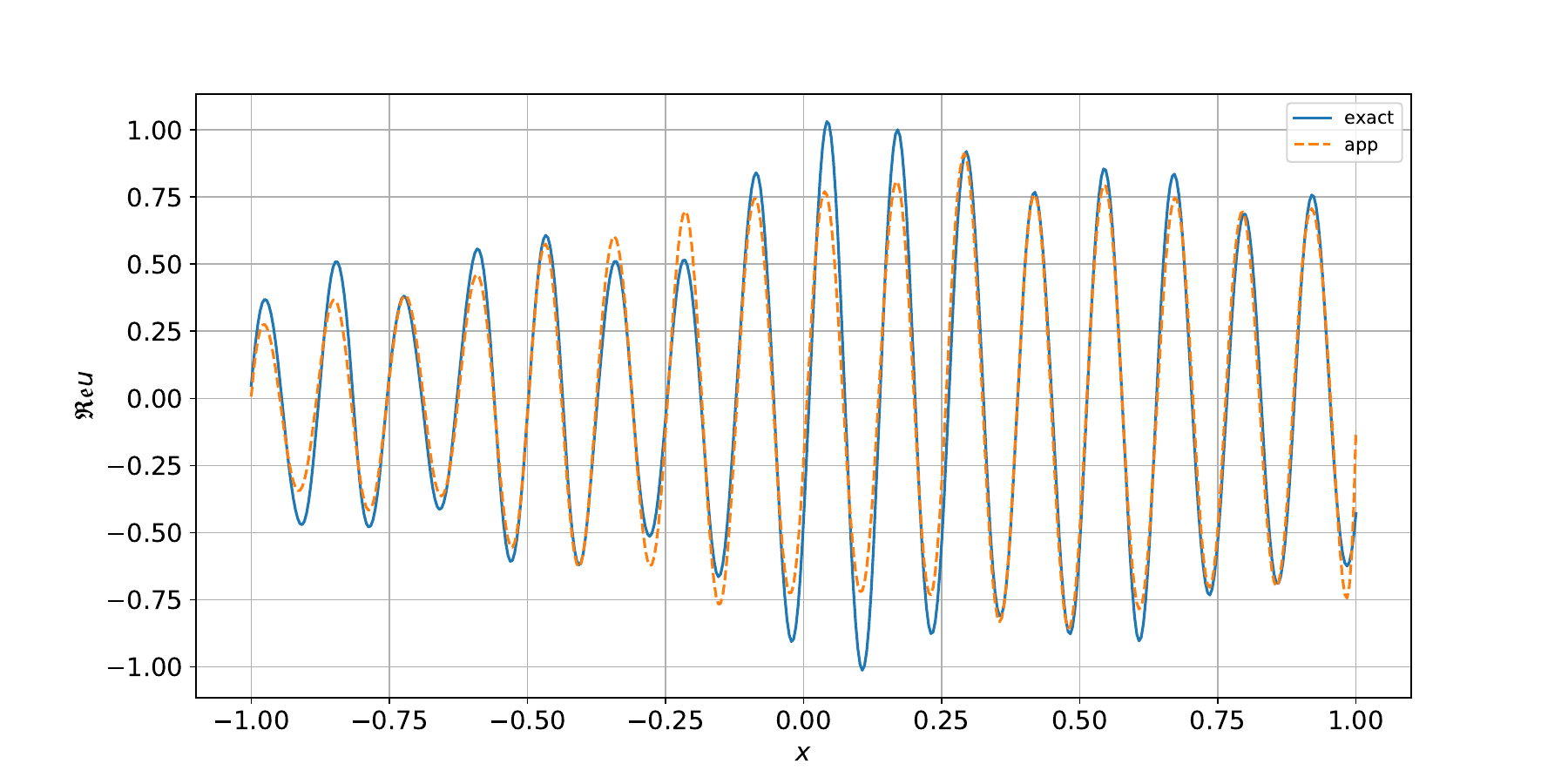}}
\subfigure[imaginary part on training data]{\includegraphics[scale=0.25]{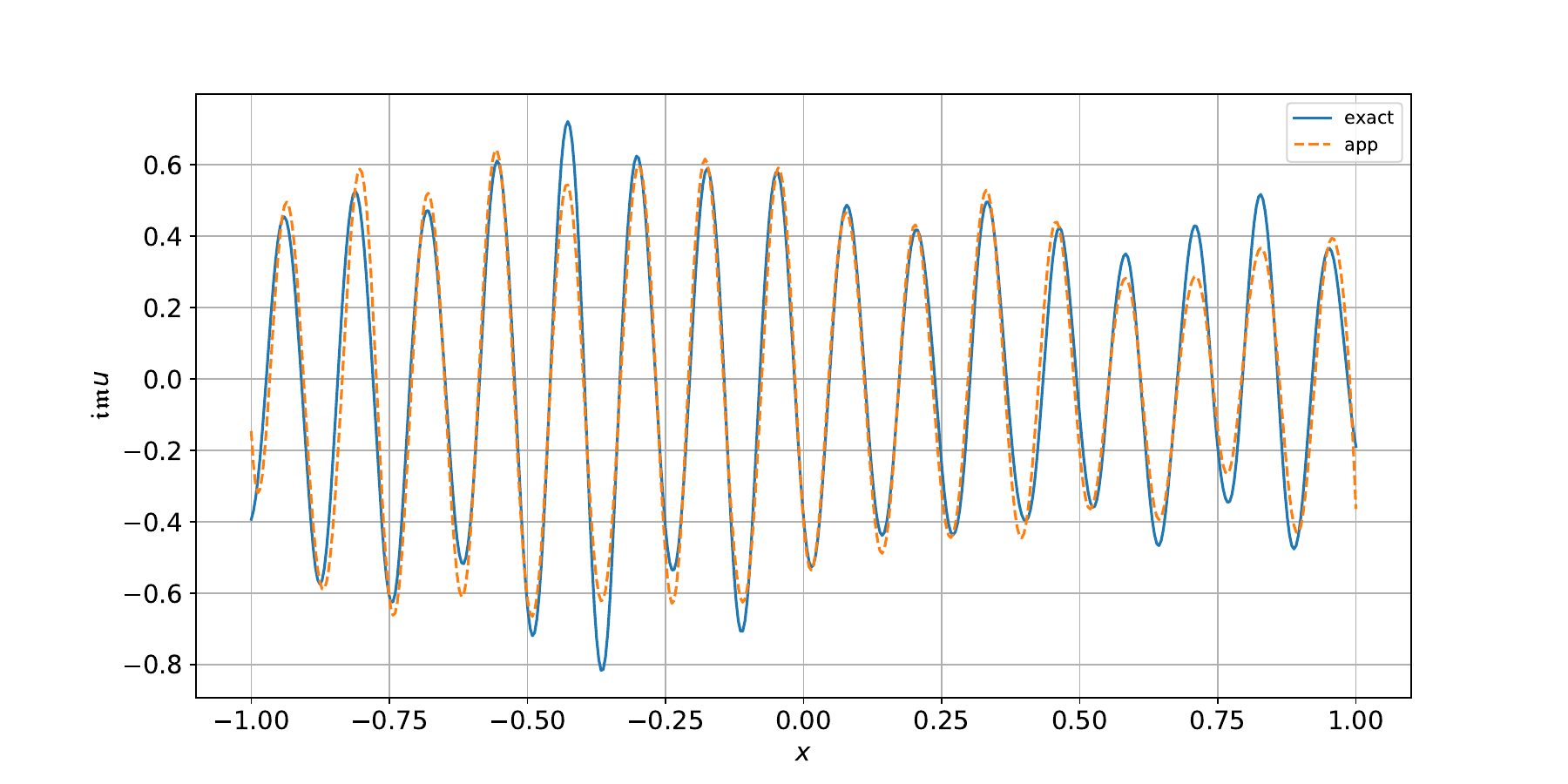}}
\subfigure[imaginary part on test data]{\includegraphics[scale=0.25]{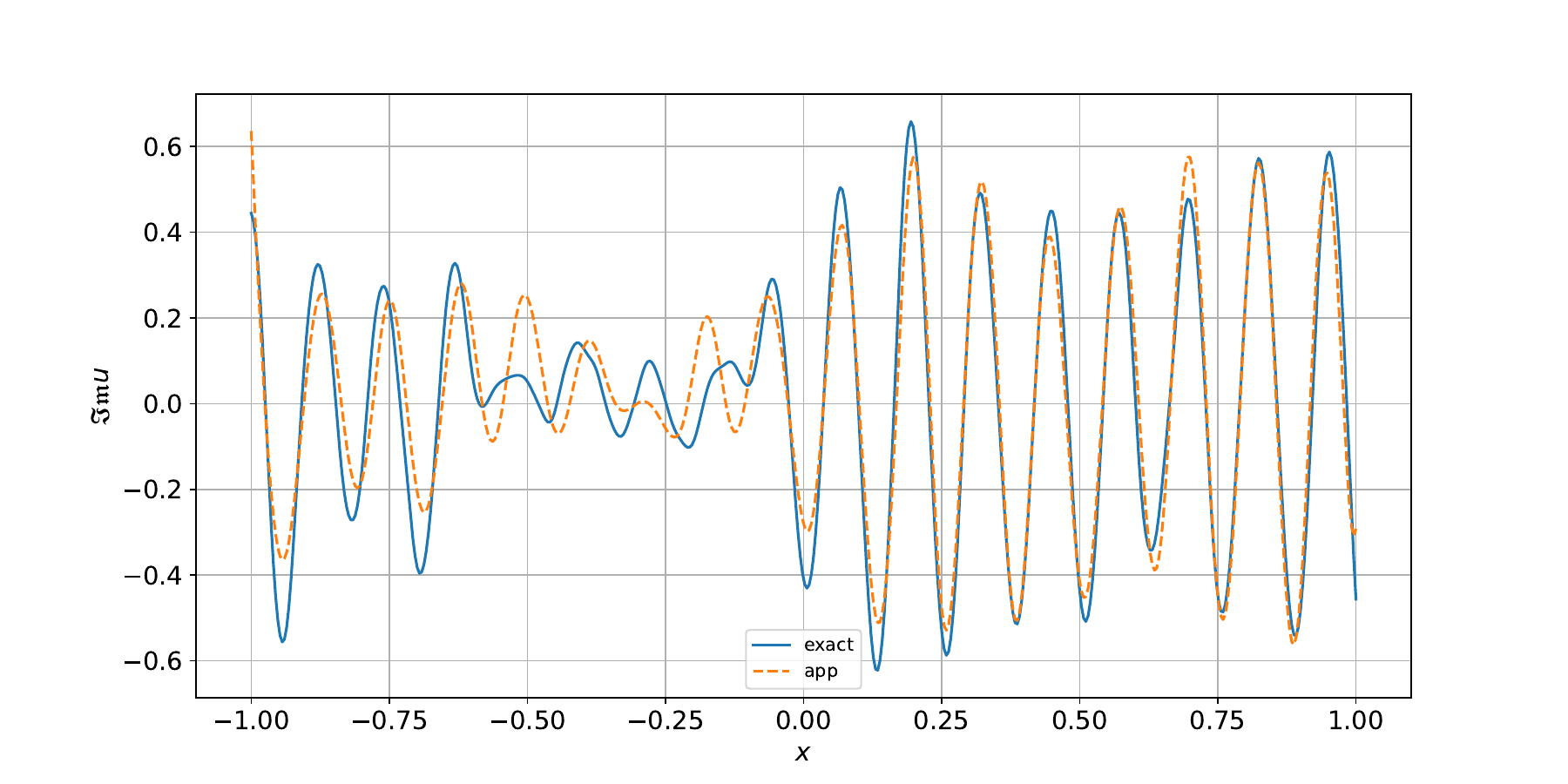}}
\caption{Training results using DeepONet ($S_{\rm branch}=1, S_{\rm trunk}=1$) for $M=50, k=50$.}%
\label{test_2_fig_2}%
\end{figure}

\begin{figure}[ht!]
\center
\subfigure[real part on training data]{\includegraphics[scale=0.25]{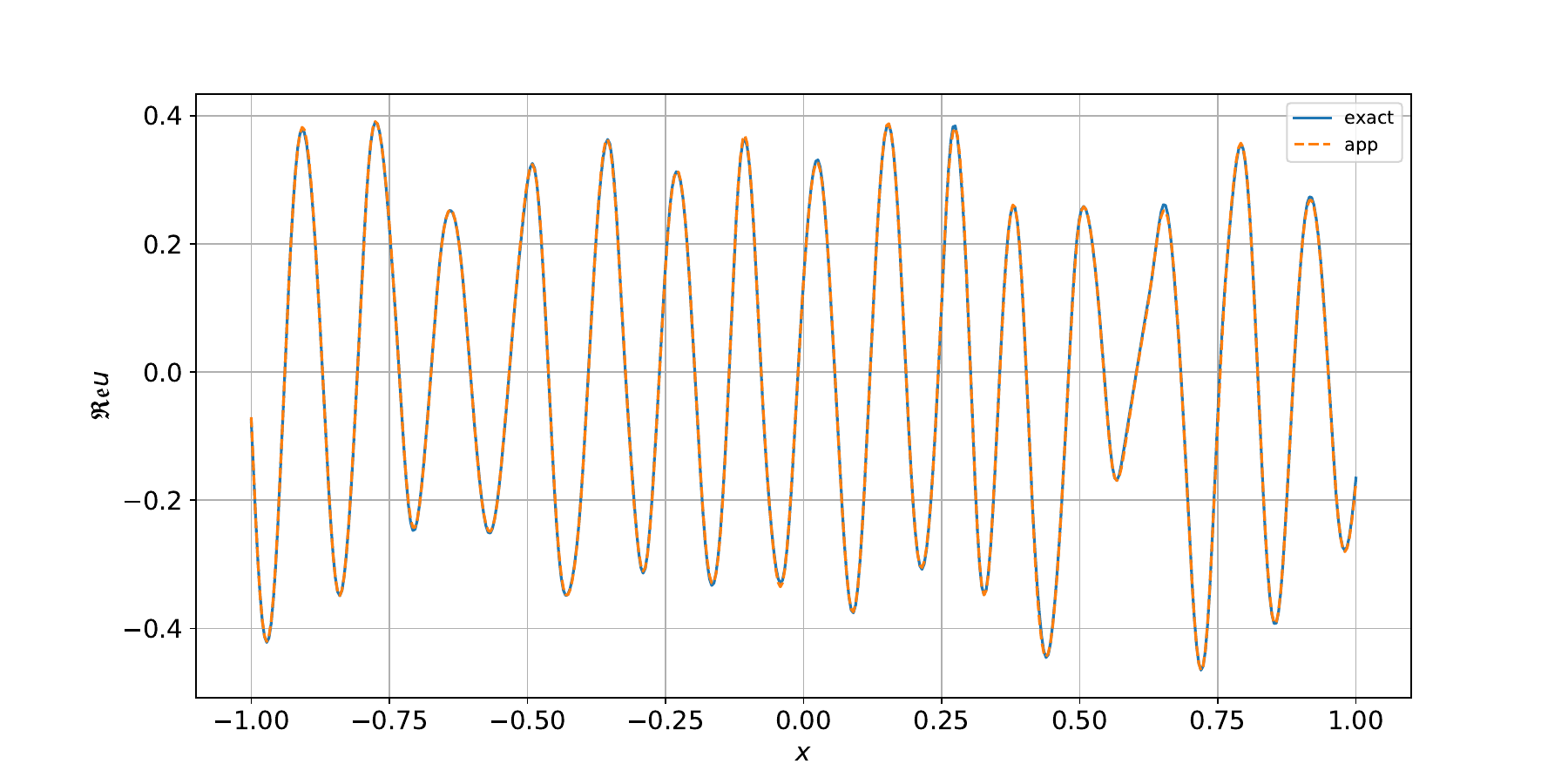}}
\subfigure[real part on test data]{\includegraphics[scale=0.25]{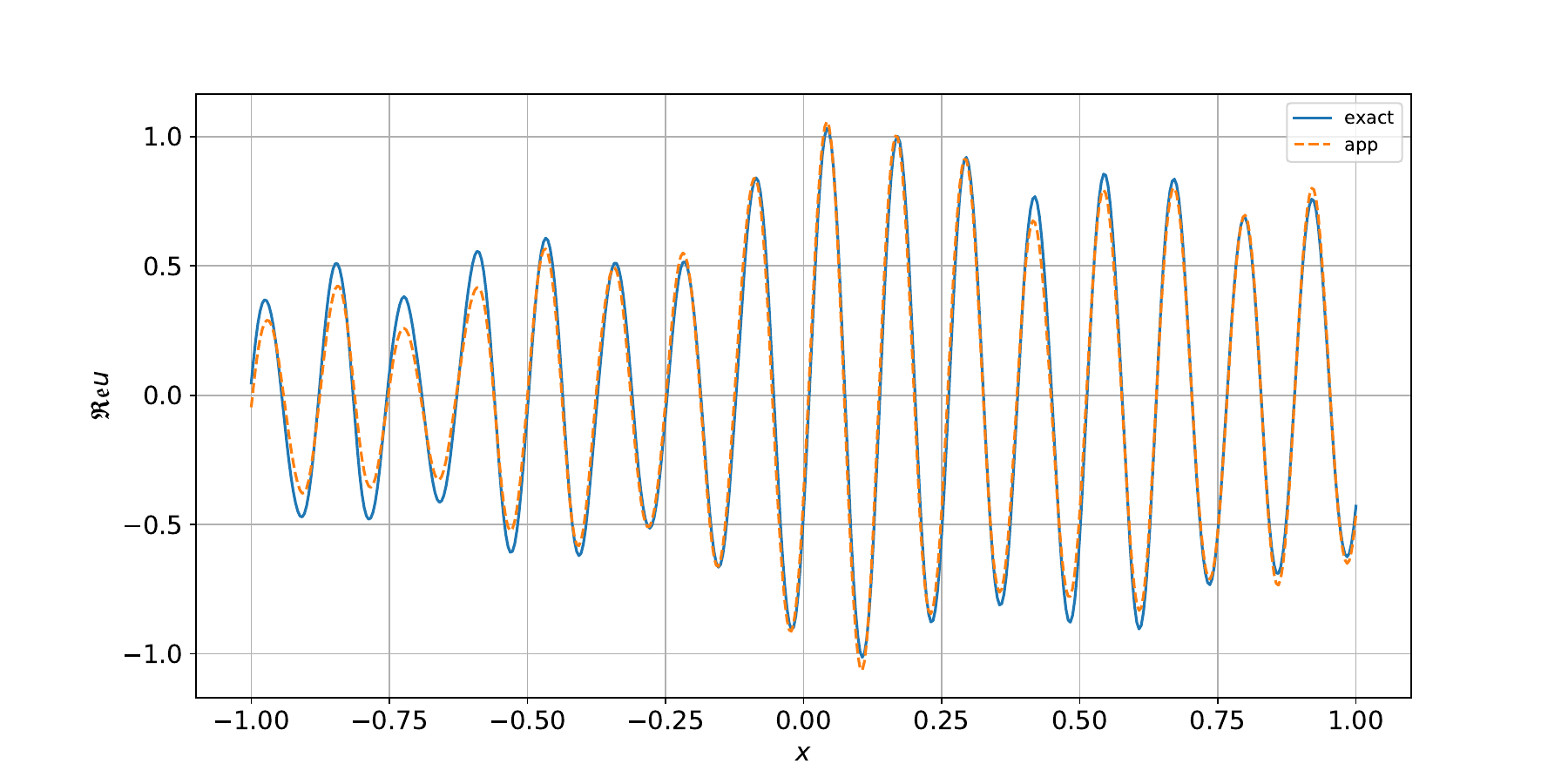}}
\subfigure[imaginary part on training data]{\includegraphics[scale=0.25]{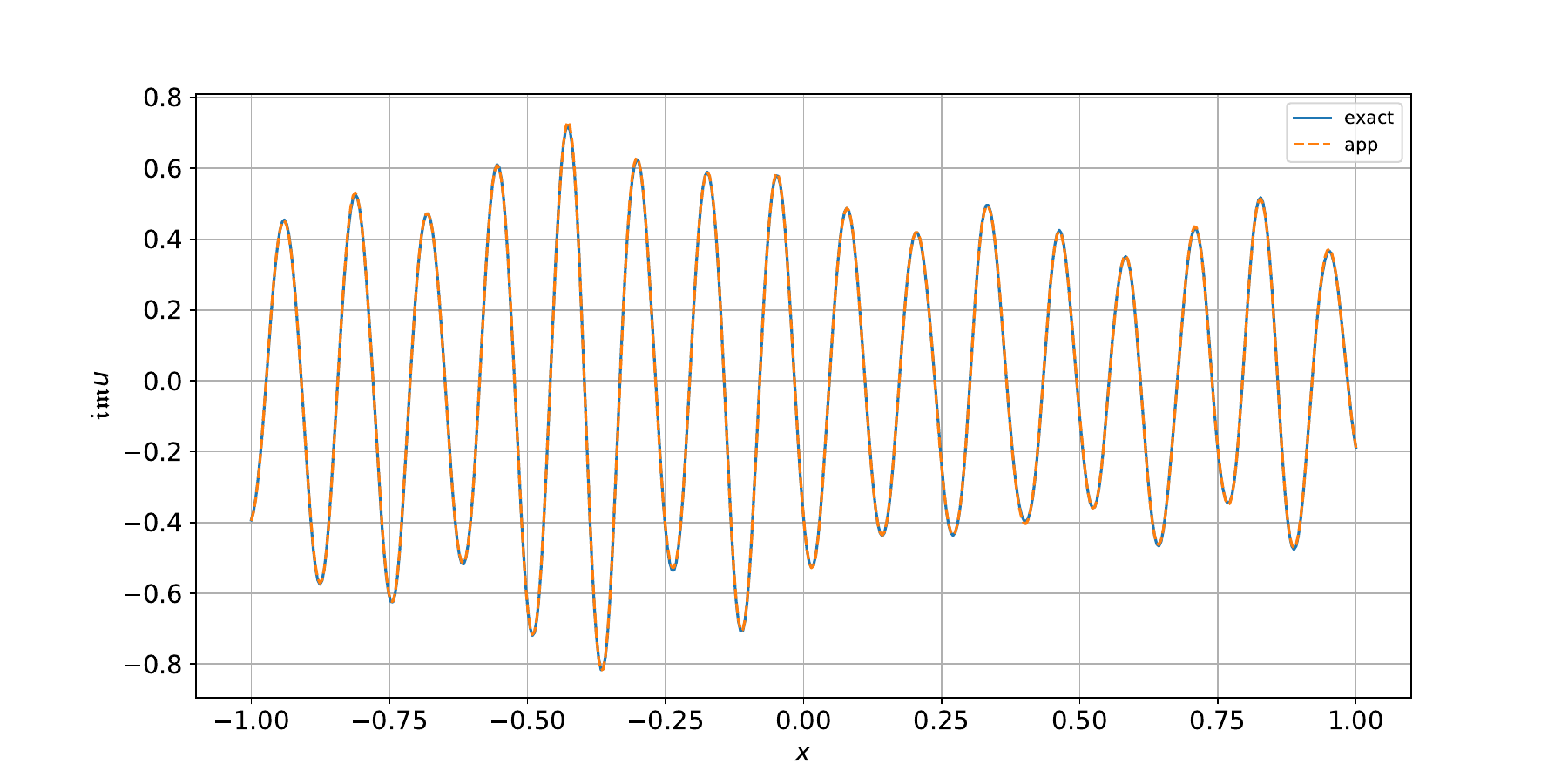}}
\subfigure[imaginary part on test data]{\includegraphics[scale=0.25]{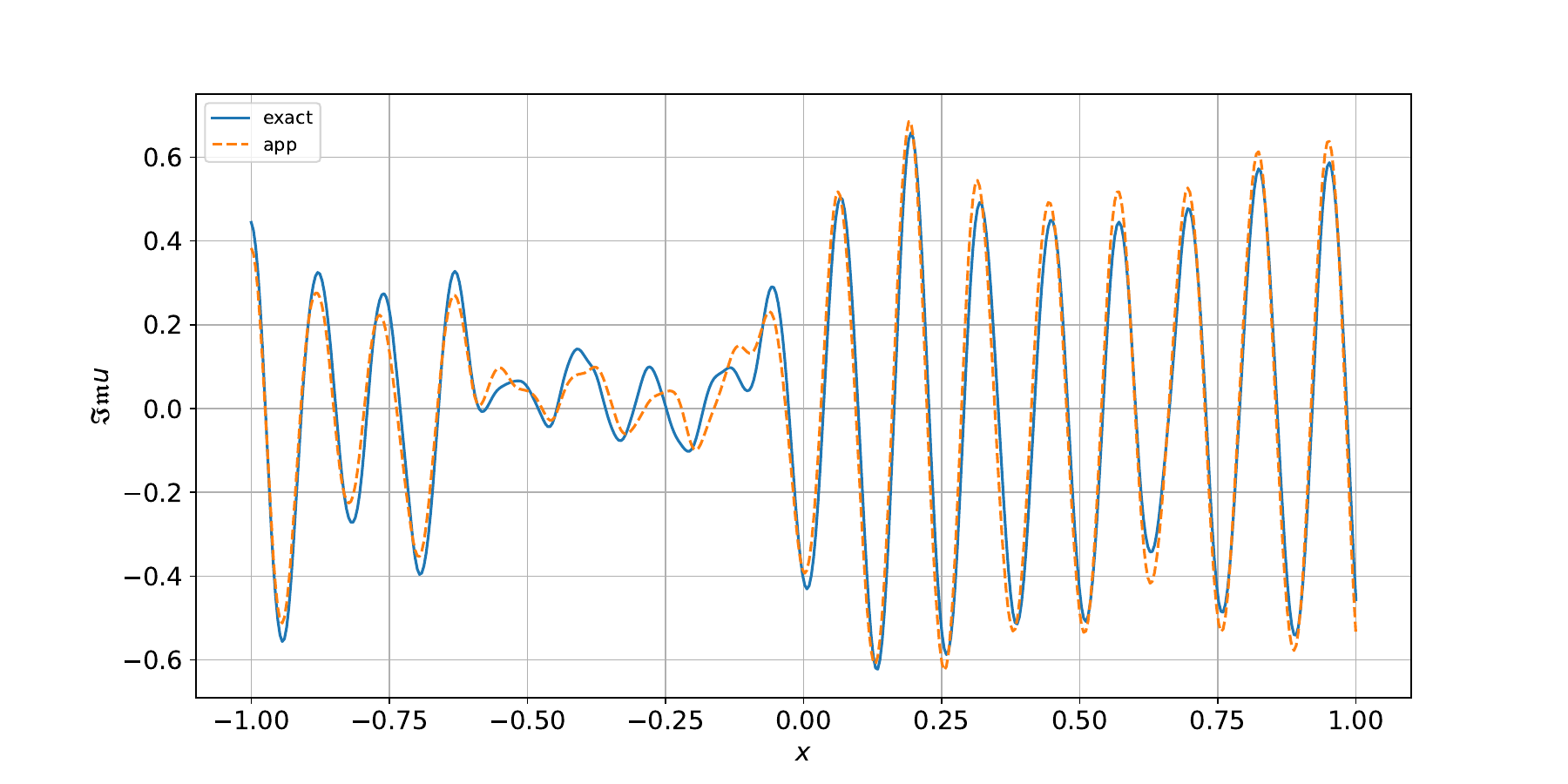}}
\caption{Training results using multi-scale DeepONet ($S_{\rm branch}=1, S_{\rm trunk}=10$) for the case $M=50, k=50$.}%
\label{test_2_fig_3}%
\end{figure}

\begin{figure}[ht!]
\center
\subfigure[real part on training data]{\includegraphics[scale=0.25]{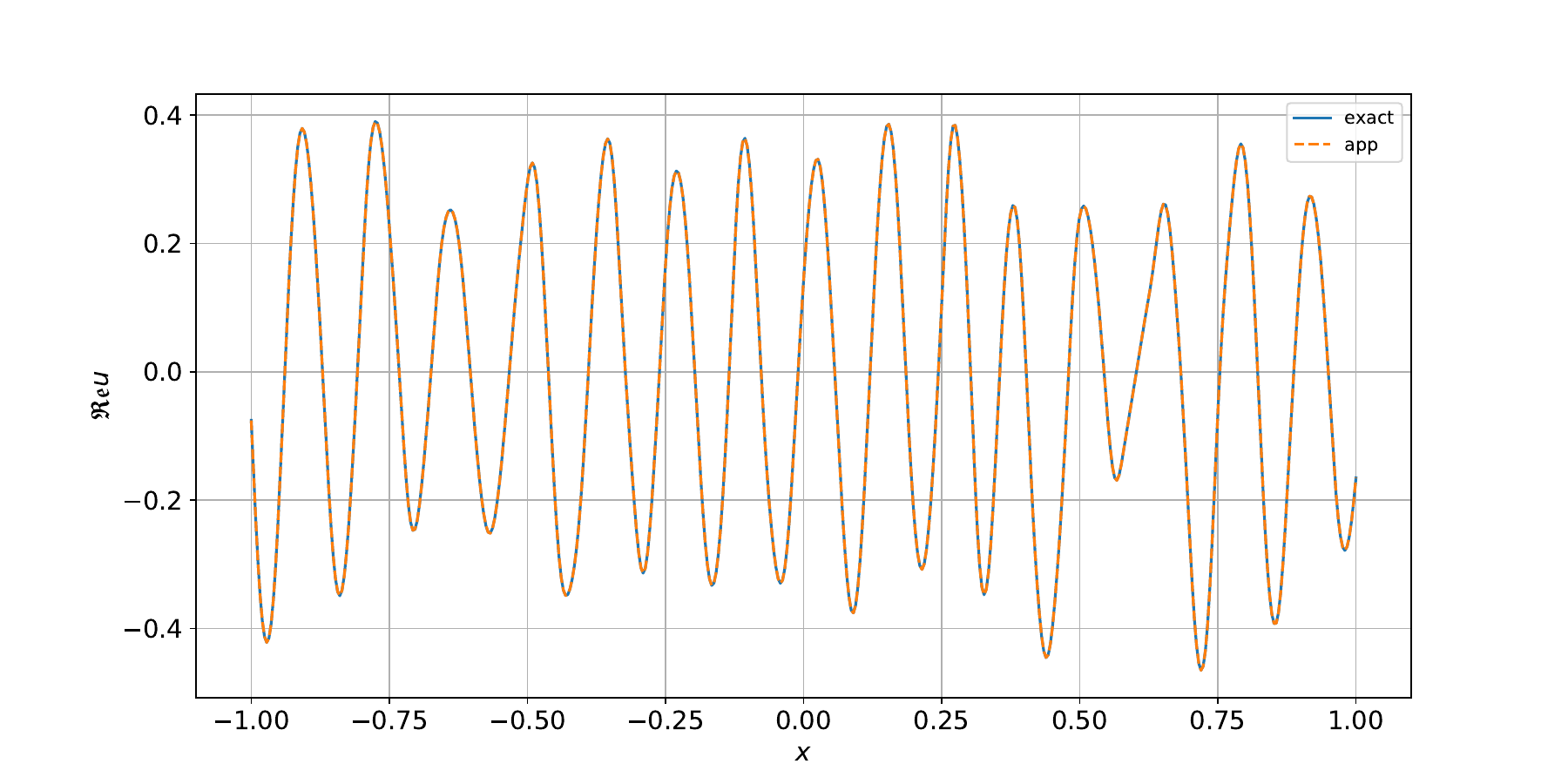}}
\subfigure[real part on test data]{\includegraphics[scale=0.25]{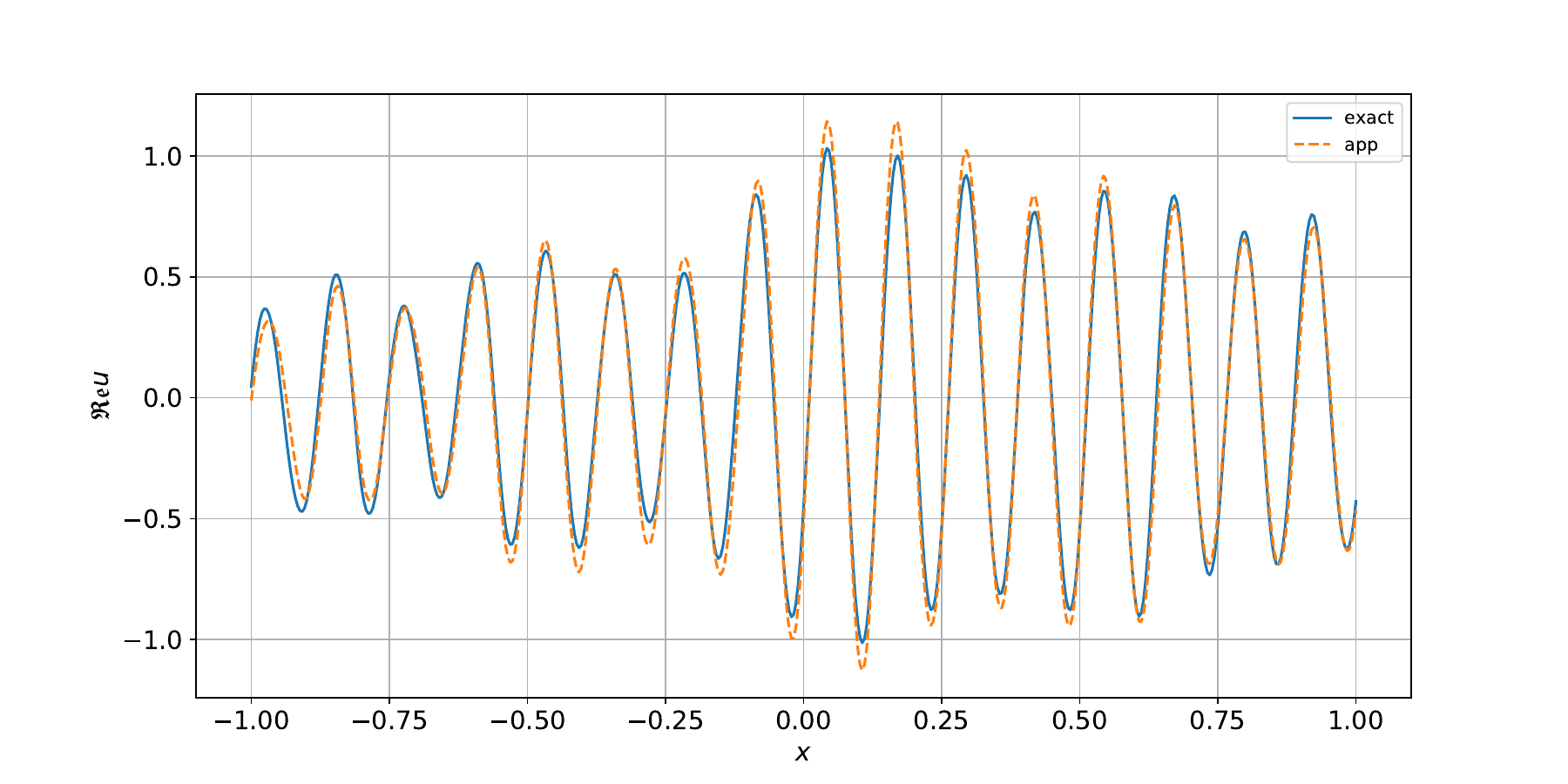}}
\subfigure[imaginary part on training data]{\includegraphics[scale=0.25]{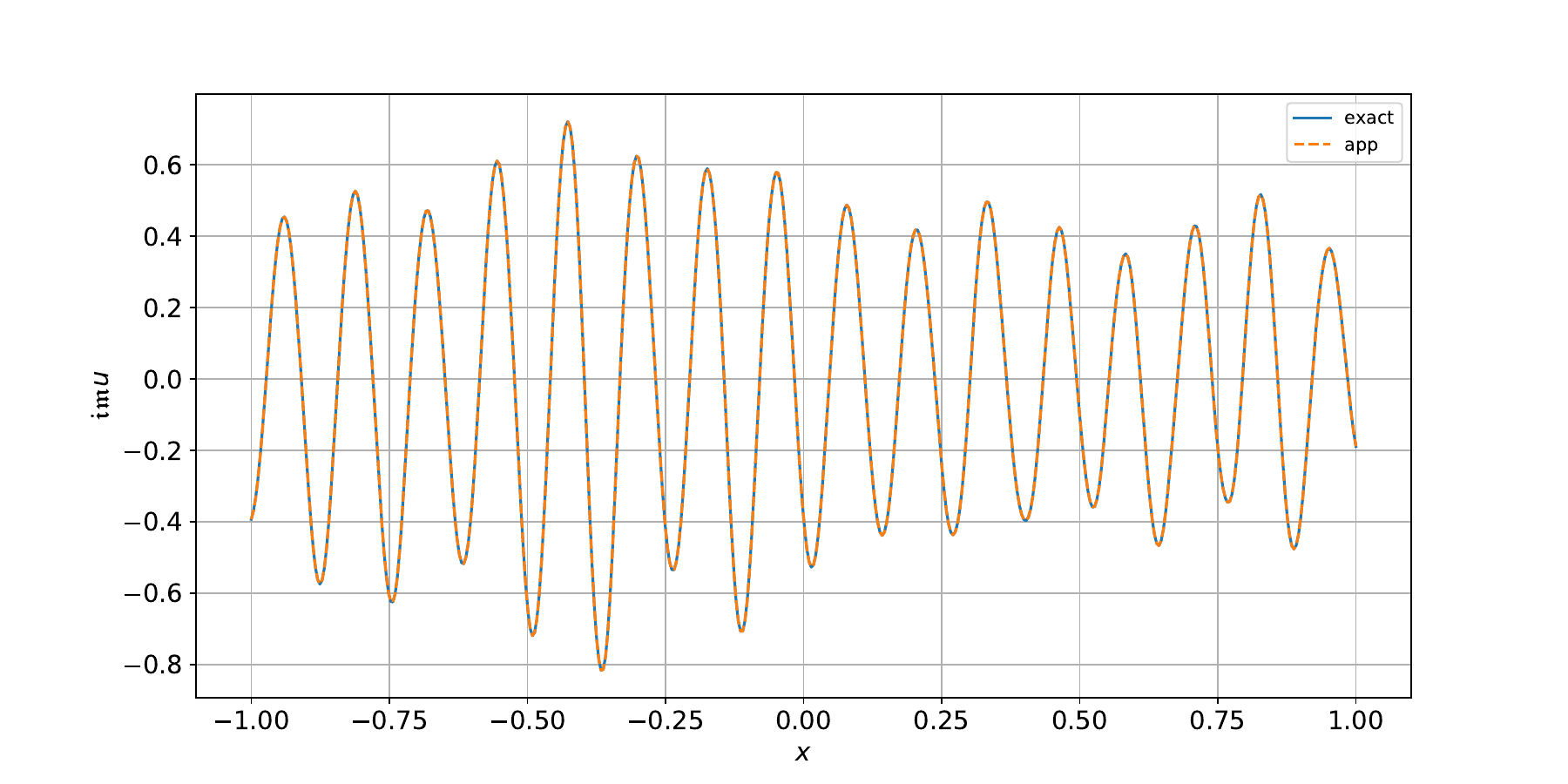}}
\subfigure[imaginary part on test data]{\includegraphics[scale=0.25]{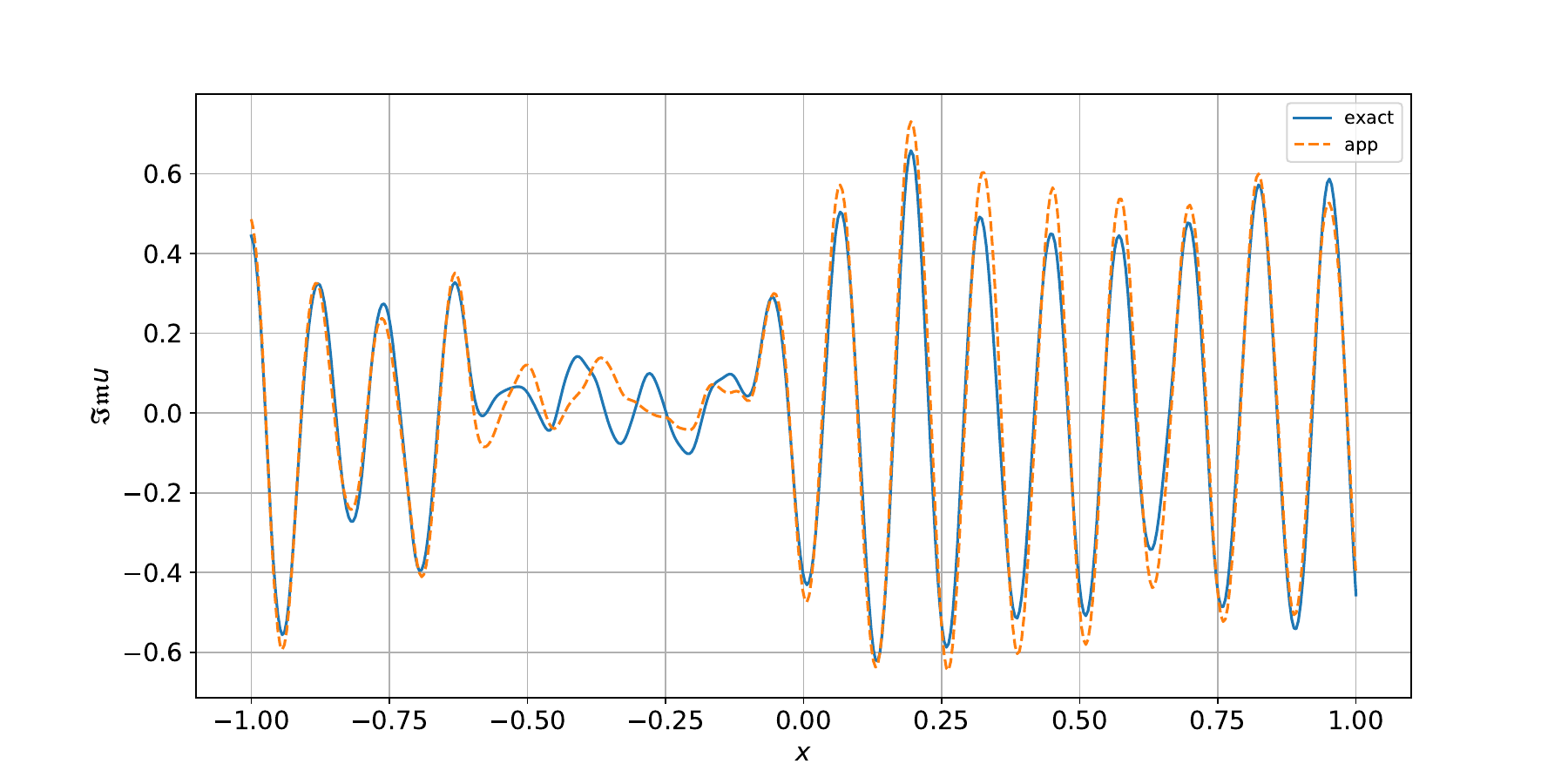}}
\caption{Training results using multi-scale DeepONet ($S_{\rm branch}=5, S_{\rm trunk}=10$) for the case $M=50, k=50$.}%
\label{test_2_fig_4}%
\end{figure}

\begin{figure}[ht!]
\center
\subfigure[real part on training data]{\includegraphics[scale=0.25]{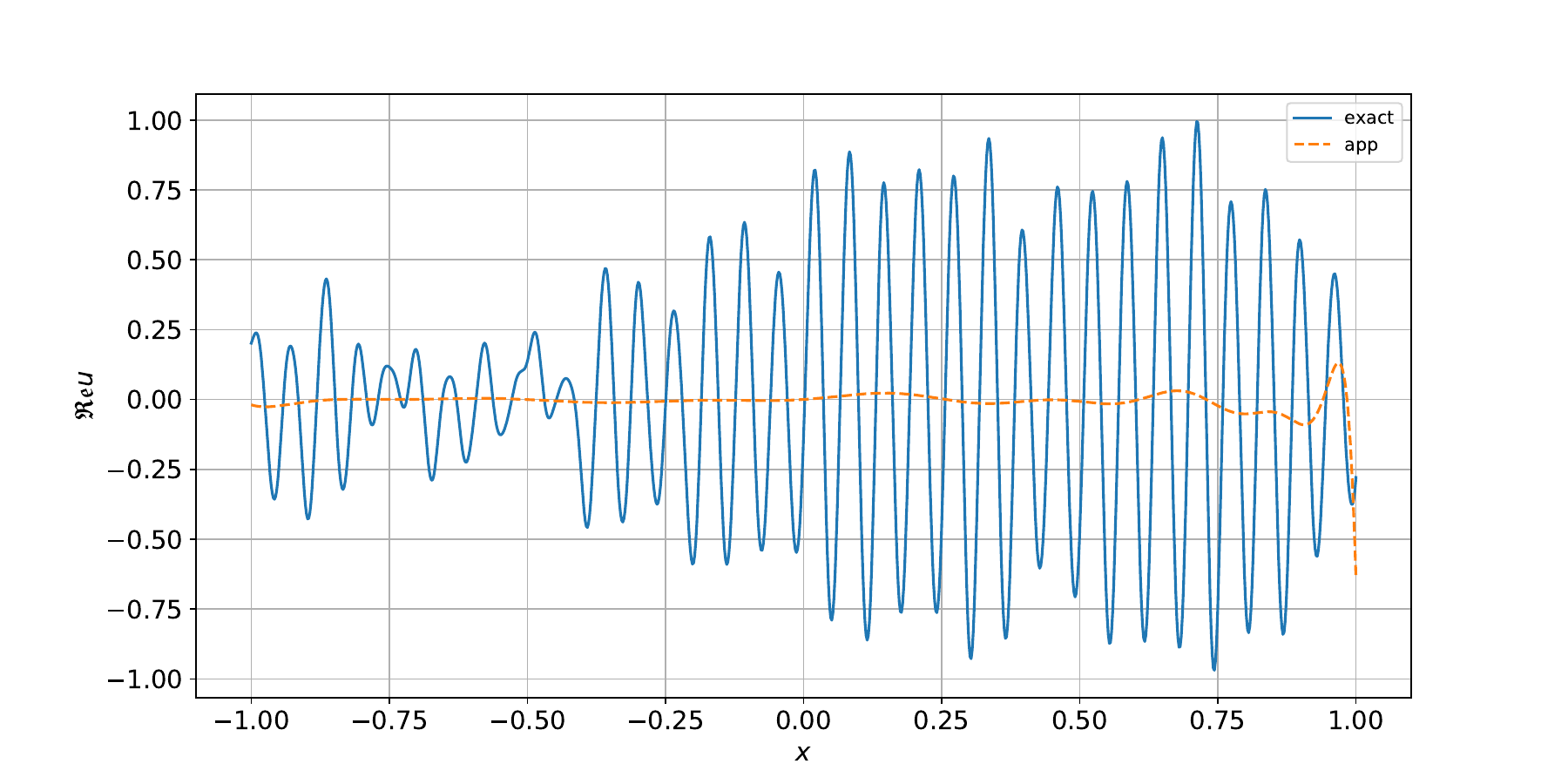}}
\subfigure[real part on test data]{\includegraphics[scale=0.25]{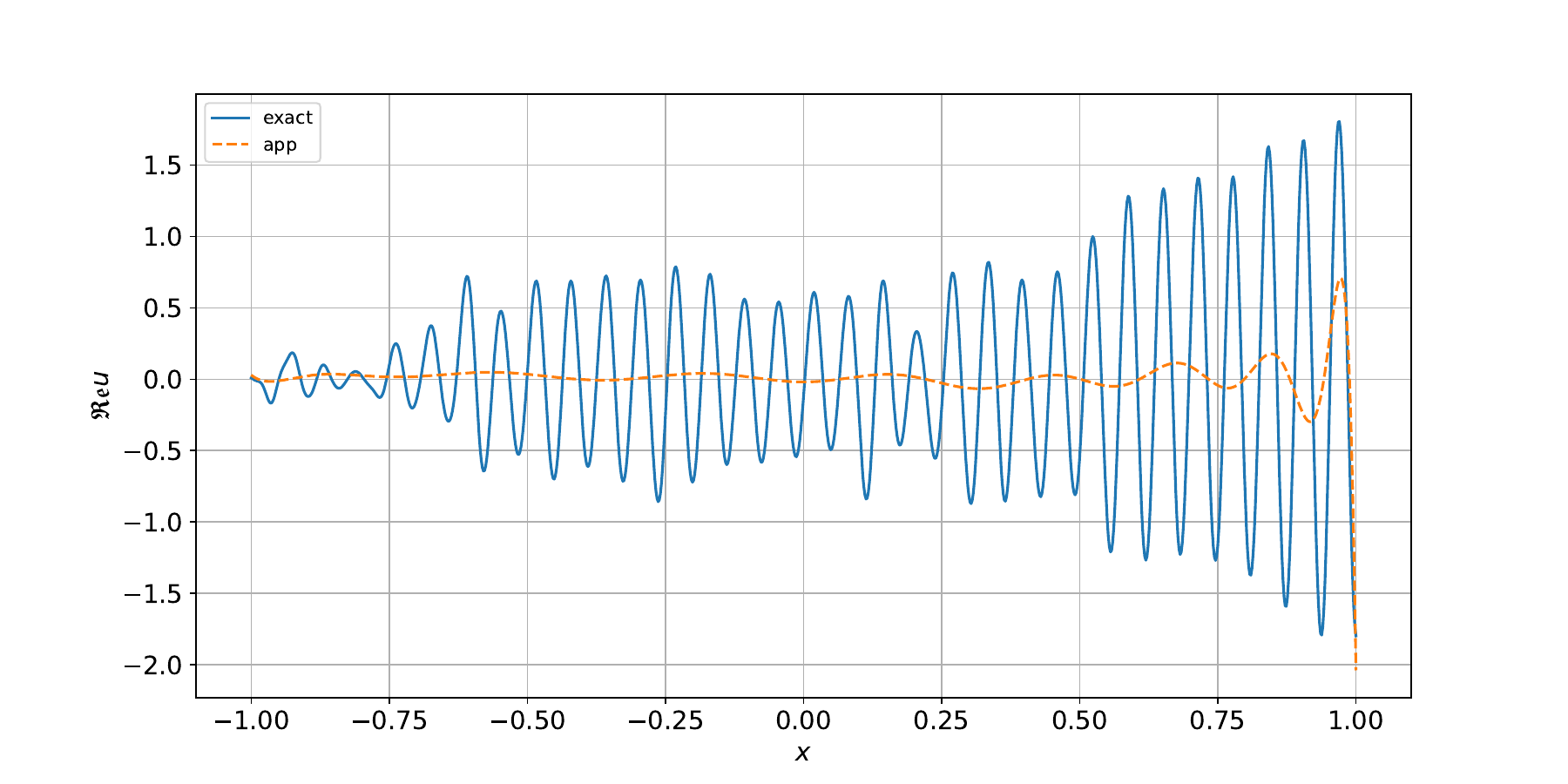}}
\subfigure[imaginary part on training data]{\includegraphics[scale=0.25]{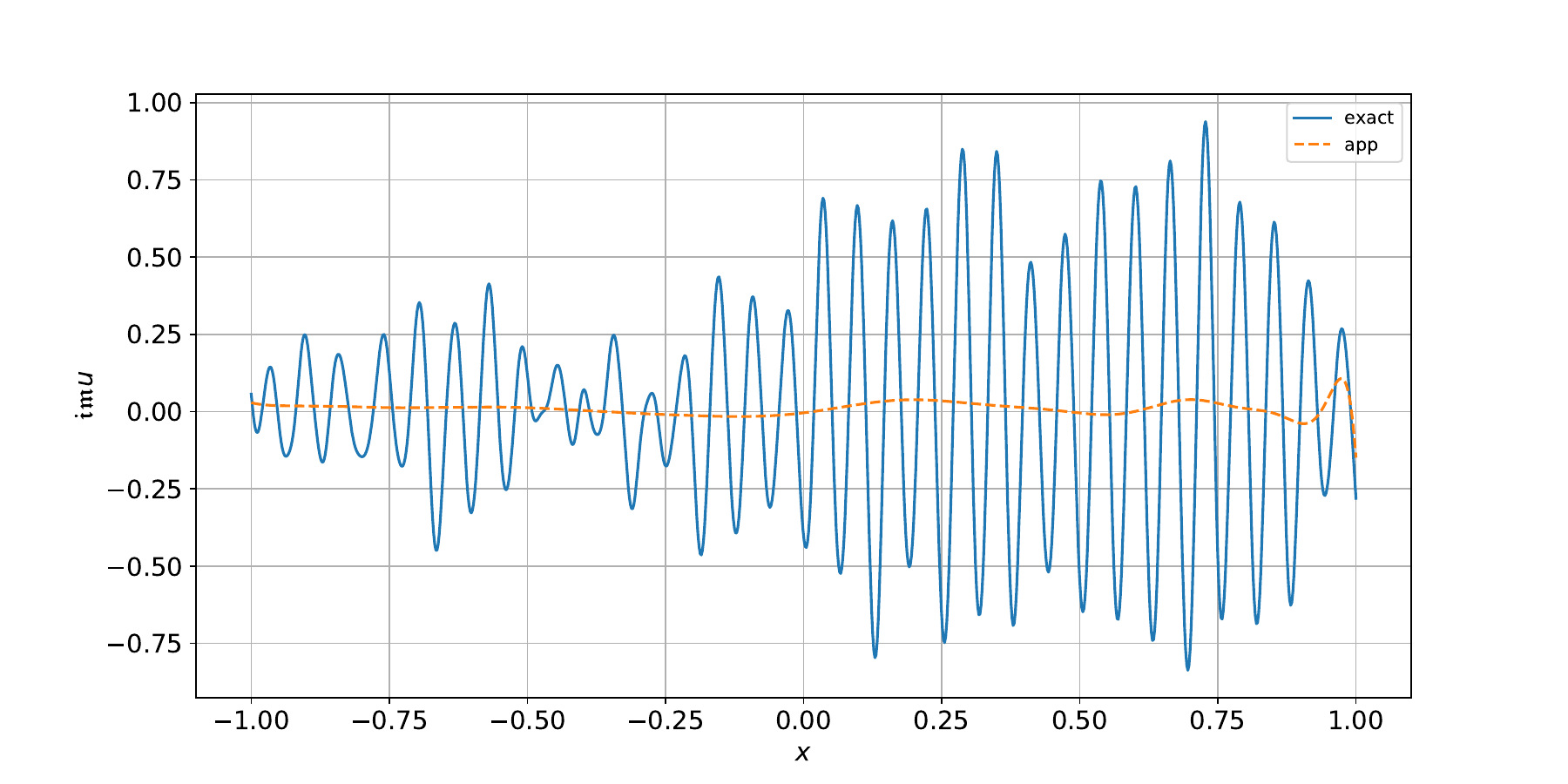}}
\subfigure[imaginary part on test data]{\includegraphics[scale=0.25]{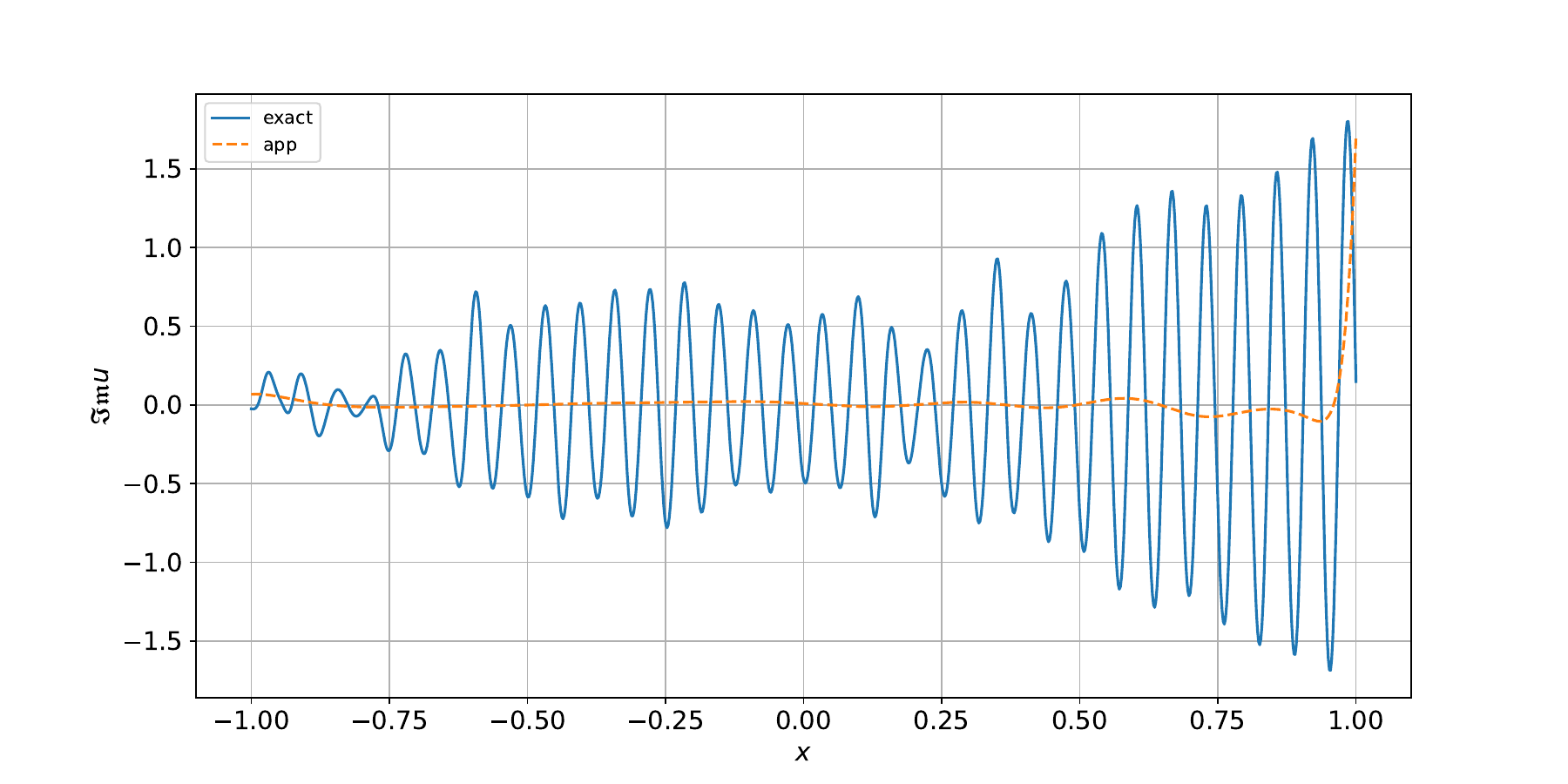}}
\caption{Training results using DeepONet ($S_{\rm branch}=1, S_{\rm trunk}=1$) for $M=50, k=100$.}%
\label{test_2_fig_6}%
\end{figure}

\begin{figure}[ht!]
\center
\subfigure[real part on training data]{\includegraphics[scale=0.25]{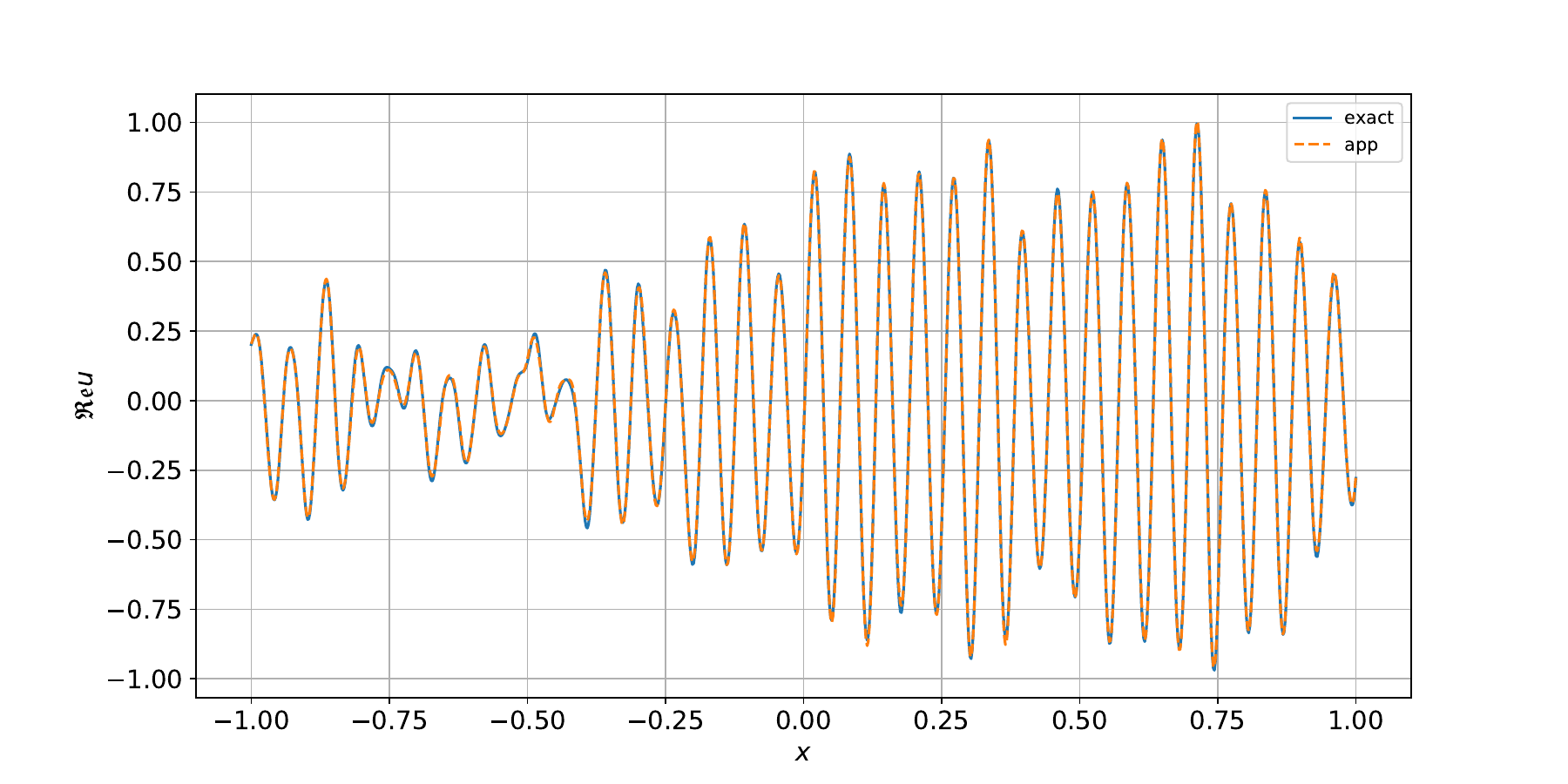}}
\subfigure[real part on test data]{\includegraphics[scale=0.25]{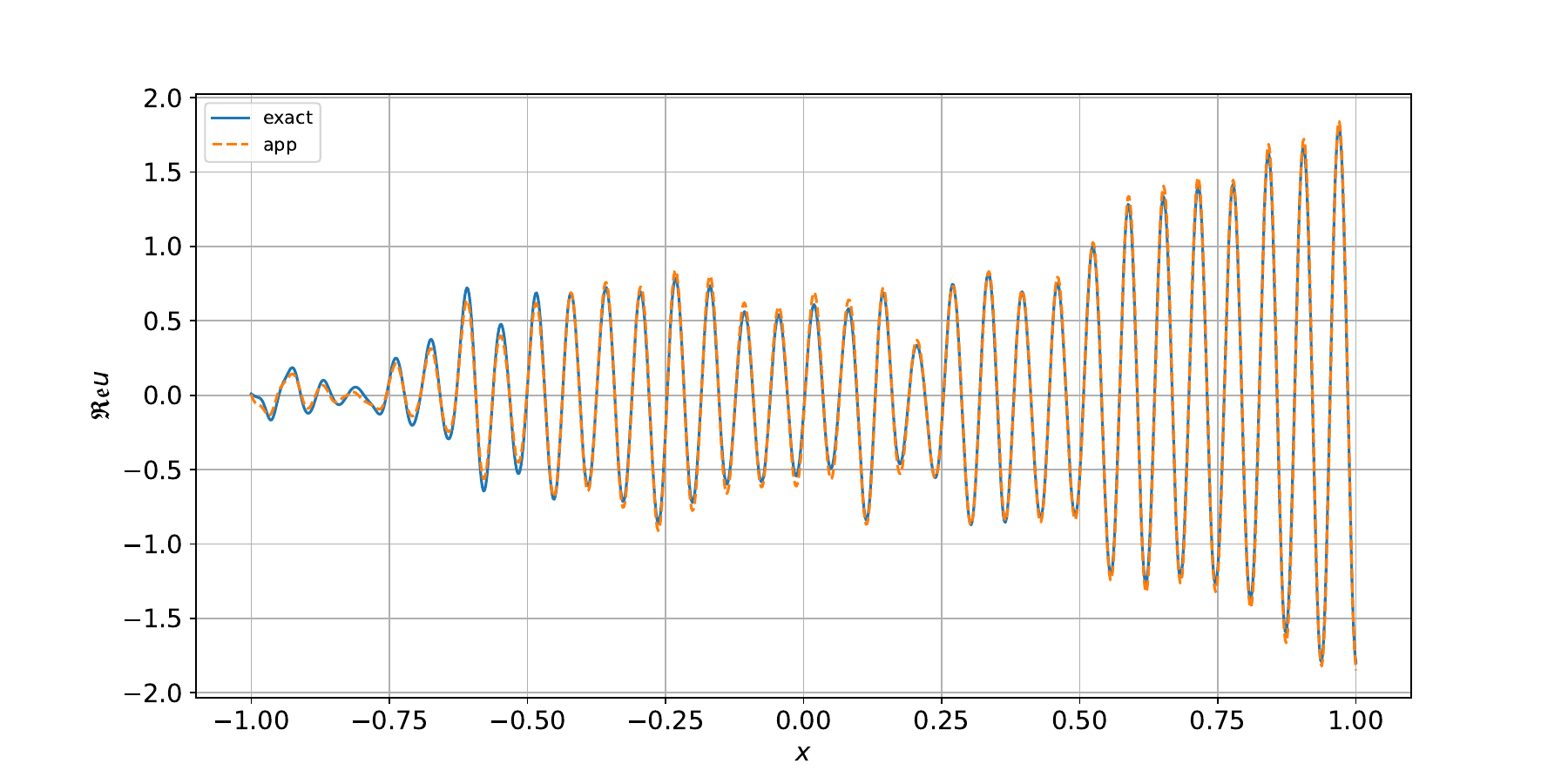}}
\subfigure[imaginary part on training data]{\includegraphics[scale=0.25]{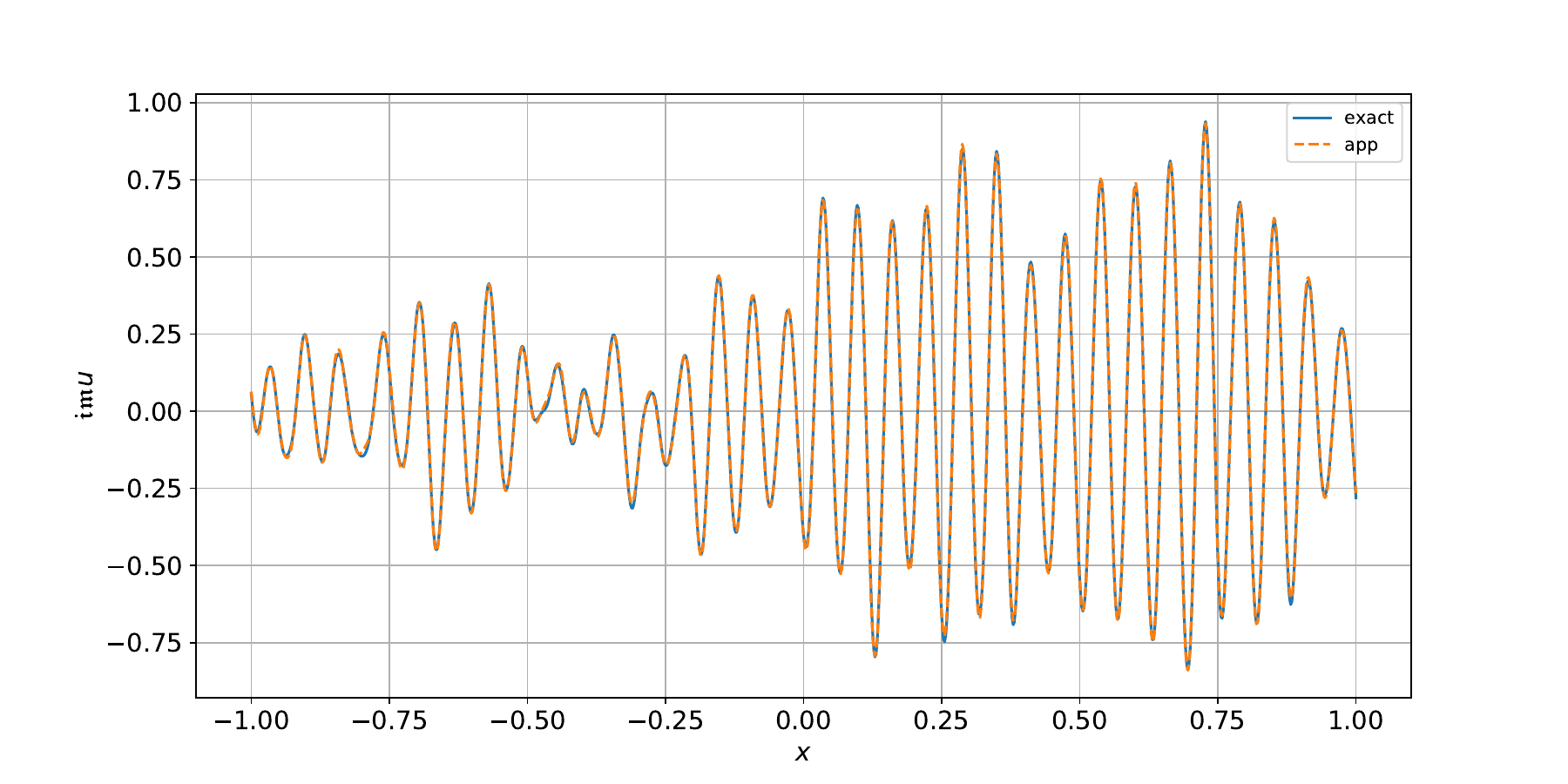}}
\subfigure[imaginary part on test data]{\includegraphics[scale=0.25]{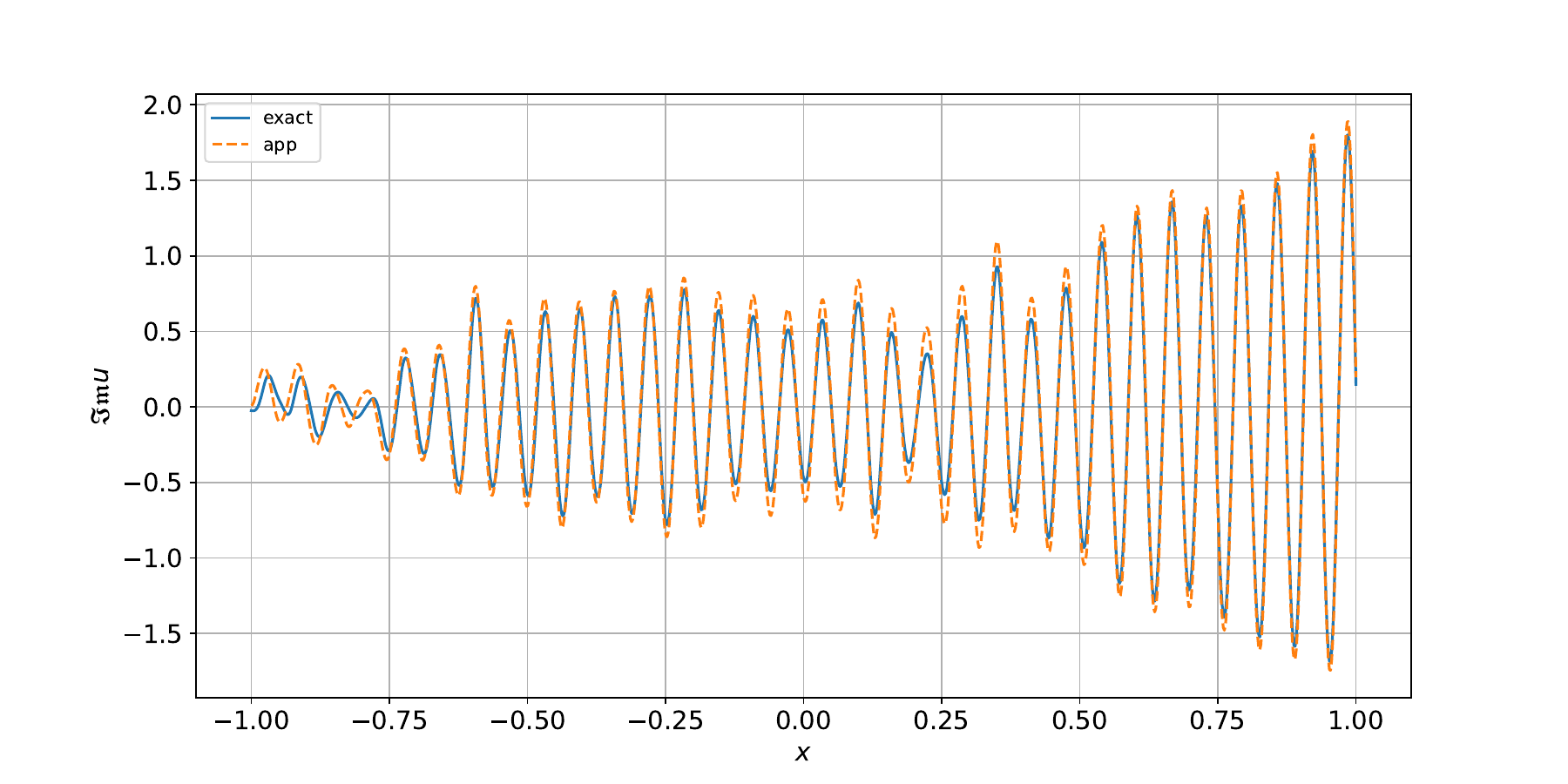}}
\caption{Training results using multi-scale DeepONet ($S_{\rm branch}=1, S_{\rm trunk}=10$) for the case $M=50, k=100$.}%
\label{test_2_fig_7}%
\end{figure}

\begin{figure}[ht!]
\center
\subfigure[real part on training data]{\includegraphics[scale=0.25]{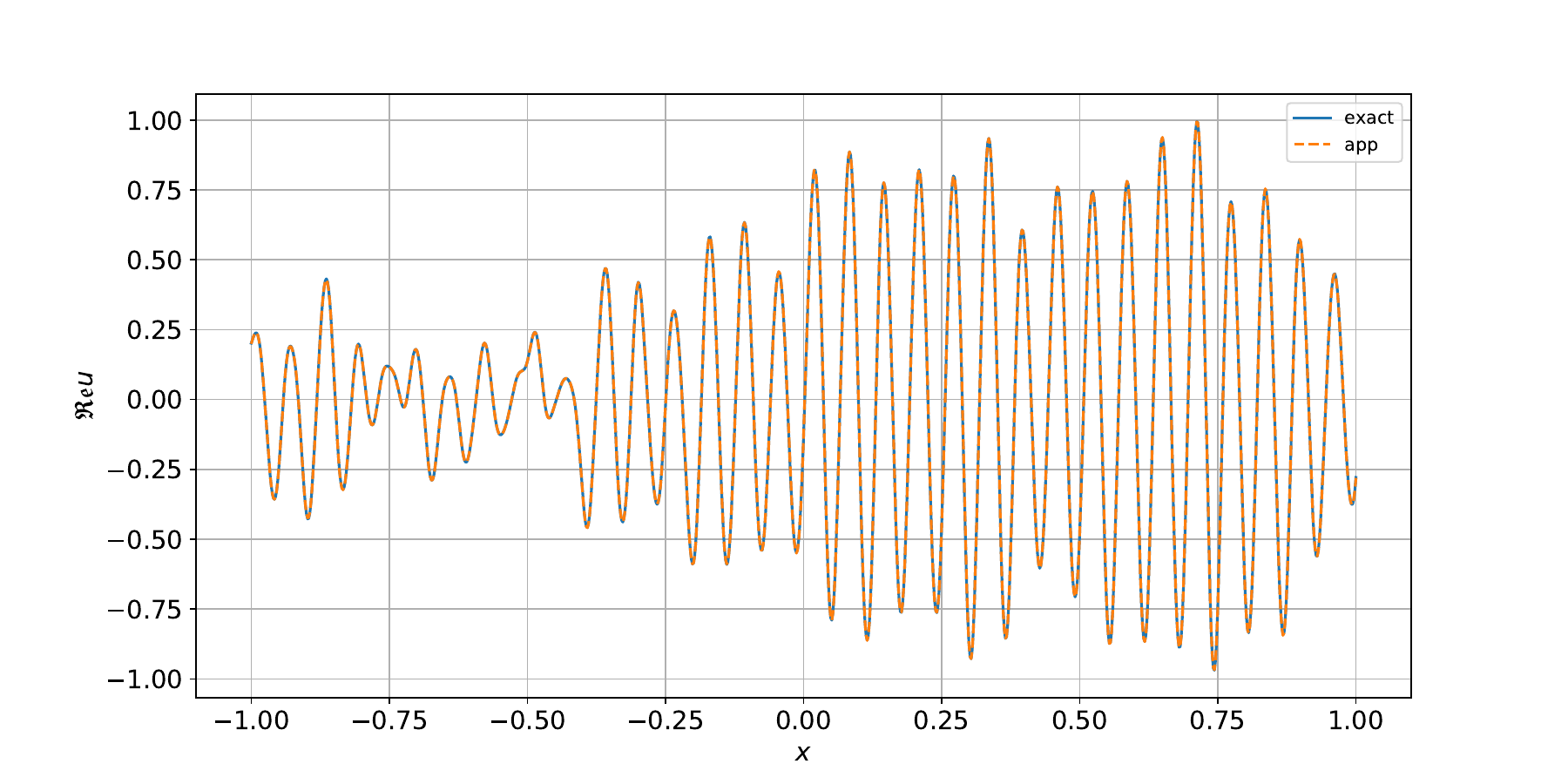}}
\subfigure[real part on test data]{\includegraphics[scale=0.25]{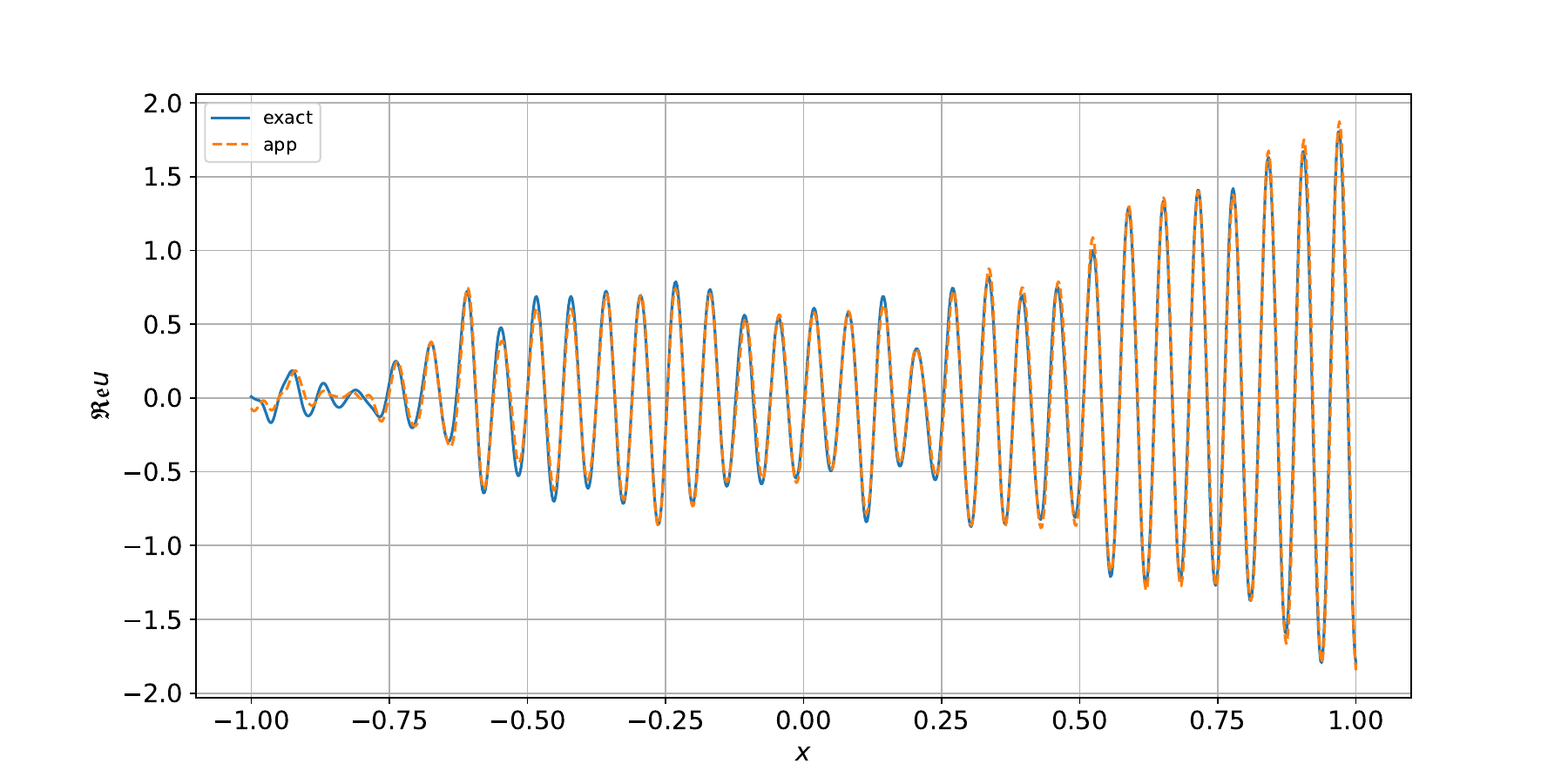}}
\subfigure[imaginary part on training data]{\includegraphics[scale=0.25]{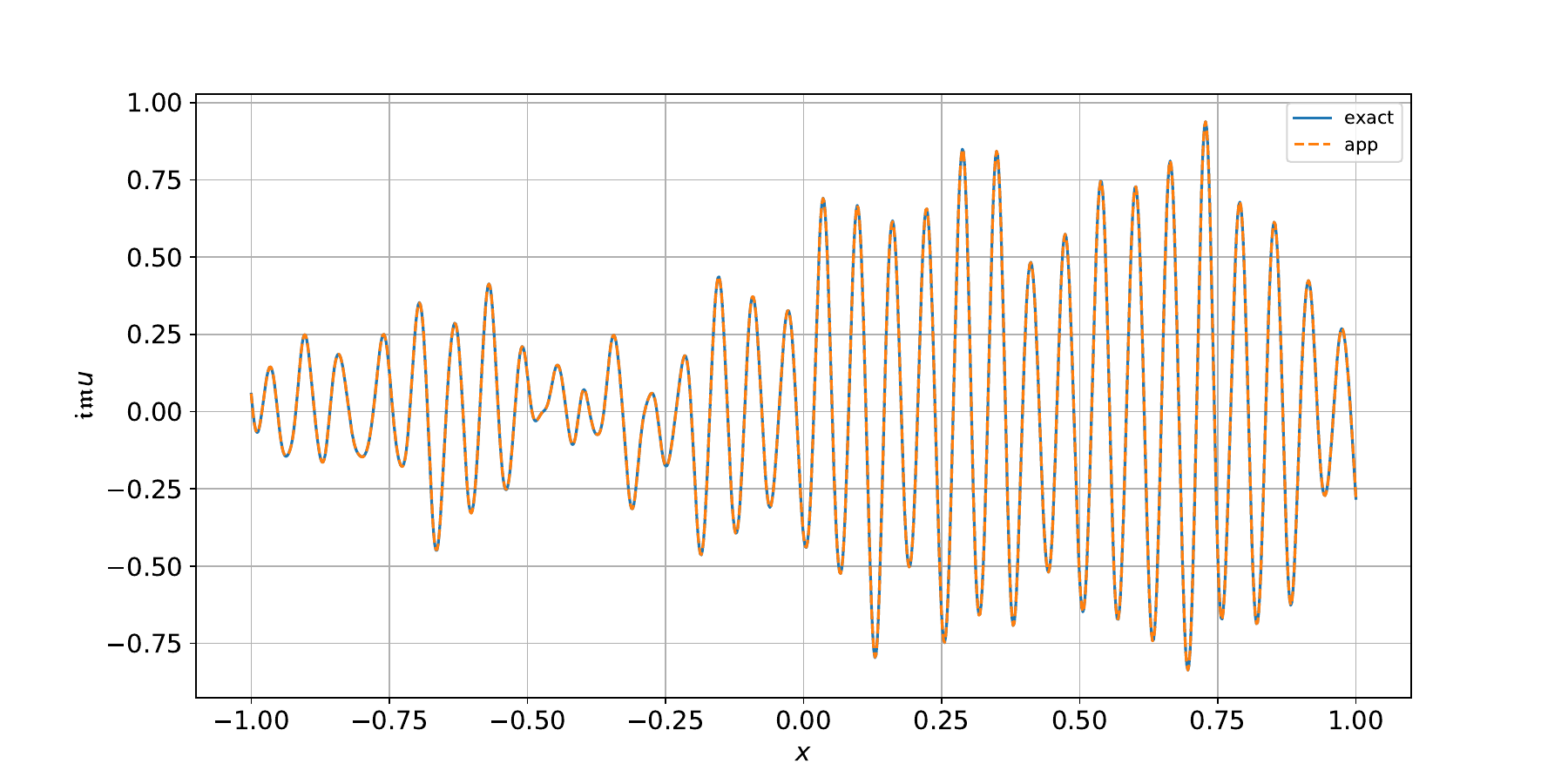}}
\subfigure[imaginary part on test data]{\includegraphics[scale=0.25]{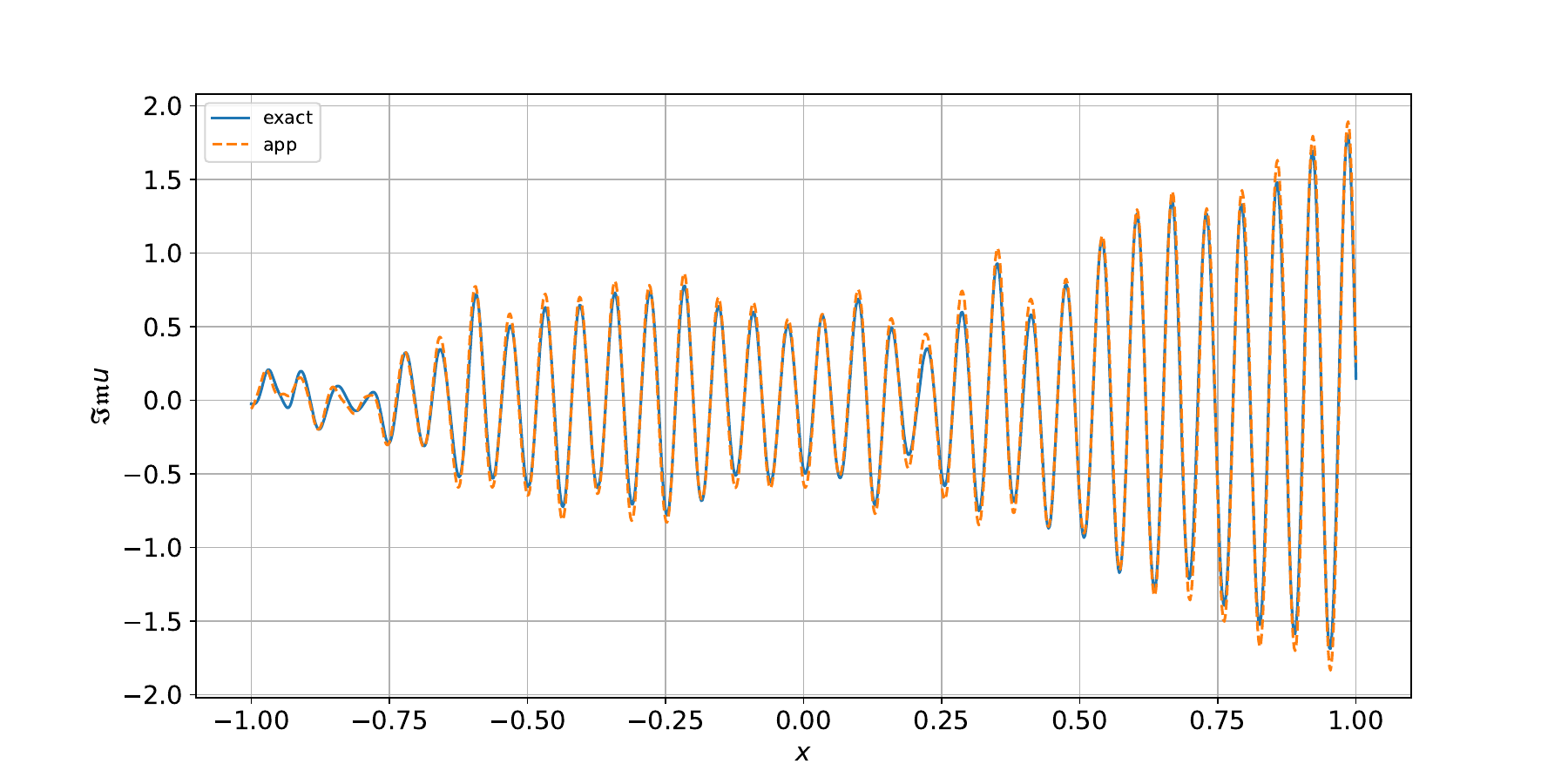}}
\caption{Training results using multi-scale DeepONet ($S_{\rm branch}=5, S_{\rm trunk}=10$) for the case $M=50, k=100$.}%
\label{test_2_fig_8}%
\end{figure}

The training and testing losses for various cases are compared in Fig. \ref{test_1_fig_1}. It is evident that the multiscale technique becomes increasingly important as the wave number $k$ increases for high frequency scattering. A comparison of the performance between the original and multiscale DeepONets is shown in Figs. \ref{test_1_fig_2}–\ref{test_2_fig_8}. In general, the multiscale DeepONet demonstrates superior performance over the original version, particularly in the medium-frequency scenarios.

For the case $M=50,k=100, M=50,k=100$, the original DeepONet fails to learn effectively after 1500 epochs, whereas the multiscale DeepONet achieves high accuracy, comparable to that in the low-frequency case. As with the approximation of  nonlinear mappings, the multiscale technique in the branch network improves accuracy on the training data but may also introduce some overfitting issues.





\section{Conclusion and future work}
In this paper, we have proposed a multi-scale DeepOnet, where multi-scale DNN are introduced in the branch and trunk networks of the normal DeepOnet. Numerical results have shown significant improvement of the capability of the Mscale-DeepOnet in learning high frequency nonlinear mapping including the one between the scatterer's material property and scattering field for high-frequency scattering problems, especially in the regime of high-frequency scattering. 

In future work, we will test the multiscale feature in the branch network for cases where the magnitude of the input function for the operator have large range of values such as in scattering and porous media problems in high contrast materials \cite{hou97}.

\end{document}